%% file: manuscript_icml.tex
\documentclass{article}

\usepackage{microtype}
\usepackage{graphicx}
\usepackage{subfigure}
\usepackage{booktabs} 

\usepackage{color}
\usepackage{hyperref}
\hypersetup{colorlinks=true,citecolor=cyan, linkcolor=blue}

\usepackage[accepted]{icml2025}

\usepackage{amsmath,amsthm,amssymb,amsfonts}
\usepackage{mathtools}
\usepackage{bm}

\usepackage{multirow}
\usepackage{multicol}
\usepackage{makecell}
\usepackage{tcolorbox}
\usepackage{wrapfig}
\usepackage{hhline}
\usepackage{colortbl}
\definecolor{LightGray}{gray}{0.85}
\definecolor{White}{gray}{1.0}
\definecolor{Celadon}{RGB}{175, 225, 175}
\definecolor{mygreen}{RGB}{240, 248, 255}
\definecolor{myorange}{RGB}{255, 240, 240}
\definecolor{myyellow}{RGB}{255, 240, 240}

\definecolor{cleanModel}{RGB}{198, 219, 239}  
\definecolor{poisonedModel}{RGB}{255, 200, 200}  

\definecolor{seabornOrange}{RGB}{221, 132, 82}   
\definecolor{seabornGreen}{RGB}{85, 168, 104}    
\definecolor{seabornPurple}{RGB}{129, 114, 179}  

\newcommand{\light}{\cellcolor{poisonedModel}}

\newcommand{\grc}{\cellcolor{cleanModel}}

\newtcolorbox{mycolorbox}{
  colback=cleanModel,
  colframe=cleanModel!70!black,
  boxrule=1pt
}
\newtcolorbox{mycolorbox2}{
  colback=poisonedModel,
  colframe=poisonedModel!70!black,
  boxrule=1pt
}

\usepackage[capitalize,noabbrev]{cleveref}

\theoremstyle{plain}
\newtheorem{theorem}{Theorem}[section]

\newtheorem{corollary}[theorem]{Corollary}
\theoremstyle{definition}
\newtheorem{definition}[theorem]{Definition}

\theoremstyle{remark}

\newcommand{\fullname}{Two-stage Symmetry Connectivity}
\newcommand{\shortname}{TSC}
\newcommand{\poisoning}{data-poisoning attack}
\newcommand{\code}{training-manipulation attack}
\newcommand{\filtering}{training-time defenses}
\newcommand{\post}{post-purification defenses}
\newcommand{\test}{test-time defenses}

\usepackage[textsize=tiny]{todonotes}
\usepackage{bbding}

\begin{document}

\twocolumn[
\icmltitle{Circumventing Backdoor Space via Weight Symmetry}



\icmlsetsymbol{equal}{*}

\begin{icmlauthorlist}
\icmlauthor{Jie Peng}{harbin}
\icmlauthor{Hongwei Yang}{harbin}
\icmlauthor{Jing Zhao}{harbin}
\icmlauthor{Hengji Dong}{harbin}
\icmlauthor{Hui He}{harbin}
\icmlauthor{Weizhe Zhang}{harbin,pc}
\icmlauthor{Haoyu He}{mona}

\end{icmlauthorlist}

\icmlaffiliation{harbin}{School of Cyberspace Science, Harbin Institute of Technology, Harbin, China.}
\icmlaffiliation{pc}{Pengcheng Laboratory, Shenzhen, China.}
\icmlaffiliation{mona}{Department of Data Science and AI, Faculty of IT, Monash University, Melbourne, Australia}

\icmlcorrespondingauthor{Hui He}{hehui@hit.edu.cn}

\icmlkeywords{Machine Learning, ICML}

\vskip 0.3in
]



\printAffiliationsAndNotice{} 

\begin{abstract}
Deep neural networks are vulnerable to backdoor attacks, 
where malicious behaviors are implanted during training. 
While existing defenses can effectively purify compromised models, 
they typically require labeled data or specific training procedures, 
making them difficult to apply beyond supervised learning settings. 
Notably, recent studies have shown successful backdoor attacks across various learning paradigms, 
highlighting a critical security concern.
To address this gap, 
we propose \textit{\fullname}\ (\shortname), 
a novel backdoor purification defense that operates independently of data format 
and requires only a small fraction of clean samples.
Through theoretical analysis, we prove that by leveraging permutation invariance in neural networks and quadratic mode connectivity, 
\shortname\ amplifies the loss on poisoned samples while maintaining bounded clean accuracy. 
Experiments demonstrate that \shortname\ 
achieves robust performance 
comparable to state-of-the-art methods in supervised learning scenarios. 
Furthermore, 
\shortname\ generalizes to self-supervised learning frameworks, 
such as SimCLR and CLIP, maintaining its strong defense capabilities.
Our code is available at \url{https://github.com/JiePeng104/TSC}.
\end{abstract}

\vspace*{-2em}
\input{sections/1_introduction.tex}

\input{sections/2_related_work.tex}

\input{sections/3_method.tex}

\input{sections/4_experiments.tex}

\section*{Acknowledgements}
This work was supported in part by
the National Key Research and Development Program of China (Grant No. 2024YFB31NL00101), 
the National Natural Science Foundation of China (Grant No. U22A2036), 
and the National Natural Science Foundation of China (Grant No. 62472122).

\section*{Impact Statement}
This paper aims to advance the field of Machine Learning Security by proposing a novel purification method for removing implanted backdoors. 
The potential societal benefits include providing a 
framework for eliminating backdoor behavior across various machine learning scenarios, 
thereby enhancing model security.
However, new attacks targeting our method may emerge in the future. 
Further work is needed to validate the effectiveness of our approach on a broader scale.

\bibliography{manuscript_icml}
\bibliographystyle{icml2025}

\newpage
\appendix
\onecolumn


\input{sections/5_appendix.tex}

\end{document}

%% file: sections/1_introduction.tex
\section{Introduction}
Modern classifiers require substantial data and computational resources to achieve high accuracy,
providing adversaries with opportunities to implant backdoors into deep neural networks \cite{badnets,TargetGlasses}.
This vulnerability arises from either injecting poisoned data \cite{badnets, firstCleanLabel} (\textit{i.e.,} \textit{\poisoning}), or manipulating the training process \cite{Tale,Inputaware} (\textit{i.e.,} \textit{\code}).
For instance, recent advancements in self-supervised learning strategies often rely on large volumes of training data, 
which, while circumventing the need for curated or labeled datasets, 
can be time-intensive and incur high computational costs \cite{clip, MAE, simCLR}.
Therefore, 
many users prefer to delegate model training to third-party providers
or fine-tune publicly available models on downstream tasks. 
This practice exposes models to backdoor attacks. 
For example, malicious providers can manipulate the training process to implant backdoors 
and then force downstream classifiers 
to output adversarial labels \cite{Latent, backdoor_ssl_cvpr, embarrassing}.

\begin{figure}[t]
    \centering
    \includegraphics[width=1\linewidth]{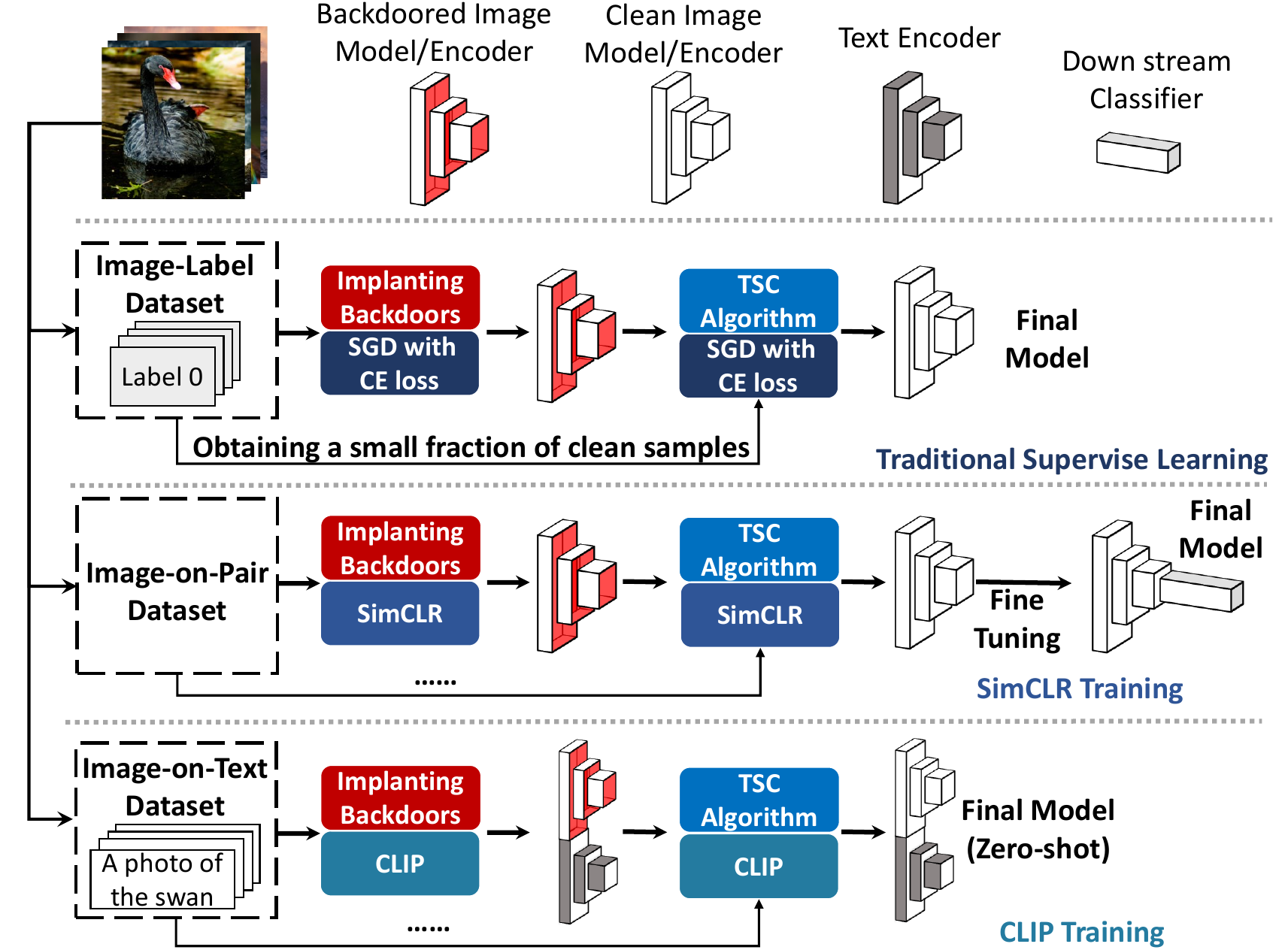}
    \vspace*{-1.2em} 
    \caption{
        Illustration of the application of \shortname~ in three popular learning settings. 
        We assume adversaries can perform either \poisoning~ or \code~ to implant a backdoor into the weights of a classification model or an encoder. 
        Unlike most existing defenses requiring specific training procedures,
        \shortname~ provides a framework to remove backdoors using only a small fraction of clean samples and the original training process.
        For instance, in SimCLR \cite{simCLR} setting, 
        a \shortname~ defender can directly remove the backdoor hidden in the image encoder by combining \shortname~ with the SimCLR training procedure. 
        This allows for training the downstream classifier without the adversarial influence inherited from upstream.
    }
       \label{fig:overview}
       \vspace*{-1.2em}
  \end{figure}

Consequently, 
many defenses have been developed to eliminate 
backdoors hidden within models. 
One of the most prominent mechanisms is \textit{\post},
which remove backdoors through post-training processes \cite{FinePruning, NeuralCleanse, ANP,I_BAU}. 
Usually, \post\ require only a small amount of clean data and are effective against both data-poisoning and training-manipulation attacks. 
However, existing approaches focus primarily on supervised learning scenarios and rely on training methods requiring labeled data \cite{NeuralCleanse,ANP,NAD,AWM} 
(\textit{e.g.,} methods working as analogues to adversarial training). 
Thus, they are not directly applicable to learning regimes like self-supervised or unsupervised learning.
Moreover, some studies have shown that most current defense mechanisms,
whether categorized as \post~ or not, 
are vulnerable to attacks using small poisoning rates or adaptively designed triggers \cite{SPECTRE, adap_Backdoor, towardsStable}.

Previous studies have intentionally or unintentionally tried to address the challenges of robustness and transferability in backdoor defense. 
Recent work by \citet{towardsStable} re-investigated fine-tuning (FT) based methods, 
proposing to enhance robustness under low poisoning rates. 
While achieving promising performance, 
their methods remain confined to supervised learning settings. 
The mode connectivity repair (MCR) \cite{MCR}, 
which leverages quadratic mode connectivity \cite{ModeConnectivity}, 
offers a data-format agnostic procedure but shows limited robustness against various attack methods. 
To address all these challenges, in this paper, we explore
\textit{a robust post-purification defense that operates independently of the data format.}

We propose \fullname\ (\shortname), 
a novel defense mechanism that leverages weight symmetry in neural networks to guide compromised models away from backdoor behaviors without directly unlearning backdoor patterns (\textit{i.e.,} circumventing the backdoor space). 
Our approach builds upon two key properties: 
permutation invariance, 
which allows equivalent model representations through weight layer permutations \cite{permutation}; 
and quadratic mode connectivity \cite{ModeConnectivity}, 
which connects model states through low-loss quadratic paths.

Our defense process begins by projecting a copy of the compromised model into a distinct yet symmetrical loss basin, leveraging the permutation invariance property of neural networks. 
These two models serve as endpoints for a Bézier curve that is trained using a small clean dataset. 
We then pick a point on this trained curve as the purified model, which completes the first stage of our defense.
As the endpoints are in different loss basins and this curve training utilizes only clean samples, we show that the loss of poisoned samples along the curve is amplified. 
Subsequently, we merge the purified model and the backdoored model in the original loss basin to maintain clean accuracy, which completes the second stage of our defense.
\Cref{fig:overview} illustrates how \shortname\ operates 
across different learning settings, 
using only a small fraction of clean samples and 
a procedure aligned with the original training process.
Overall, our contributions are as follows:
\vspace*{-1em}
\begin{itemize}
    \item 
    We discover that the property of weight symmetry enable effective 
    backdoor space circumvention.
    Based on this insight, \shortname\ provides a unified framework for backdoor defense across various learning settings.
    \vspace*{-0.4em}
    \item Through analysis of the mechanisms behind permutation invariance and mode connectivity, 
    we provide theoretical guarantees that \shortname\ can amplify the upper bound of loss values on poisoned samples 
    while maintaining accuracy on the initial task.
    \vspace*{-0.4em}
    \item Experiments on CIFAR10 \cite{CIFAR10}, GTSRB \cite{gtsrb}, and ImageNet100 \cite{ImageNet} under supervised learning demonstrate that \shortname\ 
    achieves performance comparable to state-of-the-art methods while maintaining robustness against various attack settings, 
    including small poisoning rates and adaptively designed attacks.
    Moreover, \shortname\ successfully counters attacks on image encoders across different learning frameworks.
\end{itemize}

%% file: sections/2_related_work.tex
\vspace*{-1em}
\section{Related Work}
\paragraph{Mode Connectivity.}
Merging two models with different initializations usually involves the concept of \textit{mode connectivity}, 
which lies at the heart of finding a low-loss linear or nonlinear path connecting two models \cite{LMC, ModeConnectivity, SGDR_mc, lossLandscape}. 
Empirical studies have shown that aligning two models through permutation before merging can greatly enhance 
the generalization of models along a linear path \cite{permutation, geometry_lossLandscape}, or a quadratic curve \cite{optMC_na}.
Notably, recent research demonstrates that models aligned within the same loss basin can be merged effectively by averaging their weights \cite{git_rebasin, model_fusion, repair}.

\vspace*{-0.8em}
\paragraph{Backdoor Attacks.}

A backdoor adversary aims to make the victim model maintain accuracy on clean inputs while assigning target labels to trigger-embedded inputs \cite{badnets, TargetGlasses}.
Based on the attacker's capabilities, 
backdoor attacks can be categorized as \textit{\poisoning} and \textit{\code}.
In \textit{data-poisoning attacks}, 
the adversary poisons a portion of the training set 
by injecting pre-designed triggers and modifying their labels. 
The trigger patterns can range from pixel squares to real-world objects \cite{FindingNaturally} or invisible patterns \cite{BackdoorStega, HiddenTrigger, ISSBA}.
To avoid detection of mislabeled samples as outliers, some studies have developed backdoor attacks that maintain original labels \cite{firstCleanLabel, SIG, PoisonFrog, Transferable_clean-label}.
\textit{Training-manipulation} attackers have full access to the training process, 
enabling them to effectively inject backdoors through specific training-based methods \cite{Inputaware,Tale}.

Most backdoor attack studies focus on supervised learning scenarios.
However, recent work has indicated that models in self-supervised learning settings are vulnerable to backdoor attacks, 
which may involve the adversary controlling the training process \cite{BadEncoder}, or merely injecting poisoned samples \cite{backdoor_ssl_cvpr, poisoningCLIP, embarrassing,web_scale_poisoning}.

\vspace*{-0.8em}
\paragraph{Backdoor Defenses.}

Current backdoor defenses can be divided into three categories: \textit{\filtering}, \textit{\test}, and \textit{\post}. 
Training-time defenses require training clean models on a polluted dataset to counteract data-poisoning attacks \cite{Clustering, PCA, SPECTRE, ABL, Rethinking_BD}.
A recent approach, ASSET \cite{asset}, extends this concept to handle various attacks and learning settings.
Test-time defenses aim to filter out malicious inputs during inference  
rather than directly eliminating backdoor threats \cite{STRIP, scaleup, IBD}.

In this study, we focus on \textit{\post} \cite{FinePruning, ANP, I_BAU, NeuralCleanse, towardsStable} aiming to remove the backdoor injected into the weights of a model. 
Existing visual \textit{\post}~ mostly require labels accompanied with the training images, such as adversarial training based methods \cite{ANP, I_BAU}, 
limiting their applicability to other training frameworks without labels.
While \citet{detect_encoder} proposed a method to identify backdoors in self-supervised learning settings, it focuses on detection rather than removal.

Recently, \citet{SSL_Cleanse} extended unlearning methods to self-supervised settings. 
Instead of unlearning backdoor patterns,
our method leverages weight symmetry to purify models while preserving performance, 
generalizing robustly across various learning scenarios and attack settings.

%% file: sections/3_method.tex
\vspace*{-0.8em}
\section{Preliminaries}
\subsection{Minimum Loss Path}
\label{pre_mode_connectivity_pre}
\vspace*{-0.5em}
Here, we consider the connecting method with respect to quadratic mode connectivity \cite{ModeConnectivity,optMC_na}.
Let $\bm{\theta}_A$ and $\bm{\theta}_B$ be the weights of two trained models,
and let $\bm{\gamma}_{\bm{\theta}_{A,B}}(t)$
denote a parametric curve connecting $\bm{\theta}_A$ and $\bm{\theta}_B$
such that $\bm{\gamma}_{\bm{\theta}_{A,B}}(0)=\bm{\theta}_A$ and $\bm{\gamma}_{\bm{\theta}_{A,B}}(1)=\bm{\theta}_B$.
To train $\bm{\gamma}_{\bm{\theta}_{A,B}}(t)$,
\citet{ModeConnectivity} proposed finding the set of parameters $\bm{\theta}_{A,B}$
that minimizes the expectation of the loss $\mathcal{L}(\bm{\gamma}_{\bm{\theta}_{A,B}}(t))$ over the distribution $p_{\bm{\theta}_{A,B}}(\cdot)$ on the curve,
\vspace*{-0.3em}
\begin{equation}
    \label{curve_loss}
    \ell({\bm{\theta}_{A,B}}) = \int_{0}^{1} \mathcal{L}(\bm{\gamma}_{\bm{\theta}_{A,B}}(t)) p_{\bm{\theta}_{A,B}}(t) \,\mathrm{d}t, 
    \vspace*{-0.2em}
\end{equation}
where the $p_{\bm{\theta}_{A,B}}(t)$ is the distribution for sampling the models on the curve indexed by $t$.
For simplicity in computation, the uniform distribution $U(0,1)$ is typically chosen as $p_{\bm{\theta}_{A,B}}(\cdot)$.
To characterize the parametric curve $\bm{\gamma}_{\bm{\theta}_{A,B}}(t)$ for $t \in [0,1]$,  
the \textit{Bézier curve} with 3 bends is commonly employed and is defined as follows:
\vspace*{-0.2em}
\begin{equation}
    \label{bezier_curve}
    \bm{\gamma}_{\bm{\theta}_{A,B}}(t) = 
(1 -t)^2  \bm{\theta}_{A} + 2t(1-t) \bm{\theta}_{A,B} + t^2  \bm{\theta}_{B}.
\end{equation}
More details about mode connectivity can be found in \cref{Quadratic_mc}.

\vspace*{-0.5em}
\subsection{Permutation Invariance}
\vspace*{-0.3em}
\label{pre_permutation}
For simplicity,
we consider an \(L\)-layer feedforward neural network with an element-wise activation function $\sigma$ and weights $\bm{\theta}$.
We use $\bm{W}_l$ to denote the weight of the $l$\textsuperscript{th} layer, 
$\bm{x}_0 \in \mathbb{R}^{d_0}$ to represent the input data and \(\bm{y} \in \mathbb{R}^{d_L}\) to indicate the output logits (or features).
The \(L\)-layer feedforward network can be expressed as:
\vspace*{-0.2em}
\begin{equation}
    \label{feedforward_network}
    \bm{f}(\bm{x}_0, \bm{\theta}) = \bm{y}= \bm{W}_{L} \circ \sigma \circ \bm{W}_{L-1} \circ ... \circ \sigma \circ \bm{W}_{1} \bm{x}_0,
    \vspace*{-0.2em}
\end{equation}
Moreover, we denote the $l$\textsuperscript{th} intermediate feature as 
$\bm{x}_l\in \mathbb{R}^{d_l}$ and $\bm{x}_l= \bm{W}_{l}\circ \sigma \circ\bm{x}_{l-1}$.

One of the key techniques central to our defense method is \textit{permutation invariance} of neural networks \cite{permutation,optMC_na,git_rebasin}.
Let $\bm{P}_l\in \Pi_{d_l}$ be a permutation matrix that permutes output feature $\bm{x}_l$ of the $l$\textsuperscript{th} layer,
where $\Pi_{d_l}$ is the set of all possible $d_l \times d_l$ permutation matrices.
Since $\bm{P}_l^{\top} \bm{P}_l=\bm{I}$, 
without changing the final output $\bm{y}$, we can permute $\bm{W}_{l}$ to $\bm{P}_l\bm{W}_{l}\bm{P}_{l-1}^{\top}$.
We denote this operation as $\pi(\bm{\theta}, S(\bm{P}))$, where
$S(\bm{P})=\{\bm{P}_{1}, \bm{P}_{2},...,\bm{P}_{L-1}\}$ is the set of permutation matrices.
Consequently, we obtain a new network with weights $\bm{\theta}^{S(\bm{P})}=\pi(\bm{\theta}, S(\bm{P}))$ defined as:
\vspace*{-0.2em}
\begin{align}
    \label{weights_permutation}
    &\bm{W}_{1}^{S(\bm{P})} = \bm{P}_{1}\bm{W}_{1}; \quad \ \; \bm{W}_{L}^{S(\bm{P})} = \bm{W}_{L}\bm{P}_{L-1}^{\top}; \nonumber \\ 
    &\bm{W}_{l}^{S(\bm{P})} = \bm{P}_{l} \bm{W}_{l} \bm{P}_{l-1}^{\top}, \ \forall \ l \in \{2, 3, ..., L-1\}.
    \vspace*{-0.2em}
\end{align}
The property of a neural network that allows it to be transformed by such permutation, resulting in $f(\bm{x}_0, \bm{\theta}^{S(\bm{P})})=f(\bm{x}_0, \bm{\theta})$,
is known as \textit{permutation invariance}.
This property implies that two networks can be functionally identical even if the arrangement of their neurons within each layer differs.
For example, if two parameter sets, $\bm{\theta}_A$ and $\bm{\theta}_B$, satisfy $f(\bm{x}_0, \bm{\theta}_A)=f(\bm{x}_0, \bm{\theta}_B)$ for any input $\bm{x}_0 \in \mathbb{R}^{d_0}$,
permutation invariance implies that their corresponding layer weights, $\bm{W}_{i}^A$ and $\bm{W}_{i}^B$, can differ.
This observation motivates the concept of \textit{neuron alignment}.

\vspace*{-0.5em}
\subsection{Neuron Alignment}

\vspace*{-0.2em}

Previous studies have found that two functionally identical networks, $\bm{\theta}_A$ and $\bm{\theta}_B$,
when trained independently with the same architecture but using different random initializations or yielding different SGD solutions,
can be misaligned (\textit{i.e.}, their parameters, $\bm{\theta}_A$ and $\bm{\theta}_B$, correspond to different neuron arrangements).
As a result, the loss obtained from a linear interpolation of their parameters can be quite large.

Recently, research \cite{permutation, model_fusion, git_rebasin,repair} has shown that such misaligned networks
could be projected to the same loss basin using a specific set of permutation matrices $S(\hat{\bm{P}})$.
To identify such permutation set $S(\hat{\bm{P}})$ and achieve alignment,
we can minimize a cost function $c_l:\mathbb{R}^{d_l} \times \mathbb{R}^{d_l} \rightarrow \mathbb{R}^{+}$
with respect to the intermediate features $\bm{x}_l^A$ and $\bm{x}_l^B$
to get the $\bm{\hat{P}}_l$ for each layer.
In practice, given a dataset $D$ containing $n$ samples, 
we minimize the sum of $c_l$ across $D$.
The optimal $\bm{\hat{P}}_l$ can then be found by solving: 
\vspace*{-0.3em}
\begin{equation}
    \label{basic_na_minimization}
    \bm{\hat{P}}_l =  \mathop{\arg \min}\limits_{\bm{P}_l\in \Pi_{d_l}} \sum_i^n c_l(\bm{x}_{i,\; l}^{A},\ \bm{P}_l \; \bm{x}_{i,\; l}^{B}).
    \vspace*{-0.2em}
\end{equation}
This problem is a classic example in the field of optimal transport \cite{optimal_transport,model_fusion}, 
which could be solved via the Hungarian algorithm \cite{hungarian}.
Following previous work, we adopt the alignment method proposed by \citet{convergent_learning},
where $c_l=1-\text{corr}(\bm{v},\; \bm{z})$ and $\text{corr}(\cdot,\; \cdot)$ denotes the correlation between two vectors.
This makes the minimization procedure is equivalent to ordinary least squares constrained to the solution space $\Pi_{d_l}$ \cite{git_rebasin,optMC_na}.
Thus, problem (\ref{basic_na_minimization}) can be specified as:
\vspace*{-0.5em}
\begin{equation}
    \label{l2_minimization}
    \bm{\hat{P}}_l =  \mathop{\arg \min}\limits_{\bm{P}_l\in \Pi_{d_l}} \sum_i^n   \left\lVert \bm{x}_{i,\; l}^{A} - \ \bm{P}_l \; \bm{x}_{i,\; l}^{B} \right\rVert^2.
    \vspace*{-0.2em}
\end{equation}
Throughout the paper, we define $M_l(\bm{\theta}_A, \bm{\theta}_B;D)$ to measure the $L_2$ norm feature distance of the $l$\textsuperscript{th} layer between $\bm{\theta}_A$ and $\bm{\theta}_B$ given dataset $D$:
\vspace*{-0.5em}
\begin{equation}
    M_l(\bm{\theta}_A, \bm{\theta}_B;D) = \sum_i^n   \left\lVert \bm{x}_{i,\; l}^{A} - \; \bm{x}_{i,\; l}^{B} \right\rVert^2. 
    \vspace*{-0.2em}
\end{equation}
Therefore, $\bm{\hat{P}}_l$ can be regarded as the optimal solution which has the smallest $M_l(\bm{\theta}_A, \pi({\bm{\theta}_B,S(\bm{P})});D)$
among all $\bm{{P}}_l \in \Pi_l$. 
Additionally, if we denote $\bm{x}_{i,\; l}^{A}$ and $\bm{x}_{i,\; l}^{B}$ as samples from two distributions $\mathbb{P}_{l}^{A}$ and $\mathbb{P}_{l}^{B}$,
the minimal distance in (\ref{l2_minimization}) corresponds to the 2-Wasserstein distance $ W_2(\mathbb{P}_{l}^{A},\ \mathbb{P}_{l}^{B}; D)$ between $\mathbb{P}_{l}^{A}$ and $\mathbb{P}_{l}^{B}$.
Thus, we have $ W_2(\mathbb{P}_{l}^{A},\ \mathbb{P}_{l}^{B}; D) = M_l(\bm{\theta}_A, \pi({\bm{\theta}_B,S(\hat{\bm{P}})});D)$.
A brief explanation can be found in \cref{Wasserstein_distance}.

\vspace*{-0.5em}
\subsection{Threat Model and Evaluation Metrics}
\label{threat_model}
We consider a scenario where an adversary can manipulate a portion of the training data or access the model training procedure. 
The backdoor is assumed to be implanted into the parameters of a standard neural architecture rather than a model with specific malicious modules \cite{Architectural}. 
A \shortname\ defender requires access to a small fraction of clean training samples and and the ability to retrain the model using the original training procedure. 
A discussion about the applicability of the proposed defense setting can be found in \cref{applicability}. 

To evaluate the performance of various backdoor defenses, we consider two primary metrics: \textit{Attack Success Rate} (ASR), the proportion of attack samples misclassified as the target label, 
and \textit{Clean Accuracy} (ACC), the prediction accuracy on benign samples. 
An effective defense should achieve a high ACC while maintaining a low ASR.

\begin{figure*}[t]
    \centering
    \vspace*{-0.1em} 
    \includegraphics[width=0.94\linewidth]{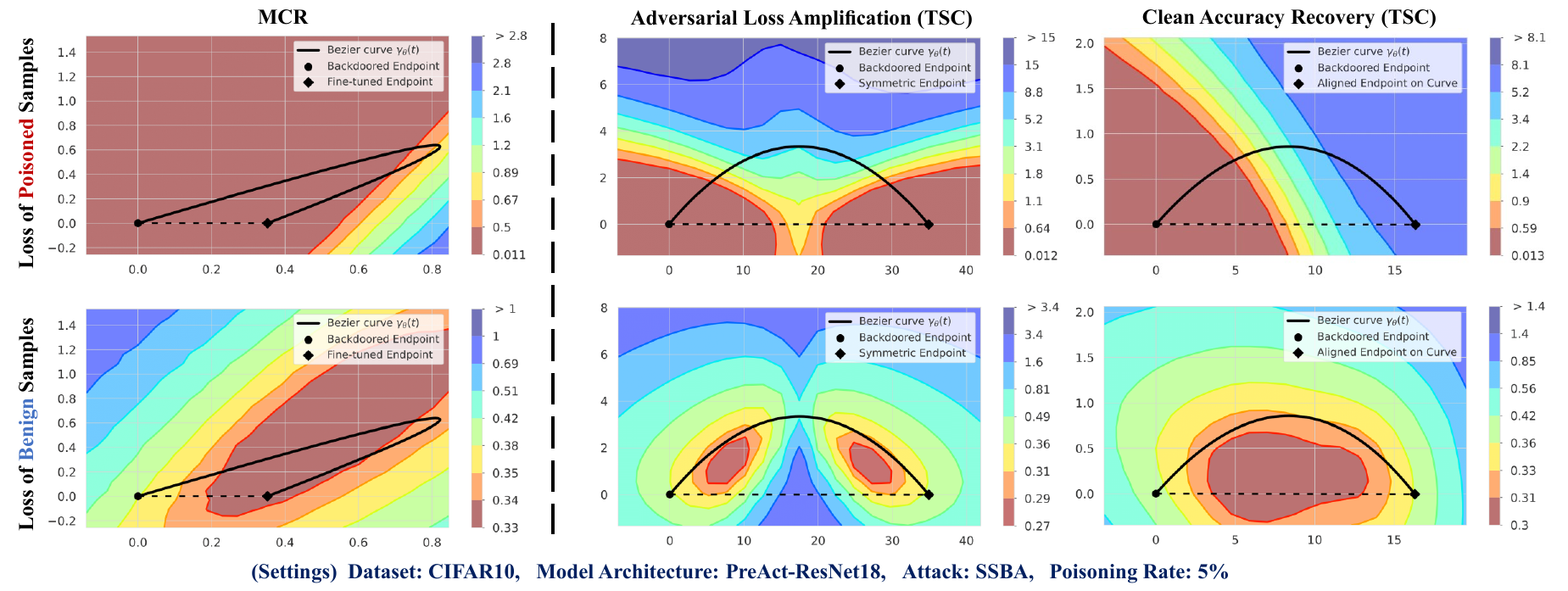}
    \vspace*{-0.8em} 
    \caption{
        Loss landscape for poisoned and benign samples,
        along with trained quadratic curves connecting distinct models.
        The backdoored model is a PreAct-ResNet18 trained on CIFAR-10, which contains 5\% SSBA poisoned samples \cite{ISSBA}.
        \textbf{Left}: the curve identified by MCR.
        \textbf{Middle}: the curve identified by the first stage of \shortname.
        \textbf{Right}: the curve identified by the second stage of \shortname.
    }
       \label{fig:ssba_loss_landscape}
       \vspace*{-0.4em}
\end{figure*}
  
\vspace*{-0.6em}
\section{Method}

Before formally describing our approach, we first provide the intuition behind \shortname. 
Removing the implanted backdoor in a model is equivalent to inducing the model to have a high loss value for the poisoned samples (\textit{i.e., adversarial loss}). 
However, performing a procedure similar to anti-backdoor learning \cite{ABL} to increase the loss of the poisoned samples could be challenging for a post-time defender, 
as the trojan method is unknown, and the adversary could inject various trigger patterns that are difficult to recover \cite{ISSBA}.

To remove the backdoor, we propose a repairing method consisting of two stages of mode connectivity. 
In the first stage, we \textit{amplify the adversarial loss} by un-aligning the malicious model $\bm{\theta}_{adv}$ with its own copy $\bm{\theta}_{adv'}$ 
on the loss landscape and then training a curve $\bm{\gamma}_1$ connecting $\bm{\theta}_{adv}$ and $\bm{\theta}_{adv'}$ with the given benign samples.
Since this unalignment process is designed to place the endpoints ($\bm{\theta}_{adv}$ and $\bm{\theta}_{adv'}$) in different loss basins, and the curve $\bm{\gamma}_1$ is trained exclusively with benign samples, models $\bm{\theta}_t$ along this curve are expected to exhibit high adversarial loss but low loss on benign samples. This outcome is consistent with established properties of quadratic mode connectivity \cite{ModeConnectivity,optMC_na}.
In the second stage, to \textit{recover the clean accuracy}, we train another curve that connects the aligned $\bm{\theta}_{adv}$ and $\bm{\theta}_t$. 
This procedure aims to find a curve $\bm{\gamma}_2$ with descending adversarial loss along the curve but a much lower loss for benign samples compared to $\bm{\gamma}_1$.
\vspace*{-0.5em}
\subsection{Adversarial Loss Amplification}
\label{first_stage}
\vspace*{-0.5em}

We present a case in the left of \cref{fig:ssba_loss_landscape}, 
where the model is attacked by SSBA \cite{ISSBA}, 
and the defender simply trains a Bézier curve connecting the initial backdoored model $\bm{\theta}_{adv}$ and its slightly fine-tuned version, $\bm{\theta}_{ft}$ \cite{MCR}. 
As shown, if the poisoned model $\bm{\theta}_{adv}$ and $\bm{\theta}_{ft}$ lie in the same loss basin, eliminating the backdoor through model connection or fusion is difficult, 
as the adversarial loss along the curve remains low.
To amplify the adversarial loss, 
we utilize the permutation invariance property to project $\bm{\theta}_{adv}$ to a distinct loss basin to obtain the other endpoint, $\bm{\theta}_{adv'}$, rather than fine-tuning $\bm{\theta}_{adv}$. 
This can be achieved by finding a set of permutation matrices $S(\bm{P}')$ that maximize the cost function $c_l$ for each layer. 
In contrast to neuron alignment in problem (\ref{basic_na_minimization}), 
our goal is to find the permutation $\bm{P}'_l$ for all layers to project the backdoored model into a distinct loss basin. 
Specifically, we formulate the following optimization problem to obtain $S(\bm{P}')$:
\vspace*{-0.5em}
\begin{equation}
    \label{l2_maximization}
    \bm{P}'_l =  \mathop{\arg \max}\limits_{\bm{P}_l\in \Pi_{d_l}} \sum_i^n   \left\lVert \bm{x}_{i,\; l}^{adv} - \ \bm{P}_l \; \bm{x}_{i,\; l}^{adv} \right\rVert^2.
    \vspace*{-0.2em}
\end{equation}
Then, we get the updated model $\bm{\theta}_{adv'} = \pi(\bm{\theta}_{adv}, S(\bm{P}'))$. 
It's important to note that this procedure does not alter the output of the backdoored model for any input; consequently, it does not increase the loss for any sample, though here we are dealing with a maximization problem.
To amplify the adversarial loss, we then train a Bézier curve, $\bm{\gamma}_1$, connecting $\bm{\theta}_{adv}$ and $\bm{\theta}_{adv'}$, and subsequently select a model corresponding to a middle point on this curve.

As shown in the middle of \cref{fig:ssba_loss_landscape}, 
when connecting $\bm{\theta}_{adv}$ and $\bm{\theta}_{adv'}$, 
the loss of poisoned samples increases significantly along the Bézier curve $\bm{\gamma}_1$ (\textit{i.e.,} circumventing the backdoor space). 
As expected, the loss landscape becomes symmetric for both poisoned and benign samples. 
However, compared to MCR, we observe an increase in the loss of benign samples, 
implying that the models selected from the curve connecting $\bm{\theta}_{adv}$ and $\bm{\theta}_{adv'}$ would perform poorly on the initial task.

\vspace*{-0.5em}
\subsection{Clean Accuracy Recovery}
\vspace*{-0.3em}
To ensure that the purified model maintains better performance on benign data, 
we re-align the model $\bm{\theta}_{t}$ found on $\bm{\gamma}_1$ to the benign loss basin of the backdoored model $\bm{\theta}_{adv}$. 
We denote the corresponding permutation set as $\bm{P}^*$. 
Furthermore, empirically, when setting $t=0.4$, the model $\bm{\theta}_{t}$ typically exhibits higher loss on both poisoned and benign samples. 
We then train another curve $\bm{\gamma}_2$ connecting $\bm{\theta}_{adv}$ and $\bm{\theta}_{t^*=0.4} = \pi(\bm{\theta}_{t=0.4}, S(\bm{P}^*))$. 
The right side of \cref{fig:ssba_loss_landscape} shows that the loss of poisoned samples gradually increases along the curve from $\bm{\theta}_{adv}$ to $\bm{\theta}_{t^*=0.4}$, 
and the model point along this curve could also attain a lower loss on benign samples than $\bm{\theta}_{t=0.4}$. 
A model along the curve with a high loss for poisoned samples and a low loss for benign samples can be selected as the ideal purified model.

\begin{figure*}[t]
    \centering
    \includegraphics[width=0.99\linewidth]{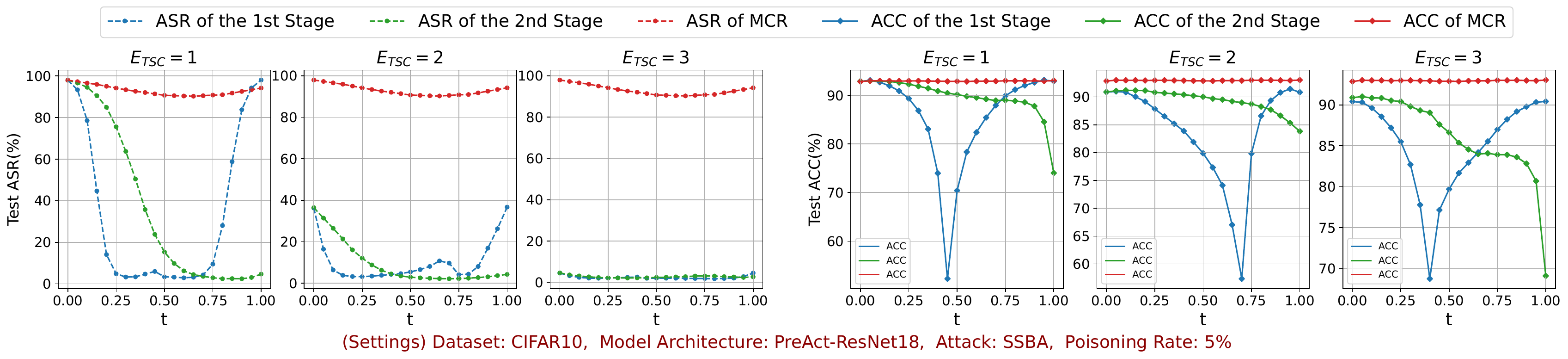}
    \vspace*{-0.6em} 
    \caption{
        Test attack success rate (ASR) and accuracy (ACC) on benign samples are evaluated as functions of the points along the Bézier curve found by MCR and \shortname. 
        We assess the performance against SSBA on CIFAR-10 with 5\% poison rate using PreAct-ResNet18. 
        We select model points along the curve at $t=0.4$ for each stage and round.
        Since MCR only trains a single curve for model purification, we plot the results of MCR at each round of \shortname\ for better comparison.
    }
       \label{fig:curve_acc}
    \vspace*{-0.15em} 
\end{figure*}

Additionally, as shown in \cref{fig:curve_acc}, we find that performing one round of \shortname\ is insufficient to remove the backdoor. 
In practice, we slightly fine-tune the model obtained from the second stage of \shortname, 
then use this fine-tuned model as the input for the next round to mitigate the backdoor threat step by step. 
It is evident that as the global epoch $E_{\shortname}$ increases, 
the ASR of model points along the second-stage curve of \shortname\ decreases. 
Empirically, setting $E_{\shortname} = 3$ can effectively eliminate the backdoor while maintaining good performance on the benign task.

We give the pseudocode in \cref{alg:tsc}. 
Since we assume the defender only has access to a small fraction of clean samples $D_c$, 
both the computation of the permutation and the training of the curve are conducted over $D_c$.
The function $\textsc{PermuteLayers}(\bm{\theta}_A, \bm{\theta}_B, D, \textsc{opt})$ in the pseudocode 
returns a model by permuting the layers of $\bm{\theta}_B$, aligning or un-aligning it with $\bm{\theta}_A$. 
The function $\textsc{FitQuadCurve}(\bm{\theta}_A, \bm{\theta}_B, \mathcal{F}, D, e)$ returns the quadratic Bézier curve connecting $\bm{\theta}_A$ and $\bm{\theta}_B$, trained using method $\mathcal{F}$ over $D$ for $e$ epochs.
Moreover, $\mathcal{F}$ can be various training methods tailored to the corresponding data format, as shown in \cref{fig:overview}. 
The function $\textsc{RetrievePoint}(\bm{\gamma}_{A,B}, t)$ returns the model point along $\bm{\gamma}_{A,B}$ at index $t$, as described in \cref{bezier_curve}. 
See more details in \cref{practical_algorithm}.
\vspace*{-0.3em}

\vspace*{-0.5em}
\subsection{Theoretical Analysis of TSC}
\vspace*{-0.5em}
\input{algorithms/tsc.tex}

We consider two feedforward neural networks, as defined in \cref{feedforward_network}, with weights \( \bm{\theta}_0 \) and \( \bm{\theta}_1 \).
We say that \( \bm{\theta}_0 \) and \( \bm{\theta}_1 \) are \textit{$L_2$-norm-consisten} if they satisfy the following condition:
\begin{definition}[Weight $L_2$-norm consistency condition]
    \label{weight_consistency}
    Let \( \bm{W}_l^{0} \) and \( \bm{W}_l^{1} \) be the parameters of the \( l \)\textsuperscript{th} layer of \( \bm{\theta}_0 \) and \( \bm{\theta}_1 \), respectively. 
    We define that the parameters satisfy the condition that the $L_2$ norms of the corresponding layer weights are equal:
    $\|\bm{W}_l^{0}\|_2 = \|\bm{W}_l^{1}\|_2, \forall l \in \{ 1, 2, ..., L \}$,
    where the $L_2$ norm of the matrix refers to the element-wise $L_2$ norm (\textit{i.e.,} Frobenius norm).
\end{definition}
\vspace*{-0.4em}
This condition implies that the magnitude of the weights in each layer is preserved between the two models. 
Consequently, both models effectively operate at the same scale, 
which can be critical for their comparative performance.

Next, considering three optimal independent feedforward networks with weights \( \bm{\theta}_A \), \( \bm{\theta}_B \) and  \( \bm{\theta}_C\),
we can present a theorem regarding the upper bounds of the loss $\ell(\cdot)$ in \cref{curve_loss} over the quadratic curves $\bm{\gamma}_{\bm{\theta}_{A,B}}(t)$ and $\bm{\gamma}_{\bm{\theta}_{A,C}}(t)$ across dataset $D$.
Before stating the theorem, we first reformulate the parametric Bézier curve \(\bm{\gamma}_{\bm{\theta}_{A,B}}(t)\) by replacing \(\bm{\theta}_{A,B}\) with its deviation \(\tilde{\bm{\theta}}_{A,B}\):
\vspace*{-0.4em}
\begin{equation}
    \label{bezier_curve_deviation_brief}
    \bm{\gamma}_{\tilde{\bm{\theta}}_{A,B}}(t) = (1 - t)\bm{\theta}_{A} + t\bm{\theta}_{B} + 2t(1 - t)\tilde{\bm{\theta}}_{A,B}. 
\end{equation}
Similarly, we can reformulate \(\bm{\gamma}_{\bm{\theta}_{A,C}}(t)\) to \(\bm{\gamma}_{\tilde{\bm{\theta}}_{A,C}}(t)\).
Such reformulation allows us to express the Bézier curves in a considerably simpler form, thus facilitating theoretical analysis.
The detailed derivation is provided in \cref{bezier_curve_deviation}.
Moreover, we refer to the quantity \(M_l(\bm{\theta}_A, \bm{\theta}_B; D)\), as defined in \cref{l2_minimization}, which measures the $L_2$ norm of the feature distance at the $l$\textsuperscript{th} layer between models $\bm{\theta}_A$ and $\bm{\theta}_B$ for a given dataset $D$.
We can then derive the following theorem:

\begin{theorem}
    \label{general_theorem}
    We assume that the activation function \(\sigma\) in \cref{feedforward_network} and the loss function \(\mathcal{L}\) in \cref{curve_loss} are Lipschitz continuous. 
    Let \(\bm{\gamma}_{\tilde{\bm{\theta}}_{A,B}}(t)\) and \(\bm{\gamma}_{\tilde{\bm{\theta}}_{A,C}}(t)\) be two Bézier curves defined in \cref{bezier_curve_deviation_brief}. 
    Then, under the following conditions:
    
    \noindent(1) \(M_l(\bm{\theta}_A, \bm{\theta}_B; D) \leq M_l(\bm{\theta}_A, \bm{\theta}_C; D)\), \(\forall l \in \{1, 2, \ldots, L\}\);
    \noindent(2) \( \bm{\theta}_B\) and \(\bm{\theta}_C\) are \(L_2\)-norm-consistent;
    \noindent(3) \( \tilde{\bm{\theta}}_{A,B}\) and \(\tilde{\bm{\theta}}_{A,C}\) are \(L_2\)-norm-consistent;    
    \noindent there exists upper bounds $U_{{A,B}}$ and $U_{{A,C}}$ for $\ell(\cdot)$ such that:
    $\ell(\tilde{\bm{\theta}}_{A,B}) \le U_{{A,B}}, \ \ell(\tilde{\bm{\theta}}_{A,C}) \le U_{{A,C}}$, where $U_{{A,B}} \le U_{{A,C}}$.
\end{theorem}
\vspace*{-1em}
\begin{proof}
    See \Cref{proof_for_general_theorem}.
\end{proof}
\vspace*{-1em}
A similar theorem can be found in \cite{optMC_na}.
It is important to note that the theorem demonstrated in \cite{optMC_na} provides only the upper bound relations between linear paths connecting aligned and unaligned models, 
despite \citet{optMC_na} claiming that their theorem pertains to quadratic Bézier curves. 
Moreover, their theorem represents only the left inequality of \Cref{permutation_lemma}, 
which is a special case of \Cref{general_theorem} and is insufficient for theoretical analysis of \shortname.

\paragraph{Increasing the Adversarial Loss.}
Now, 
we redirect our focus to our method
and compare the scenarios where one endpoint is fixed by model $\bm{\theta}_A$ 
while the other is settled by $\bm{\theta}_B$, $\bm{\theta}_{\hat{B}}=\pi(\bm{\theta}_B, S(\bm{\hat{P}}))$ or $\bm{\theta}_{B'}=\pi(\bm{\theta}_B, S(\bm{P}'))$.
$S(\bm{\hat{P}})$ and $S(\bm{P}')$ correspond to the solutions in problems (\ref{l2_minimization}) and (\ref{l2_maximization}), respectively.
Then, we have the following corollary:
\begin{corollary}
    \label{permutation_lemma}
    We assume that the activation function \(\sigma\) and the loss function \(\mathcal{L}\) are Lipschitz continuous.
    Let \(\bm{\gamma}_{\tilde{\bm{\theta}}_{A,B}}(t)\), \(\bm{\gamma}_{\tilde{\bm{\theta}}_{A,\hat{B}}}(t)\) and \(\bm{\gamma}_{\tilde{\bm{\theta}}_{A,{B'}}}(t)\) be three Bézier curves defined in \cref{bezier_curve_deviation_brief}. 
    We also assume that \( \tilde{\bm{\theta}}_{A,B}\), \(\tilde{\bm{\theta}}_{A,\hat{B}}\) and \(\tilde{\bm{\theta}}_{A,B'}\) are \(L_2\)-norm-consistent with each other.
    Then, there exists upper bounds $U_{{A,B}}$, $U_{{A, \hat{B}}}$ and $U_{{A, B'}}$ for $\ell(\cdot)$ such that:
    $U_{{A,\hat{B}}} \le U_{{A,B}} \le U_{{A,B'}}.$
\end{corollary}
\vspace*{-1em}
\begin{proof}
    See \Cref{proof_for_specific_proof}.
\end{proof}
\vspace*{-1em}
Intuitively, 
to increase the loss with respect to poison samples along the curve,
we can project the copy of backdoored model $\bm{\theta}_{adv}$ to loss basin different from that of the original.
Theoretically, \Cref{permutation_lemma} 
implies that applying the permutation by $S(\bm{P}')$ can enlarge the upper bound of the backdoor loss over the curve,
resulting in a more robust model on the curve when faced with attack samples.

However, 
\Cref{permutation_lemma} also indicates that permuting the layers via $S(\bm{P}')$, 
leads to a looser upper bound for the loss on benign samples compared to models that reside in the same loss basin. 
Therefore, an extended method is needed to ensure that the purified model maintains strong performance on clean samples.
\vspace*{-1.2em}

\paragraph{Reducing the Clean Loss.}
In the second stage,
we train another curve connecting the origianl backdoored model and re-aligned model found previously.
As demonstrated in \cref{permutation_lemma}, re-aligning the model found
on the curve to the original loss basin of the backdoored model 
can reduce the upper bound of loss value along the curve.
When aligning the model with benign samples, 
the curve is more likey to be trained in a loss basin of benign data rather than poisoned ones.
Thus, the second stage successfully improves the performance on benign task.

Moreover, let $\bm{\theta}_{adv*}$ be a model satisfies the following three conditions: 
(1) it is aligned with $\bm{\theta}_{adv}$;
(2) it is \(L_2\)-norm-consistent with the model $\bm{\theta}_{t^*=0.4}$;
and (3) it achieves higher accuracy on the poisoned dataset $D_{adv}$ than $\bm{\theta}_{t^*=0.4}$. 
Also, consider a curve $\bm{\gamma}_{adv}$ 
that connects $\bm{\theta}_{adv*}$ and $\bm{\theta}_{adv}$. 
We can say that the 2-Wasserstein distance $W_2(\mathbb{P}_{l}^{\ adv}, \mathbb{P}_{l}^{\ adv*}; D_{adv})$ 
is smaller than $W_2(\mathbb{P}_{l}^{\ adv}, \mathbb{P}_{l}^{\ t=0.4}; D_{adv})$ for each layer, 
as $\bm{\theta}_{adv}$ is more functionally identical to $\bm{\theta}_{adv}$ over $D_{adv}$. 
Thus, we have $M_l(\bm{\theta}_{adv}, \bm{\theta}_{adv*}; D_{adv}) \le M_l(\bm{\theta}_{adv}, \bm{\theta}_{t^*=0.4}; D_{adv})$. 
According to \Cref{general_theorem}, 
if the parameters of $\bm{\gamma}_{adv}$ and $\bm{\gamma}_{2}$ 
are \(L_2\)-norm-consistent, 
we conclude that the upper bound of $\bm{\gamma}_{2}$ is higher than that of $\bm{\gamma}_{adv}$ over the poisoned dataset.
This finding implies that a high loss can be maintained for poisoned samples along the curve $\bm{\gamma}_{2}$.

\cref{fig:curve_acc} compares model performance along curves identified by MCR and \shortname. 
For each epoch of \shortname, 
as $t$ approaches 0.5, 
both ACC and ASR of model points from the first stage decrease more rapidly than those of MCR and \shortname's second stage. 
Besides, model points from the second stage exhibit significantly lower ASR than MCR while maintaining moderate ACC decline compared to the first stage. 
These observations support our theoretical analysis.

%% file: algorithms/tsc.tex
\begin{algorithm}[thb]
    \caption{\fullname}
    \label{alg:tsc}
 \begin{algorithmic}
    \REQUIRE{backdoored model $\bm{\theta}_{adv}$, clean dataset $D_c$, global epoch $E_{\shortname}$,
    curve index $t$, curve training epoch $e$, training method $\mathcal{F}$;
    }
    \ENSURE{purified model $\bm{\theta}_{p}$;
    }
    \STATE Initialize \; $\bm{\theta}_{p} \gets \bm{\theta}_{adv}$;
    \FOR{$i=1$ {\bfseries to} $E_{\shortname}$}
    \STATE $\vartriangleright $ \; Adversarial Loss Amplification
    \STATE $\bm{\theta}_{p'} \gets \textsc{PermuteLayers}(\bm{\theta}_{p}, \bm{\theta}_{p}, D_c, \textsc{max})$;
    \STATE $\bm{\gamma}_{\bm{\theta}_{p, p'}} \gets \textsc{FitQuadCurve}(\bm{\theta}_{p}, \bm{\theta}_{p'}, \mathcal{F}, D_c, e)$;
    \STATE $\bm{\theta}_{t} \gets \textsc{RetrievePoint}(\bm{\gamma}_{\bm{\theta}_{p, p'}} \ ,\ t)$;
    \STATE $\vartriangleright $ \; Clean Accuracy Recovery
    \STATE $\bm{\theta}_{t^*} \gets \textsc{PermuteLayers}(\bm{\theta}_{p}, \bm{\theta}_{t}, D_c, \textsc{min})$;
    \STATE $\bm{\gamma}_{\bm{\theta}_{p, t^*}} \gets \textsc{FitQuadCurve}(\bm{\theta}_{p}, \bm{\theta}_{t^*}, \mathcal{F}, D_c, e)$;
    \STATE $\bm{\theta}_{p} \gets \textsc{RetrievePoint}(\bm{\gamma}_{\bm{\theta}_{p, t^*}} \ ,\ t)$;
    \ENDFOR 
 \end{algorithmic}
\end{algorithm}

%% file: sections/4_experiments.tex
\section{Experiments}
\vspace*{-0.2em}
\subsection{Experimental Settings}
\vspace*{-0.1em}
\paragraph*{Attack Setup.}
(1) \textbf{Supervised Learning.}
We consider eleven typical backdoor attacks, including eight label-flipping attacks (BadNet \cite{badnets}, Blended \cite{TargetGlasses}, SSBA \cite{ISSBA}, LF \cite{lf}, WaNet \cite{wanet}, Inputaware \cite{Inputaware}, SBL \cite{sbl} and SAPA \cite{sapa}) and three clean label attacks (LC \cite{firstCleanLabel}, SIG \cite{SIG} and Narcissus \cite{narcissus}). 
These attacks are conducted on CIFAR10 \cite{CIFAR10} using PreAct-ResNet18 \cite{Pre-Act-ResNet} and ImageNet100 \cite{ImageNet} using ResNet50 \cite{ResNet} with various poisoning rates. 
(2) \textbf{Self-Supervised Learning.}
We evaluate two self-supervised learning attacks: BadEncoder \cite{BadEncoder} and CTRL \cite{embarrassing}.
BadEncoder is conducted using two typical SSL training methods, SimCLR \cite{simCLR} and CLIP \cite{clip}.
For SimCLR, we utilize publicly available backdoored ResNet18 and ResNet50 encoders on CIFAR10 and ImageNet, respectively, evaluating ACC and ASR through linear probing \cite{linear_probe} on downstream datasets STL10 \cite{STL10}, GTSRB, and SVHN \cite{SVHN}. 
For CLIP, following \cite{BadEncoder}, 
we fine-tune a pre-trained CLIP ResNet50\footnote{https://github.com/openai/CLIP} on ImageNet100 using SimCLR to inject backdoors, 
and assess performance through linear probing and zero-shot evaluation on STL10, Food101 \cite{food101}, and VOC2007 \cite{pascal-voc-2007}. 
The CTRL attack is evaluated using the same SimCLR settings. 
Implementation details are provided in \Cref{appendix:Experimental_Detials}.

\input{tab/cifar10_exp.tex}

\vspace*{-1em}
\paragraph*{Defense Setup.}
In our experiments, we focus on \post\ methods and provide all defenses with 5\% of the clean training dataset, except for the defenses for the CLIP model. 
To address the backdoored CLIP visual model, 
we employ the entire MS-COCO dataset \footnote{As the dataset used to train CLIP \cite{clip} is not publicly available and involves 400M images, 
we conduct defenses with the much smaller MS-COCO-2017 dataset \cite{MSCOCO}, 
which contains approximately 120K images with 5 captions each.} \cite{MSCOCO}.
(1) \textbf{Supervised Learning.} We evaluate six \post\ methods as baselines: FP \cite{FinePruning}, NC \cite{NeuralCleanse}, MCR \cite{MCR}, ANP \cite{ANP}, FT-SAM \cite{FT_SAM}, I-BAU \cite{I_BAU}, and SAU \cite{sau}. 
For \shortname, we set the global epoch $E_{\shortname}=3$, curve index $t=0.4$, and curve training epoch $e=200$.
(2) \textbf{Self-Supervised Learning.} We consider MCR \cite{MCR} and SSL-Cleanse \cite{SSL_Cleanse} as baselines. 
For \shortname, we set the global epoch $E_{\shortname}=2$, curve index $t=0.25$, and curve training epoch $e=200$ for both SimCLR and CLIP. 
Comprehensive settings for all defenses are provided in \Cref{appendix:Experimental_Detials}.
For the following experiments, 
we consider a defense against an attack \colorbox{cleanModel}{successful} if the ASR is reduced to below 15\%.

\vspace*{-0.5em}
\subsection{Results for Supervised Learning}
\vspace*{-0.2em}
\cref{tab:cifar10_imagenet} 
compares the performance of \shortname\ with existing defenses on CIFAR10 and ImageNet100 under supervised learning scenarios.
Comprehensive results for GTSRB and of other attack settings are provided in \Cref{additional_sl_results}.

It's clear that \shortname\ successfully reduces the ASR to below 15\% for all attacks on CIFAR10 GTSRB and ImageNet100, demonstrating its robustness and effectiveness.
Among all attacks, Blended attack with small poisoning rates proves most challenging to defend against. 
While other defenses struggle to contain ASRs below 25\%, 
\shortname\ reduces the ASR to 12.46\% on CIFAR10 (1\% poisoning rate) and 12.63\% on ImageNet100 (0.5\% poisoning rate).
SAU shows strong backdoor removal capabilities against most attacks but suffers from ACC instability.
For example, under BadNet Attack with 1\% poisoning rate on CIFAR10,
SAU's accuracy falls below 70\%.
We attribute this behavior to SAU's aggressive unlearning strategy,
which leads to catastrophic decreases in both ACC and ASR.
ANP, FT-SAM, and I-BAU also demonstrate effectiveness, 
reducing ASRs to below 15\% for most attacks on CIFAR10 and GTSRB.
However, their performance diminishes on ImageNet100, particularly against Blended, Inputaware and LF attacks.
While \shortname\ occasionally yields lower initial task accuracy than FT-SAM, 
it outperforms ANP and I-BAU in most cases.

To further explore the impact of \shortname\ on ACC, we provide results for ACC drops on non-backdoored models in \cref{appendix:clean_acc_drop}.
These results indicate that the ACC drop for \shortname\ is acceptable in scenarios without data poisoning.

\subsection{Results for Self-supervised Learning}
\vspace*{-0.2em}
\cref{tab:simclr_defense,tab:clip_defense} show the results against BadEncoder attacks under SimCLR and CLIP training scenarios, respectively.
We employ different settings for SimCLR and CLIP to showcase the flexibility of \shortname.
For SimCLR, encoders are trained and backdoored independently with specific target labels for each downstream task. 
For CLIP, we backdoor a single visual encoder using `truck' as the target label/caption 
and evaluate performance before and after defenses across downstream tasks. 
Since Food101 and VOC2007 lack the `truck' label,
 we augment their training sets with truck images from STL10 for evaluation of linear probing. 
 Results indicate the attack remains effective even without targeting specific downstream datasets.
More results for CTRL attack and SSL-Cleanse are provided in \Cref{additional_ssl_results}. 

\input{tab/simCLR.tex}

\input{tab/CLIP.tex}

Under SimCLR, while MCR successfully removes backdoors targeting STL10 and GTSRB downstream tasks, 
it fails against attacks targeting the SVHN dataset. 
In contrast, \shortname\ effectively reduces the ASR to below 11\% across all downstream tasks.
For CLIP, \shortname\ achieves remarkable performance, reducing ASR below 2\%, 
while MCR proves ineffective against BadEncoder attacks.

\subsection{Resistance to Potential Adaptive Attacks}
\vspace*{-0.6em}
The previous experiments demonstrate the effectiveness of \shortname\ against existing backdoor attacks. 
However, it is essential to consider adpative attacks against \shortname.
The core defense mechanism of \shortname\ relies on increasing adversarial loss along the quadratic Bézier curve by projecting model $\bm{\theta}_{adv}$ 
to a distinct loss basin to find $\bm{\theta}_{adv'}$ during the first stage. 
An adaptive attack would attempt to train and backdoor a model that maintains low backdoor loss along the curve identified by \shortname.

To design such an adaptive attack, 
we build upon the neural network subspace learning approach proposed in \cite{learning_subspace}, 
originally developed for improving accuracy and calibration.
Our strategy involves simultaneously training a curve and updating its endpoints $\bm{\theta}_{adv}$ and $\bm{\theta}_{adv'}$ 
using the mixture of benign and poisoned data (\textit{i.e.}, learning a backdoored subspace).
After training, we select one endpoint as the final model.
We convert all previously evaluated attacks into adaptive versions under both supervised and self-supervised learning settings.
Experimental results reveal that \shortname\ remains robust against such attack.
The implementation details and results are provided in \Cref{adaptive_attack}.

To further validate the robustness of \shortname\ against adaptive attacks, we present corresponding loss landscape visualizations in \cref{adap_loss_landscape}.
The analysis reveals that the combination of the permutation mechanism and training exclusively with benign samples contributes to amplifying the loss on poisoned samples, even for adaptive attacks.
Moreover, given that our defense involves the concept of loss landscapes, we also conduct experiments against more advanced attacks, including SBL \cite{sbl}, Narcissus \cite{narcissus}, and SAPA \cite{sapa}.
These modern backdoor attacks aim to optimize flatter loss landscapes or entangle benign and backdoor features.
The results demonstrate that \shortname\ effectively reduces the ASR to below 15\% even against these sophisticated attacks.

\vspace*{-0.5em}
\subsection{Ablation Studies}
We conduct comprehensive ablation studies to analyze the impact of \shortname's key hyperparameters: the number of global epochs $E_{\shortname}$ and the curve index $t$. 
Through extensive experiments, 
we find that increasing $t$ (up to 0.5) and $E_{\shortname}$ improves backdoor removal performance 
but meanwhile reducing accuracy on benign samples.
Moreover, a larger $E_{\shortname}$ leads to more stable performance.
Based on empirical results, we recommend $t=0.4$ and $E_{\shortname}=3$ for supervised learning scenarios, 
and $t=0.25$ and $E_{\shortname}=2$ for self-supervised learning. 
Detailed analysis and additional experimental results can be found in \cref{ablation_hyperparameters}.

To validate the stability of \shortname, 
we conduct experiments with VGG19-BN \cite{vgg19} and InceptionV3 \cite{inceptionv3} on CIFAR10. 
Results in \cref{ablation_model_results} show the robustness of \shortname\ across different model architectures.

We opt to use the same $t$ for both stages to maintain a simpler parameter design.
Employing distinct $t$ values for each stage would lead to numerous parameter combinations, potentially complicating the algorithm's overall structure.
Moreover, as shown in \cref{fig:curve_acc}, the ACC/ASR values in the first stage exhibit a roughly symmetric pattern with respect to $t$, whereas in the second stage, they decrease as $t$ increases. 
Although the overall trends for the two stages differ across $t \in [0,1]$, they both demonstrate a decreasing tendency within $t \in [0,0.5]$.
Notably, in Stage 2, ASR decreases effectively while ACC remains high for $t$ values near $0.5$.
Considering this observation from Stage 2, the symmetry in Stage 1, and the consistent trend within $t \in [0,0.5]$, we suggest selecting $t$ from this range for both stages.

\section{Conclusion and Limitation}
In this paper, we propose \shortname, 
a novel defense mechanism leveraging permutation invariance. 
Unlike previous post-purification defenses, 
\shortname\ utilizes weight symmetry to remove backdoors 
and is applicable to both supervised and self-supervised learning scenarios, 
with potential extensions to other learning paradigms. 
Our experiments demonstrate the robustness of \shortname\ under diverse attack settings,
achieving comparable or superior performance to existing defenses. 
However, \shortname\ occasionally trades off accuracy on benign samples for backdoor removal. 
Future work could focus on optimizing such trade-off to maintain high ASR reduction while improving ACC.

%% file: tab/cifar10_exp.tex
\begin{table*}[th]
    \vspace*{-1em}
    \caption{
        Results on CIFAR10 and ImageNet100 under supervised learning scenarios. 
        Attack Success Rates (ASRs) below 15\% are highlighted in blue to indicate a \colorbox{cleanModel}{successful} defense, 
        while ASRs above 15\% are denoted in red as \colorbox{poisonedModel}{failed} defenses.} 
    \centering 
    \renewcommand{\arraystretch}{1.14}
    \setlength\tabcolsep{6pt}
    \resizebox{0.99\linewidth}{!}{
        \setlength\arrayrulewidth{1pt}
        \begin{tabular}{ccc *{9}{cc}}
        \hline
        \multirow{2}*{} &
        \multirow{2}*{Attacks} & \multirow{2}*{\shortstack{Poison\\Rate}} &
        \multicolumn{2}{c}{No Defense} & 
        \multicolumn{2}{c}{FP} &
        \multicolumn{2}{c}{NC} &
        \multicolumn{2}{c}{MCR} &
        \multicolumn{2}{c}{ANP} &
        \multicolumn{2}{c}{FT-SAM} &
        \multicolumn{2}{c}{I-BAU} &
        \multicolumn{2}{c}{SAU} &
        \multicolumn{2}{c}{TSC (ours)} \\[2pt]
        \cmidrule(r){4-5} \cmidrule(lr){6-7} \cmidrule(lr){8-9} \cmidrule(lr){10-11} 
        \cmidrule(lr){12-13} \cmidrule(lr){14-15} \cmidrule(lr){16-17} \cmidrule(lr){18-19} \cmidrule(l){20-21}
        & & & ACC($\uparrow$) & ASR($\downarrow$) & ACC($\uparrow$) & ASR($\downarrow$) & 
        ACC($\uparrow$) & ASR($\downarrow$) & ACC($\uparrow$) & ASR($\downarrow$) & 
        ACC($\uparrow$) & ASR($\downarrow$) & ACC($\uparrow$) & ASR($\downarrow$) & 
        ACC($\uparrow$) & ASR($\downarrow$) & ACC($\uparrow$) & ASR($\downarrow$) & ACC($\uparrow$) & ASR($\downarrow$) \\[2pt]
        \hline
        \multicolumn{1}{c|}{\multirow{16}*{\rotatebox{90}{CIFAR10}}}
        &\multirow{2}*{\shortstack{BadNet}}	&5\%	&92.64	&\light 88.74	&92.26	&\grc 1.17	&90.53	&\grc 1.01	&92.17	&\grc 7.62	&86.45	&\grc 0.02	&92.19	&\grc 3.50	&88.66	&\grc 0.92	&89.32	&\grc 1.74	&89.19	&\grc 1.90	\\
        \multicolumn{1}{c|}{}&&1\%	&93.14	&\light 74.73	&92.59	&\grc 2.29	&92.07	&\grc 0.77	&92.90	&\light 18.06	&85.82	&\grc 0.04	&92.39	&\grc 1.57	&87.80	&\grc 2.29	&65.38	&\grc 2.06	&90.71	&\grc 1.26	\\ \cline{2-21}
        \multicolumn{1}{c|}{}&\multirow{2}*{\shortstack{Blended}}	&5\%	&93.66	&\light 99.61	&92.70	&\light 49.47	&93.67	&\light 99.61	&93.23	&\light 99.01	&88.95	&\light 18.76	&93.00	&\light 29.59	&88.07	&\light 34.86	&90.69	&\grc 7.74	&90.14	&\grc 10.53	\\
        \multicolumn{1}{c|}{}&&1\%	&93.76	&\light 94.88	&92.92	&\light 69.74	&93.76	&\light 94.88	&93.62	&\light 93.10	&89.69	&\light 60.52	&93.00	&\light 49.36	&89.62	&\light 25.74	&90.02	&\light 36.16	&91.12	&\grc 12.46	\\ \cline{2-21}
    \multicolumn{1}{c|}{}&\multirow{2}*{\shortstack{LF}}	&5\%	&93.36	&\light 98.03	&92.84	&\light 59.12	&90.98	&\grc 2.43	&93.07	&\light 97.32	&84.20	&\grc 2.46	&92.89	&\grc 7.44	&88.64	&\light 45.66	&90.60	&\grc 1.71	&88.50	&\grc 3.78	\\
    \multicolumn{1}{c|}{} &&1\%	&93.56	&\light 86.44	&92.45	&\light 65.80	&93.56	&\light 86.46	&93.09	&\light 84.11	&86.27	&\grc 11.28	&93.47	&\grc 11.71	&90.53	&\light 69.28	&91.58	&\light 18.12	&90.68	&\grc 11.67	\\ \cline{2-21}
    \multicolumn{1}{c|}{} &\multirow{2}*{\shortstack{SSBA}}	&5\%	&93.27	&\light 94.91	&92.55	&\light 16.27	&93.27	&\light 94.91	&92.94	&\light 92.06	&88.72	&\grc 0.13	&92.71	&\grc 2.87	&89.65	&\grc 1.54	&91.30	&\grc 2.06	&89.43	&\grc 2.18	\\
    \multicolumn{1}{c|}{} &&1\%	&93.43	&\light 73.44	&93.01	&\grc 7.68	&91.60	&\grc 0.46	&93.33	&\light 65.88	&85.33	&\grc 0.31	&93.02	&\grc 1.49	&89.56	&\grc 4.87	&91.38	&\grc 0.99	&91.18	&\grc 2.18	\\ \cline{2-21}
    \multicolumn{1}{c|}{}&\multirow{2}*{\shortstack{SBL-BadNet}}	&5\%	&90.79	&\light 93.48	&92.59	&\grc 1.13	&92.22	&\grc 0.59	&92.26	&\light 91.82	&82.82	&\light 51.63	&92.16	&\light 60.03	&90.67	&\light 27.06	&91.31	&\grc 0.60	&91.02	&\grc 1.12	\\
	\multicolumn{1}{c|}{}&&1\%	&91.71	&\light 88.64	&93.10	&\light 31.77	&91.82	&\grc 0.72	&93.23	&\light 86.11	&82.71	&\light 81.48	&92.77	&\light 59.58	&90.63	&\grc 2.00	&92.32	&\grc 1.01	&91.54	&\grc 1.93	\\ \cline{2-21}
    \multicolumn{1}{c|}{}&\multirow{2}*{\shortstack{SAPA}}	&1\%	&94.01	&\light 99.97	&92.34	&\light 92.22	&92.80	&\grc 2.14	&93.83	&\light 100.00	&86.06	&\light 92.68	&93.06	&\light 79.80	&86.69	&\light 15.17	&91.83	&\grc 1.96	&90.37	&\grc 7.41	\\
	\multicolumn{1}{c|}{}&&0.5\%	&93.77	&\light 84.80	&88.82	&\light 82.76	&92.74	&\grc 1.52	&93.78	&\light 80.83	&87.99	&\light 81.52	&93.23	&\light 82.02	&90.16	&\light 26.48	&91.75	&\grc 0.68	&90.98	&\grc 7.32	\\ \cline{2-21}
    \multicolumn{1}{c|}{}&\multirow{2}*{\shortstack{LC}}	&5\%	&93.31	&\light 98.33	&92.19	&\light 72.99	&92.32	&\grc 0.64	&92.94	&\light 99.94	&88.15	&\grc 13.83	&92.59	&\light 57.18	&90.15	&\grc 1.99	&91.53	&\grc 1.50	&90.04	&\grc 2.38	\\
    \multicolumn{1}{c|}{}   &&1\%	&93.79	&\light 75.93	&92.86	&\light 29.86	&92.31	&\grc 0.68	&93.67	&\light 82.54	&86.58	&\light 31.46	&92.83	&\light 39.40	&89.78	&\grc 0.71	&92.16	&\grc 3.77	&90.08	&\grc 5.78	\\ \cline{2-21}
    \multicolumn{1}{c|}{}&\multirow{2}*{\shortstack{Narcissus}}	&1\%	&93.68	&\light 82.87	&92.29	&\light 44.88	&93.68	&\light 47.87	&93.61	&\light 49.79	&92.01	&\light 27.01	&93.05	&\light 26.80	&90.21	&\light 18.67	&91.36	&\grc 3.24	&90.65	&\grc 7.88	\\
	\multicolumn{1}{c|}{}&&0.5\%	&93.68	&\light 80.58	&92.94	&\light 29.59	&93.67	&\light 32.57	&93.69	&\light 32.96	&89.35	&\light 16.78	&93.06	&\grc 14.08	&89.16	&\light 21.09	&91.74	&\grc 5.81	&91.71	&\grc 8.02	\\ 

\hline
\multicolumn{1}{c|}{\multirow{16}*{\rotatebox{90}{ImageNet100}}}
 &\multirow{2}*{\shortstack{BadNet}}	&0.5\%	&84.30	&\light 99.78	&83.36	&\grc 9.80	&81.92	&\grc 0.52	&85.24	&\light 99.66	&78.44	&\light 94.18	&83.70	&\grc 9.45	&73.70	&\grc 8.34	&73.86	&\grc 0.28	&80.20	&\grc 0.22	\\
\multicolumn{1}{c|}{}	&&1\%	&84.56	&\light 99.86	&83.10	&\grc 9.58	&81.92	&\grc 0.49	&85.08	&\light 99.86	&79.48	&\light 93.64	&83.88	&\light 24.14	&71.46	&\light 43.66	&72.84	&\grc 0.26	&78.06	&\grc 0.14	\\ \cline{2-21}
\multicolumn{1}{c|}{}&\multirow{2}*{\shortstack{Blended}}	&0.5\%	&84.44	&\light 94.32	&82.80	&\light 63.25	&84.44	&\light 94.32	&85.58	&\light 94.97	&84.56	&\light 93.27	&83.40	&\light 75.43	&74.22	&\light 62.34	&73.84	&\grc 3.72	&76.58	&\grc 12.63	\\
\multicolumn{1}{c|}{}	&&1\%	&84.90	&\light 98.04	&83.36	&\light 69.21	&80.21	&\light 70.21	&85.04	&\light 97.58	&84.54	&\light 97.70	&83.86	&\light 82.00	&73.10	&\light 61.25	&69.24	&\grc 0.53	&75.88	&\grc 6.35	\\ \cline{2-21}
\multicolumn{1}{c|}{}&\multirow{2}*{\shortstack{LF}}	&0.5\%	&84.24	&\light 98.87	&83.10	&\light 50.26	&84.24	&\light 98.87	&85.70	&\light 97.70	&81.32	&\light 86.20	&83.80	&\light 70.48	&74.36	&\light 74.97	&75.22	&\grc 0.18	&78.78	&\grc 5.39	\\
\multicolumn{1}{c|}{}	&&1\%	&83.92	&\light 99.56	&83.00	&\light 35.82	&76.76	&\light 49.87	&85.30	&\light 99.03	&81.10	&\light 88.53	&83.40	&\light 70.69	&71.06	&\light 22.32	&67.38	&\grc 2.93	&78.58	&\grc 5.41	\\ \cline{2-21}
\multicolumn{1}{c|}{}&\multirow{2}*{\shortstack{SSBA}}	&0.5\%	&84.30	&\light 95.31	&83.34	&\light 46.75	&84.30	&\light 95.31	&85.04	&\light 95.13	&76.96	&\grc 6.18	&83.16	&\light 15.70	&71.52	&\grc 1.19	&76.12	&\grc 0.89	&79.56	&\grc 1.45	\\
\multicolumn{1}{c|}{}	&&1\%	&84.02	&\light 99.43	&83.34	&\light 59.68	&78.47	&\light 70.78	&85.14	&\light 97.72	&80.22	&\light 22.77	&83.50	&\light 20.30	&72.58	&\grc 7.45	&73.94	&\grc 0.36	&79.88	&\grc 4.91	\\ \cline{2-21}
\multicolumn{1}{c|}{}&\multirow{2}*{\shortstack{SBL-Blended}}	&0.5\%	&72.52	&\light 97.56	&85.10	&\light 89.29	&72.52	&\light 97.56	&82.54	&\light 92.63	&68.28	&\light 97.05	&83.84	&\light 89.07	&70.78	&\light 37.35	&73.66	&\grc 14.87	&79.42	&\grc 7.18	\\
\multicolumn{1}{c|}{}	&&1\%	&72.68	&\light 99.17	&83.41	&\light 68.24	&71.42	&\light 70.14	&82.82	&\light 95.78	&72.72	&\light 99.15	&83.92	&\light 92.69	&73.52	&\light 20.30	&76.78	&\light 39.29	&77.16	&\grc 8.87	\\ \cline{2-21}
\multicolumn{1}{c|}{}&\multirow{2}*{\shortstack{SAPA}}	&0.5\%	&85.04	&\light 98.83	&83.44	&\light 20.53	&78.57	&\grc 9.20	&85.60	&\light 93.05	&80.42	&\light 96.59	&83.82	&\light 30.04	&69.32	&\light 18.34	&73.34	&\grc 1.07	&79.00	&\grc 1.74	\\
\multicolumn{1}{c|}{}	&&1\%	&85.50	&\light 98.83	&83.34	&\light 27.86	&77.42	&\grc 3.52	&85.42	&\light 95.80	&83.12	&\light 93.76	&83.96	&\light 45.88	&69.12	&\light 41.64	&75.42	&\grc 1.23	&78.14	&\grc 1.41	\\ \cline{2-21}
\multicolumn{1}{c|}{}&\multirow{2}*{\shortstack{LC}}	&0.5\%	&84.22	&\grc 0.61	&83.34	&\grc 0.20	&84.22	&\grc 0.61	&85.04	&\grc 0.93	&84.48	&\grc 0.57	&83.86	&\grc 0.24	&69.72	&\grc 0.36	&76.20	&\grc 0.16	&80.36	&\grc 0.22	\\
\multicolumn{1}{c|}{}	&&1\%	&84.10	&\light 32.97	&83.48	&\grc 4.28	&81.13	&\grc 0.42	&85.48	&\light 76.75	&84.22	&\light 32.42	&83.58	&\grc 8.06	&73.52	&\grc 1.76	&70.50	&\grc 0.79	&80.18	&\grc 0.57	\\ \cline{2-21}
\multicolumn{1}{c|}{}&\multirow{2}*{\shortstack{SIG}}	&0.5\%	&84.20	&\light 16.22	&83.26	&\grc 2.22	&78.48	&\grc 0.43	&85.18	&\light 18.34	&84.00	&\light 15.86	&83.88	&\grc 4.51	&70.76	&\grc 3.43	&75.22	&\grc 0.12	&77.20	&\grc 0.55	\\
\multicolumn{1}{c|}{}	&&1\%	&84.16	&\light 70.08	&83.40	&\light 20.48	&79.58	&\grc 0.89	&85.02	&\light 77.84	&80.40	&\light 65.68	&83.36	&\light 44.81	&70.98	&\light 19.98	&73.76	&\grc 0.30	&80.22	&\grc 9.98	\\ 

\hline
        \end{tabular}
        }
    \label{tab:cifar10_imagenet}
    \vspace*{-0.1em}
\end{table*}

%% file: tab/simCLR.tex
\begin{table}[th]
    \vspace*{-0.5em}
    \caption{
        Defense results under SimCLR training scenario, 
        where linear probing is used to evaluate the downstream tasks.
    } 
    \vspace*{0.5em}
    \centering 
    \renewcommand{\arraystretch}{1.05}
    \resizebox{1\linewidth}{!}{
        \setlength\arrayrulewidth{1.2pt}
        \begin{tabular}{cc *{3}{cc}}
        \hline
        \multirow{2}*{\shortstack{Pre-training\\Dataset}} &
        \multirow{2}*{\shortstack{Downstream\\Dataset}} &
        \multicolumn{2}{c}{No Defense} & 
        \multicolumn{2}{c}{MCR} &
        \multicolumn{2}{c}{TSC (ours)} \\[2pt]
        \cmidrule(r){3-4} \cmidrule(lr){5-6} \cmidrule(l){7-8}
        & & ACC($\uparrow$) & ASR($\downarrow$) & ACC($\uparrow$) & ASR($\downarrow$) & 
        ACC($\uparrow$) & ASR($\downarrow$) \\[2pt]
        \hline
        \multirow{3}*{\shortstack{CIFAR10}}	&STL10	&76.74	&\light 99.65	&74.93	&\grc 7.92	&71.11	&\grc 4.44	\\
        &GTSRB	&81.12	&\light 98.79	&75.51	&\grc 0.54	&77.57	&\grc 1.68	\\
        &SVHN	&63.12	&\light 98.71	&57.35	&\light 65.58	&64.13	&\grc 10.26	\\ \hline
    \multirow{3}*{\shortstack{ImageNet}}	&STL10	&94.93	&\light 98.99	&90.20	&\grc 2.08	&86.99	&\grc 3.11	\\
        &GTSRB	&75.94	&\light 99.76	&72.38	&\grc 0.13	&69.47	&\grc 6.47	\\
        &SVHN	&72.64	&\light 99.21	&71.27	&\light 34.15	&66.44	&\grc 3.64	\\ \hline
           
        \end{tabular}
        }
    \label{tab:simclr_defense}
    \vspace*{-0.5em}
\end{table}

%% file: tab/CLIP.tex
\begin{table}[ht]
    \vspace*{-0.5em}
    \caption{
        Defense results under CLIP training scenario, 
        where linear probing and zero-shot learning are used to evaluate the downstream tasks.
        } 
    \vspace*{0.5em}
    \centering 
    \renewcommand{\arraystretch}{1.05}
    \resizebox{1\linewidth}{!}{
        \setlength\arrayrulewidth{1.2pt}
        \begin{tabular}{cc *{3}{cc}}
        \hline
        \multirow{2}*{\shortstack{Pre-training\\Dataset}} &
        \multirow{2}*{\shortstack{Downstream\\Dataset}} &
        \multicolumn{2}{c}{No Defense} & 
        \multicolumn{2}{c}{MCR} &
        \multicolumn{2}{c}{TSC (ours)} \\[2pt]
        \cmidrule(r){3-4} \cmidrule(lr){5-6} \cmidrule(l){7-8}
        & & ACC($\uparrow$) & ASR($\downarrow$) & ACC($\uparrow$) & ASR($\downarrow$) & 
        ACC($\uparrow$) & ASR($\downarrow$) \\[2pt]
        \hline
        \multirow{3}*{\shortstack{CLIP\\(linear probe)}}	&STL10	&97.07	&\light 99.33	&96.43	&\light 99.86	&94.15	&\grc 0.67	\\
        &Food101	&72.58	&\light 97.91	&72.36	&\light 96.62	&69.33	&\grc 1.04	\\
        &VOC 2007	&76.07	&\light 99.83	&75.47	&\light 99.92	&78.42	&\grc 0.34	\\ \hline
    \multirow{3}*{\shortstack{CLIP\\(zero-shot)}}	&STL10	&94.06	&\light 99.86	&91.51	&\light 99.85	&90.25	&\grc 0.88	\\
        &Food101	&67.72	&\light 99.96	&66.51	&\light 99.56	&61.69	&\grc 0.28	\\
        &VOC 2007	&71.22	&\light 99.92	&70.09	&\light 99.12	&75.08	&\grc 1.45	\\ \hline

        \end{tabular}
        }
    \label{tab:clip_defense}
    \vspace*{-0.5em}
\end{table}

%% file: sections/5_appendix.tex
\section{Applicability of the Defense Setting of TSC}
\label{applicability}

As stated in \cref{threat_model}, this paper considers two attack scenarios: one where an adversary can only poison a portion of the training data, and another where the adversary gains control over the training procedure.
In our defense setting, we assume that the defender has knowledge of the original basic training methods such as Stochastic Gradient Descent (SGD) using a Cross-Entropy loss function.
To address potential concerns about the practical relevance of this defense setting, we provide here a detailed discussion of how our proposed defense approach applies in real-world scenarios.

\subsection{Data-poisoning Attacks} 
For adversaries employing data poisoning (who have no control over the training process), 
defenders may repair the backdoored model post-training using defenses such as TSC or other post-purification techniques.
In this scenario, as the defender has control of the training it's natural the defender has the knowledge of the training procedure. 
Thus, the setting in our paper is applicable in such data poison scenario.

\subsection{Training-manipulation Attack} 
For adversaries have control over the training, we provides two common examples here:
\begin{itemize}
    \item Public Pre-trained Models: 
    Public repositories or research papers release pre-trained models that may contain backdoors. 
    Since these sources typically provide detailed descriptions of the model-training procedure, 
    defenders can leverage this information to apply TSC effectively. 
    Using public large-scale image encoders for downstream tasks is increasingly common, making our setting practically relevant. 
    Advanced zero-shot deployment models (\textit{e.g.}, CLIP) further exemplify this applicability.
    \item Internal Adversary in Organizations: 
    Consider an internal adversary scenario within an organization where malicious attackers backdoor a model without others' awareness. 
    Typically, benign team members possess knowledge of the basic training process but lack insight into the malicious manipulations. 
    In this context, defenders within the organization can deploy TSC to purify the model without taking retraining from scratch.
\end{itemize}

Moreover, beyond traditional learning scenarios, our method shows potential for application in other settings.
For example, in federated learning, models are trained collaboratively across many distributed devices, with participants computing and sending their local gradients for global aggregation.
Malicious participants could inject poisoned updates into this system, thereby introducing backdoors, even without direct access to the overall training procedure.
Since a common underlying training methodology is typically employed by both clients and the server in such federated architectures, defenders in this setting could apply our method to remove backdoors using only a small amount of data.

\section{Quadratic Mode Connectivity}
\label{Quadratic_mc}
In \cref{pre_mode_connectivity_pre}, 
we can find a path connecting \(\bm{\theta}_A\) and \(\bm{\theta}_B\) using \cref{curve_loss}. 
However, since \(p_{\bm{\theta}_{A,B}}(t)\) depends on \(\bm{\theta}_{A,B}\), 
it is intractable to compute the stochastic gradients of \(\ell({\bm{\theta}_{A,B}})\) in \cref{curve_loss}. 
To address this, Garipov \citet{ModeConnectivity} choose the uniform distribution \(U(0,1)\) over the interval \([0,1]\) 
to replace \(p_{\bm{\theta}_{A,B}}(t)\), leading to the following loss:
\begin{equation}
   \label{eq2}
   \ell'({\bm{\theta}_{A,B}}) = \int_{0}^{1} \mathcal{L}(\bm{\gamma}_{\bm{\theta}_{A,B}}(t)) \,\mathrm{d}t= \mathbb{E}_{t \sim U(0,1)} \mathcal{L}(\bm{\gamma}_{\bm{\theta}_{A,B}}(t)),
\end{equation}
The primary contrast between \eqref{curve_loss} and \eqref{eq2} is that the former calculates the average loss $\mathcal{L}(\bm{\gamma}_{\bm{\theta}_{A,B}}(t))$ over a uniform distribution along the curve, 
while the latter calculates the average loss over a uniform distribution within the interval $[0,1]$ for the variable $t$.
To minimize $\ell'(\bm{\theta}_{A,B})$, at each step one can randomly select a sample $\hat t$ from the uniform distribution over the interval $[0,1]$ and update the value of $\bm{\theta}_{A,B}$ based on the gradient of the loss function $\mathcal{L}(\bm{\gamma}_{\bm{\theta}_{A,B}}(\hat t))$. 
This implies that we can use $\nabla_{\bm{\theta}_{A,B}} \mathcal{L}(\bm{\gamma}_{\bm{\theta}_{A,B}}(\hat t))$ to estimate the actual gradient of $\ell'(\bm{\theta}_{A,B})$,
\begin{align}
       \nabla_{\bm{\theta}_{A,B}} \ell'(\theta) &= \nabla_{\bm{\theta}_{A,B}} \mathbb{E}_{t \sim U(0, 1)} \mathcal{L}(\bm{\gamma}_{\bm{\theta}_{A,B}}(t))\\
       & = \mathbb{E}_{t \sim U(0, 1)} \nabla_{\bm{\theta}_{A,B}} \mathcal{L}(\bm{\gamma}_{\bm{\theta}_{A,B}}(t))\\
       &\backsimeq \nabla_{\bm{\theta}_{A,B}} \mathcal{L}(\bm{\gamma}_{\bm{\theta}_{A,B}}(\hat t)).
\end{align}

We can choose the \textbf{Bézier curve} as the basic parametric function to characterize the parametric curve $\bm{\gamma}_{\bm{\theta}_{A,B}}(t)$.
And we could initialize $\bm{\theta}_{A,B}$ with $\frac{1}{2}(\bm{\theta}_{A}+\bm{\theta}_{B})$.
A Bézier curve provides a convenient parametrization of smooth paths with given endpoints. 
We can reform the parametric Bézier curve $\bm{\gamma}_{\bm{\theta}_{A,B}}$ in \cref{bezier_curve} by replacing $\bm{\theta}_{A,B}$ with its deviation $\tilde{\bm{\theta}}_{A,B}$:
\begin{align}
    \label{bezier_curve_deviation}
    \bm{\gamma}_{\bm{\theta}_{A,B}}(t) &= (1 -t)^2  \bm{\theta}_{A} + 2t(1-t) \bm{\theta}_{A,B} + t^2  \bm{\theta}_{B} \nonumber 
    \\
    &= (1 -t)^2  \bm{\theta}_{A} + 2t(1-t)(\frac{\bm{\theta}_{A}+\bm{\theta}_{B}}{2}+ \tilde{\bm{\theta}}_{A,B}) + t^2  \bm{\theta}_{B} \nonumber 
    \\
    &= (1-t)\bm{\theta}_{A} + t\bm{\theta}_{B} +  2t(1-t)\tilde{\bm{\theta}}_{A,B} =\bm{\gamma}_{\tilde{\bm{\theta}}_{A,B}}(t).
\end{align}

\section{Permutation Invariance and Neuron Alignment}
\label{Permutation invariance}
After applying the permutation operation $\pi(\bm{\theta},S(\bm{P}))$,
the feedforward neural network defined in \cref{feedforward_network} 
is transformed to:
\begin{equation}
    \begin{aligned}
        \label{permuted_feedforward_network}
        \bm{y}&= \bm{W}_{L} \circ \sigma \circ \bm{P}_{L-1}^{\top} \bm{P}_{L-1} \bm{W}_{L-1} \circ ... \circ \sigma \circ  \bm{P}_{1}^{\top} \bm{P}_{1}\bm{W}_{1} \bm{x}_0 \\
        &= \bm{W}_{L}\bm{P}_{L-1}^{\top}  \circ \sigma \circ \bm{P}_{L-1} \bm{W}_{L-1} \bm{P}_{L-2}^{\top} \circ ... \circ \sigma \circ \bm{P}_{1}\bm{W}_{1} \bm{x}_0.\\ 
    \end{aligned}
\end{equation}
The second equation in (\ref{permuted_feedforward_network}) follows from the fact that \(\sigma\) is an element-wise function.
The weights of the permuted network can then be obtained as defined in \cref{weights_permutation}.

Previous studies have found that two functionally identical networks $\bm{\theta}_A$ and $\bm{\theta}_B$, 
trained independently with the same architecture but different random initializations or SGD solutions, 
could be misaligned.
And the loss of their linearly interpolated network,
represented by $\bm{\theta}_t = t \bm{\theta}_A + (1 - t) \bm{\theta}_B$ (where $0 \leq t \leq 1$),
could be quite large \cite{permutation,LMC}.
However, \citet{permutation} conjecture that 
if the permutation invariance of neural networks is taken into account,
then networks obtained by all SGD solutions could be linearly connected.

Recently, reserach \cite{model_fusion, git_rebasin,repair} has shown that such misaligned networks $\bm{\theta}_A$ and $\bm{\theta}_B$ 
could be projected to the same loss basin using a specific set of permutation matrices $S(\hat{\bm{P}})$.
Subsequently, these networks can be fused through a linear path,
\textit{i.e.,} they are linear mode connected.
For example, one can let $\bm{\theta}_B$ re-aligned with $\bm{\theta}_A$ 
by permuting $\bm{\theta}_B$ to $\bm{\theta}_B^{S(\hat{\bm{P}})}$,
enabling the linearly interpolated network of $\bm{\theta}_A$ and $\bm{\theta}_B^{S(\hat{\bm{P}})}$ to exhibit performance similar to both $\bm{\theta}_A$ and $\bm{\theta}_B$.

To solve the problem defined in \cref{basic_na_minimization},
various cost functions $c_l$ can be employed to compute $\bm{\hat{P}}_l$ \cite{convergent_learning,model_fusion,git_rebasin}.
One commonly used $c_l$ is defined as $c_l=1-\text{corr}(\bm{v},\; \bm{z})$ in \cite{convergent_learning},
where the $\text{corr}$ compute the correlation between $\bm{v} \in \mathbb{R}^{d_l}$ and $\bm{z}\in \mathbb{R}^{d_l}$.
\citet{model_fusion} utilized optimal transport to soft-align neurons before model fusion.
\citet{git_rebasin} introduced two novel alignment algorithm and compared them with the method in \cite{convergent_learning}.
\citet{repair} proposed enhancing the linear mode connectivity after alignment by renormalizing the activations. 
While these studies aimed to apply neuron alignment for linear interpolated networks,
\citet{optMC_na} found that alignment could improve both robustness and accuracy along the quadratic curve connecting adversarially robust models.
Following previous work, we continue to use this alignment method proposed by \citet{convergent_learning}.

\section{Wasserstein distance}
\label{Wasserstein_distance}
Wasserstein distance, a key concept in optimal transport (OT) theory, 
measures the distance between probability distributions by considering the cost of transforming one distribution into another \cite{optimal_transport}. 
It provides a geometric perspective on comparing distributions.
Formally, let $\mathbb{P}_r$ and $\mathbb{P}_q$ be two probability distributions,
and $\Omega (\mathbb{P}_r, \mathbb{P}_r)$ denote the set of all joint distributions \( \omega   \) that have $\mathbb{P}_r$ and $\mathbb{P}_q$ as their marginal distributions.
Then the $p$-Wasserstein distance can be expressed as:
\begin{equation}
    W_p(\mathbb{P}_r, \mathbb{P}_q) = \left( \inf_{\omega \in \Omega(\mathbb{P}_r, \mathbb{P}_q)} \mathbb{E}_{(x,y)\sim \omega} \left\lVert x-y \right\rVert^p  \right)^{1/p}.
\end{equation}
The joint distribution $\omega$ can be regarded as  
the optimal transport solution (\textit{i.e.,} the minimal distance) between $\mathbb{P}_r$ and $\mathbb{P}_q$.
When we solve the problem outlined in \cref{l2_minimization} to find the optimal permutation $\hat{\bm{P}}_l$, 
we are essentially seeking the optimal transport $\omega \in \Pi_{d_l} $ that corresponds to the 2-Wasserstein distance between $\mathbb{P}^A_{l}$ and $\mathbb{P}^B_{l}$.

\section{Omitted Proofs}
\subsection{Proof of \Cref{general_theorem}}
\label{proof_for_general_theorem}

\input{mathematics_proof/proof_1.tex}

\subsection{Proof of \Cref{permutation_lemma}}
\label{proof_for_specific_proof}
\input{mathematics_proof/proof_2.tex}

\section{Practical Algorithm}
\label{practical_algorithm}

\subsection{Computing the Permutation for Each Model Layer}
$\textsc{PermuteLayers}(\bm{\theta}_A, \bm{\theta}_B, D, \textsc{opt})$ in \cref{alg:tsc} returns a model by permuting the layers of $\bm{\theta}_B$, aligning or un-aligning it with $\bm{\theta}_A$. 
To compute the permutation of the $l$\textsuperscript{th} layer of $\bm{\theta}_A$ for alignment with $\bm{\theta}_B$, we first obtain the corresponding activations $\bm{x}_{i,\; l}^{A}$ and $\bm{x}_{i,\; l}^{B}$ for each sample $x^{(i)} \in D$. 
We then employ the cost function $c_l = 1 - \text{corr}(\bm{v}, \bm{z})$ and compute the sum of the cross-correlation matrices $R_{i,l}$ of the normalized $\bm{x}_{i,\; l}^{A}$ and $\bm{x}_{i,\; l}^{B}$ as follows:
\begin{equation}
    R_l = \sum^{|D|}_{i = 1} R_{i,l} = \sum^{|D|}_{i = 1} \frac{\bm{x}_{i,\; l}^{A} - \bm{\mu}_{\bm{x}^{A}_{{i,\; l}}}}{\bm{\Sigma}_{\bm{x}^{A}_{{i,\; l}}}} \frac{\bm{x}_{i,\; l}^{B} - \bm{\mu}_{\bm{x}^{B}_{{i,\; l}}}}{\bm{\Sigma}_{\bm{x}^{B}_{{i,\; l}}}}.
\label{correlation_matrix}
\end{equation}
Finally, we use the Hungarian algorithm \cite{hungarian} to solve the maximization problem (or equivalently the minimization problem) according to $\textsc{opt}$.
It is noteworthy that, to stay consistent with previous work, we compute the cross-correlation matrix here rather than the cost function defined in \cref{l2_minimization} or \cref{l2_maximization}. 
This minimization or maximization procedure is still equivalent to ordinary least squares constrained to the solution space $\Pi_{d_l}$ \cite{git_rebasin,optMC_na}.
We give the pseudocode of computing the permutation matrices for feedforward networks in \cref{alg:permutation_layer}.
For more complex model architectures, the same principles apply, but the implementation details may vary.
We refer readers to \citet{git_rebasin} and \citet{repair} for more details on computing the permutation matrices for other architectures, such as ResNet-18 \cite{ResNet} and VGG19-BN \cite{vgg19}.
\input{algorithms/permutation_layer.tex}

\subsection{Training the Quadratic Bézier Curve}
$\textsc{FitQuadCurve}(\bm{\theta}_A, \bm{\theta}_B, \mathcal{F}, D, e)$
returns the quadratic Bézier curve connecting $\bm{\theta}_A$ and $\bm{\theta}_B$, trained using method $\mathcal{F}$ over $D$ for $e$ epochs.
As mentioned in \cref{Quadratic_mc}, at each step, 
we randomly select a sample $\hat t$ from the uniform distribution over the interval $[0, 1]$ and update the value of $\bm{\theta}_{A,B}$ based on the gradient of the loss computed using the training method $\mathcal{F}$. 
This is achieved by computing the loss of $\bm{\theta}_{\hat t}$ over the dataset $D$ using $\mathcal{F}$, 
and then calculating the gradient of $\bm{\theta}_{A,B}$ (\textit{i.e.}, the weights of the curve $\bm{\gamma}_{\bm{\theta}_{A, B}}(t)$) via the chain rule.
The pseudocode is provided in \cref{alg:fit_curve}.
\input{algorithms/curve_training.tex}

\section{Evaluating \shortname's Robustness Against Adaptive Attacks}
\label{adaptive_attack}
\subsection{Apdative Attack Design}
The key defense mechanism of \shortname\ relies on increasing adversarial loss along the quadratic Bézier curve 
by projecting model $\bm{\theta}_{adv}$ to a distinct loss basin to find $\bm{\theta}_{adv'}$ 
during its first stage. 
An adaptive attack would aim to create a backdoored model maintaining low backdoor loss along this defensive curve.

Here, we assume the adversary has access to the model training procedure. 
Building upon the neural network subspace learning approach from \cite{learning_subspace}, 
originally developed for accuracy and calibration improvements, 
we design an adaptive attack strategy. 
The key insight is to place $\bm{\theta}_{adv}$ and 
its symmetric point $\bm{\theta}_{adv'}$ 
in a subspace that minimizes loss on poisoned samples. 
Our approach simultaneously trains a curve and updates its endpoints $\bm{\theta}_{adv}$ and $\bm{\theta}_{adv'}$ 
using mixed benign and poisoned data (\textit{i.e.}, learning a backdoored subspace). 
After training, we select one endpoint as the final model.

The implementation details are provided in \cref{alg:adaptive}. 
We first project the backdoored model $\bm{\theta}_{adv}$ 
to a symmetric loss basin to obtain $\bm{\theta}_{adv'}$ 
by solving \cref{l2_maximization}, 
similar to \shortname's initial process. 
We then train the curve connecting $\bm{\theta}_{adv}$ and 
$\bm{\theta}_{adv'}$ over $D_{adv}$ without fixed endpoints, 
ensuring the curve lies within the backdoored subspace. 
Finally, we return $\bm{\theta}_{adv}$ as the output model, 
which is expected to achieves lower loss along the curve found by \shortname\ compared to the original model.

\input{algorithms/adaptive_learning_subspace.tex}

\begin{table}[ht]
    \begin{minipage}[b]{0.45\linewidth}
    \centering
    \input{tab/adap_cifar10_sl.tex}
    \end{minipage}
    \hfill
    \begin{minipage}[b]{0.45\linewidth}
        \centering
        \input{tab/adap_in100_sl.tex}
    \end{minipage}
\end{table}

\begin{table}[ht]
    \centering
    \begin{minipage}[c]{0.58\linewidth}
    \vspace*{1.2em}
    \input{tab/adap_simclr_ssl.tex}
    \end{minipage}
\end{table}

\subsection{Empirical Evaluation Against Adaptive Attack}
To evaluate \shortname's robustness against adaptive attacks, 
we apply \cref{alg:adaptive} to convert models backdoored by 
BadNet \cite{badnets}, Blended \cite{TargetGlasses}, SSBA \cite{ISSBA}, LF \cite{lf}, WaNet \cite{wanet}, Input-aware \cite{Inputaware}, LC \cite{firstCleanLabel}, and SIG \cite{SIG} under supervised learning. 
We conduct experiments on CIFAR10 using PreAct-ResNet with 5\% poisoning rate and ImageNet using ResNet50 with 1\% poisoning rate.

For self-supervised learning, 
we adapt BadEncoder \cite{BadEncoder} 
into our adaptive attack framework. 
Using SimCLR, we utilize publicly available backdoored ResNet18 and ResNet50 encoders on CIFAR10 and ImageNet, respectively, 
evaluating ASR and ACC through linear probing on downstream datasets STL10, GTSRB, and SVHN.

Since our adaptive attack is designed to exlpore the robustness of \shortname,
we do not consider other defenses here. 
For fair evaluation, we use the default settings of \shortname : global epoch $E_{\shortname}=3$, curve index $t=0.4$, and curve training epoch $e=200$ for supervised learning; 
$E_{\shortname}=2$, $t=0.25$, and $e=200$ for SimCLR. 
We provide defenders with 5\% clean samples. 
Additional experimental settings follow \cref{appendix:Experimental_Detials}.

The defense results are presented in \cref{tab:adap_cifar10_sl,tab:adap_in100_sl,tab:adap_simclr_ssl}. 
After retraining the backdoored models using \cref{alg:adaptive}, we observe slight improvements in both ACC and ASR compared to the original models.
We suspect that such improvements are due to the benefits of the subspace learning approach \cite{learning_subspace}. 
Importantly, \shortname\ maintains its robustness against these adaptive attacks across both supervised and self-supervised learning settings.

\begin{figure*}[t]
    \centering
    \includegraphics[width=0.97\linewidth]{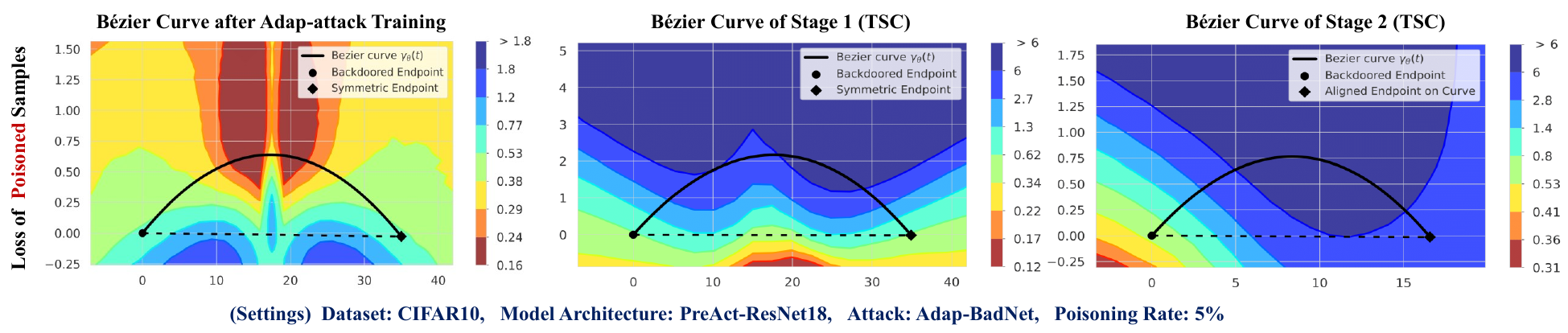}
    \vspace*{-0.7em} 
    \caption{
        Loss landscape for poisoned samples of Adap-Badnet,
        along with trained quadratic curves connecting distinct models.
        The backdoored model is a PreAct-ResNet18 trained on CIFAR-10, which contains 5\% poisoned samples.
        \textbf{Left}: the curve identified by our adaptive attack.
        \textbf{Middle}: the curve identified by the first stage of \shortname.
        \textbf{Right}: the curve identified by the second stage of \shortname.
    }
       \label{fig:adap_loss_landscape}
\end{figure*}

\subsection{Loss Landscape Visualization of Adaptive Attack}
\label{adap_loss_landscape}

To further explore the effectiveness of our defense against adaptive attacks, we present a loss landscape visualization for Adap-BadNet before and after applying \shortname, as shown in \cref{fig:adap_loss_landscape}.
The left plot shows that the adaptive attack's curve training procedure successfully places its backdoored model, $\bm{\theta}_{adv}$, and its corresponding symmetric point, $\bm{\theta}_{adv'}$, in a subspace that minimizes loss on poisoned samples.
However, as depicted in the middle plot, the curve identified by the first stage of \shortname\ traverses a loss basin characterized by considerably higher loss on poisoned samples, rather than the basin exploited by the adaptive attack.
We attribute this to \shortname's training procedure, which exclusively uses benign samples, thereby guiding the curve to circumvent the loss basin optimized by the adaptive attack.
Moreover, the permutation operation in the first stage of \shortname\ could also find an endpoint in a different loss basin from both $\bm{\theta}_{adv}$ and $\bm{\theta}_{adv'}$, 
as there are multiple permutations matrices that satisfy the maximization objective in \cref{l2_maximization} \cite{permutation}. 

Overall, the combination of the permutation mechanism and training exclusively with benign samples contributes to amplifying the loss on poisoned samples.

\section{Ablation Studies}
\label{ablation}

\subsection{Sensitivity Analysis on Global Epochs $E_{\shortname}$ and the Curve Index $t$}
\label{ablation_hyperparameters}

\shortname\ has two key hyperparameters: 
the number of global epochs $E_{\shortname}$ 
and the curve index $t$.

\begin{figure*}[th]
    \centering
    \includegraphics[width=0.9\linewidth]{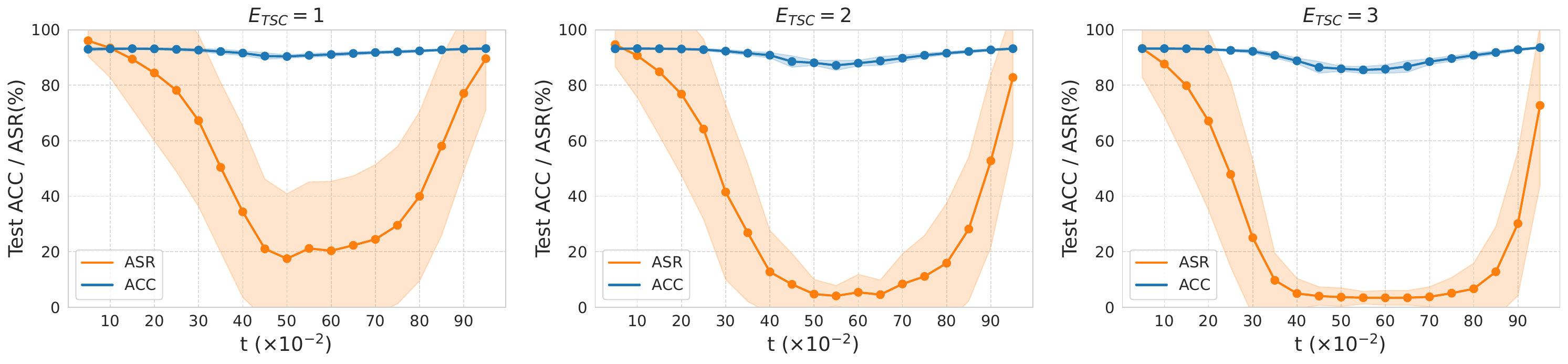}
    \vspace*{-1em}
    \caption{
        \textbf{(Supervised Learning)} Effect of global epoch $E_{\shortname}$ and the curve index $t$ on \shortname\ defense.
        We evaluate the performance of \shortname\ on CIFAR10 with 5\% poisoning rate using PreAct-ResNet18.
        Each point is averaged over the results of \shortname\ against 12 attacks.
    }
       \label{fig:ablation_cifar10}
\end{figure*}

For supervised learning,
we investigate their impact through ablation studies on CIFAR10 using PreAct-ResNet18 with a 5\% poisoning rate under supervised learning. 
\cref{fig:ablation_cifar10} illustrates the purification performance of \shortname\ under varying $E_{\shortname}$ and $t$. 
Each data point represents averaged results against twelve attacks: BadNet \cite{badnets}, Blended \cite{TargetGlasses}, SSBA \cite{ISSBA}, LF \cite{lf}, WaNet \cite{wanet}, Inputaware \cite{Inputaware}, LC \cite{firstCleanLabel}, SIG \cite{SIG}, SBL-BadNet, SBL-Blended \cite{sbl}, Narcissus \cite{narcissus} and SAPA \cite{sapa}.

\begin{figure*}[th]
    \centering
    \includegraphics[width=0.9\linewidth]{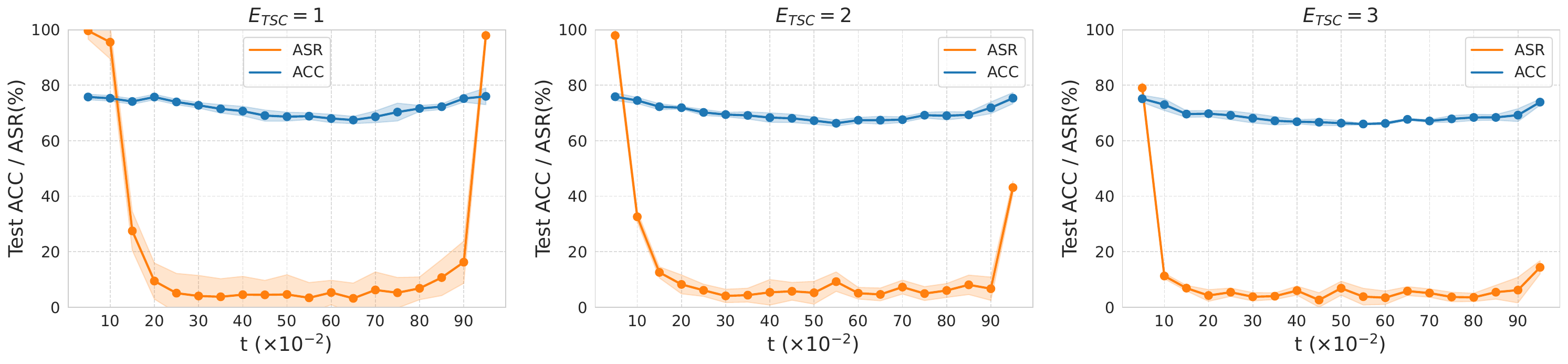}
    \vspace*{-1em}
    \caption{
        \textbf{(Self-Supervised Learning, SimCLR)} Effect of global epoch $E_{\shortname}$ and the curve index $t$ on \shortname\ defense.
        We evaluate the performance of \shortname\ against BadEncoder on pre-training dataset CIFAR10 and downstream dataset STL10 with ResNet18.
        We performed 3 defense runs for each point and averaged the results in the figure.
    }
       \label{fig:ablation_ssl_cifar10_stl10}
\end{figure*}

For self-supervised learning, 
we conduct experiments on pre-training dataset CIFAR10 and downstream dataset STL10 using ResNet18 to evaluate the performance of \shortname\ against BadEncoder \cite{BadEncoder}.
\cref{fig:ablation_ssl_cifar10_stl10} shows the results of \shortname\ under different $E_{\shortname}$ and $t$ values.

\textbf{Importantly}, each point in \cref{fig:ablation_cifar10,fig:ablation_ssl_cifar10_stl10} shows the final ACC and ASR values of \shortname\ using different combinations of $E_{\shortname}$ and $t$. 
This differs from \cref{fig:curve_acc}, \cref{fig:curve_acc_cifar10_0.1}, and similar figures, which display ACC and ASR values evaluated along the curve at each round of $E_{\shortname}$ with fixed $t$.

The results in \cref{fig:ablation_cifar10} demonstrate that increasing the curve index $t$ to $0.5$ enhances purification performance at the cost of slightly reduced clean accuracy (ACC) per epoch. 
Similarly, as $t$ increases from $0.5$ to $0.95$, ACC improves but purification effectiveness decreases.
As previously noted, when $t \geq 0.4$ (or $t \leq$ 0.6),  
a single epoch of \shortname\ proves insufficient for backdoor removal.
Furthermore, increasing the number of global epochs $E_{\shortname}$ could enhance the stability of the defense performance against various attacks.
Our empirical analysis suggests $t=0.4$ and $E_{\shortname}=3$ as optimal parameters for supervised learning settings.

We can observe that the performance of \shortname\ on self-supervised learning is consistent with that of supervised learning.
The key difference is that a smaller curve index $t$ suffices for backdoor removal in self-supervised learning. 
Though \cref{fig:ablation_ssl_cifar10_stl10} (left) demonstrates 
that setting $t=0.2$ and $E_{\shortname}=1$ achieves effective purification,
we opt for more conservative hyperparameters ($t=0.25$ and $E_{\shortname}=2$) to ensure robust performance across diverse attack scenarios.

Additional results for ACC and ASR along the curve with fixed $t=0.4$ are presented in \cref{fig:curve_acc_cifar10_0.1,fig:curve_acc_cifar10_0.05,fig:curve_acc_in100_0.01,fig:curve_acc_in100_0.005}. 
While $E_{\shortname}=1$ or $E_{\shortname}=2$ can occasionally reduce ASR to near zero, $E_{\shortname}=3$ provides more consistent robustness across different attacks.
For conservative defenders, we recommend larger values of $E_{\shortname}$ and $t$ as the hyperparameters.

\begin{table}[!h]
    \begin{minipage}[b]{0.48\linewidth}
    \centering
    \input{tab/ablation_model_cifar10_vgg.tex}
    \end{minipage}
    \hfill
    \begin{minipage}[b]{0.48\linewidth}
        \centering
        \input{tab/ablation_model_cifar10_iv3.tex}
    \end{minipage}
\end{table}

\subsection{Analysis on Model Architecture}
\label{ablation_model_results}

To evaluate the stability of \shortname\ across different architectures,
we conduct experiments using VGG19-BN \cite{vgg19} and
InceptionV3 \cite{inceptionv3} on the CIFAR10 dataset under supervised learning settings.
We set the poisoning rate to 5\% for all attacks 
and employ the default hyperparameters for MCR and \shortname. 
The results are presented in \cref{tab:ablation_model_cifar10_vgg} and \cref{tab:ablation_model_cifar10_inceptionv3} respectively.

The results demonstrate that \shortname\ maintains robust performance across different model architectures.
While \shortname\ occasionally shows marginally lower ACC compared to MCR,
it consistently demonstrates superior robustness against all considered attacks.

\section{Additional Results for Supervised Learning}
\label{additional_sl_results}

In this section, 
we provide comprehensive results for supervised learning settings.

\begin{figure*}[!h]
    \centering
    \includegraphics[width=1\linewidth]{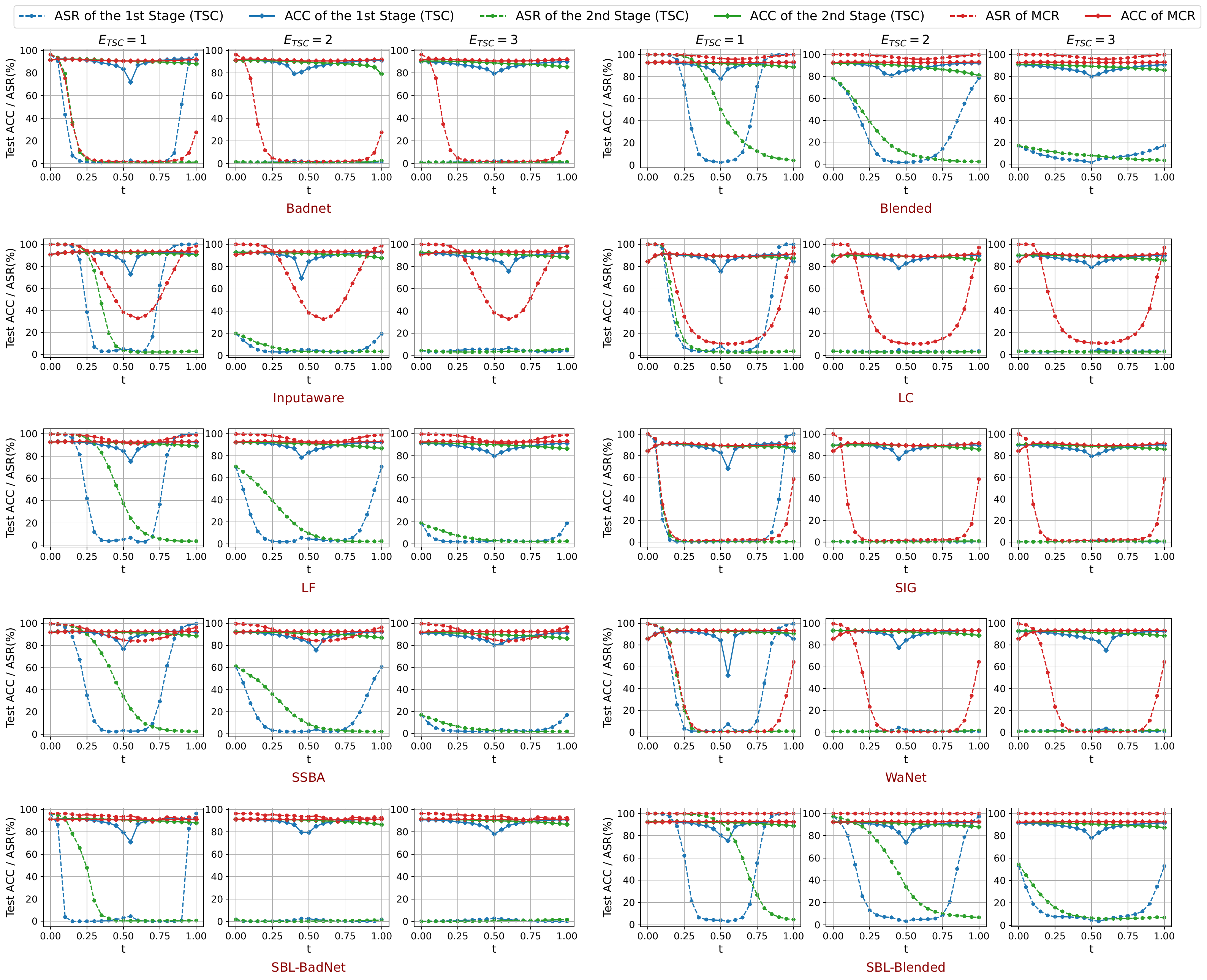}
    \vspace*{-2em}
    \caption{
        \textbf{(Supervised Learning)} Performance of \shortname\ and MCR on CIFAR-10 with 10\% poisoning rate using PreAct-ResNet18.
        Test accuracy (ACC) on benign samples and the attack success rate (ASR) are evaluated as functions of the points along the quadratic Bézier curve found by MCR and \shortname.
        We select model points along the curve at $t=0.4$ for each stage and round.
    }
       \label{fig:curve_acc_cifar10_0.1}
\end{figure*}

\begin{figure*}[!h]
    \centering
    \includegraphics[width=1\linewidth]{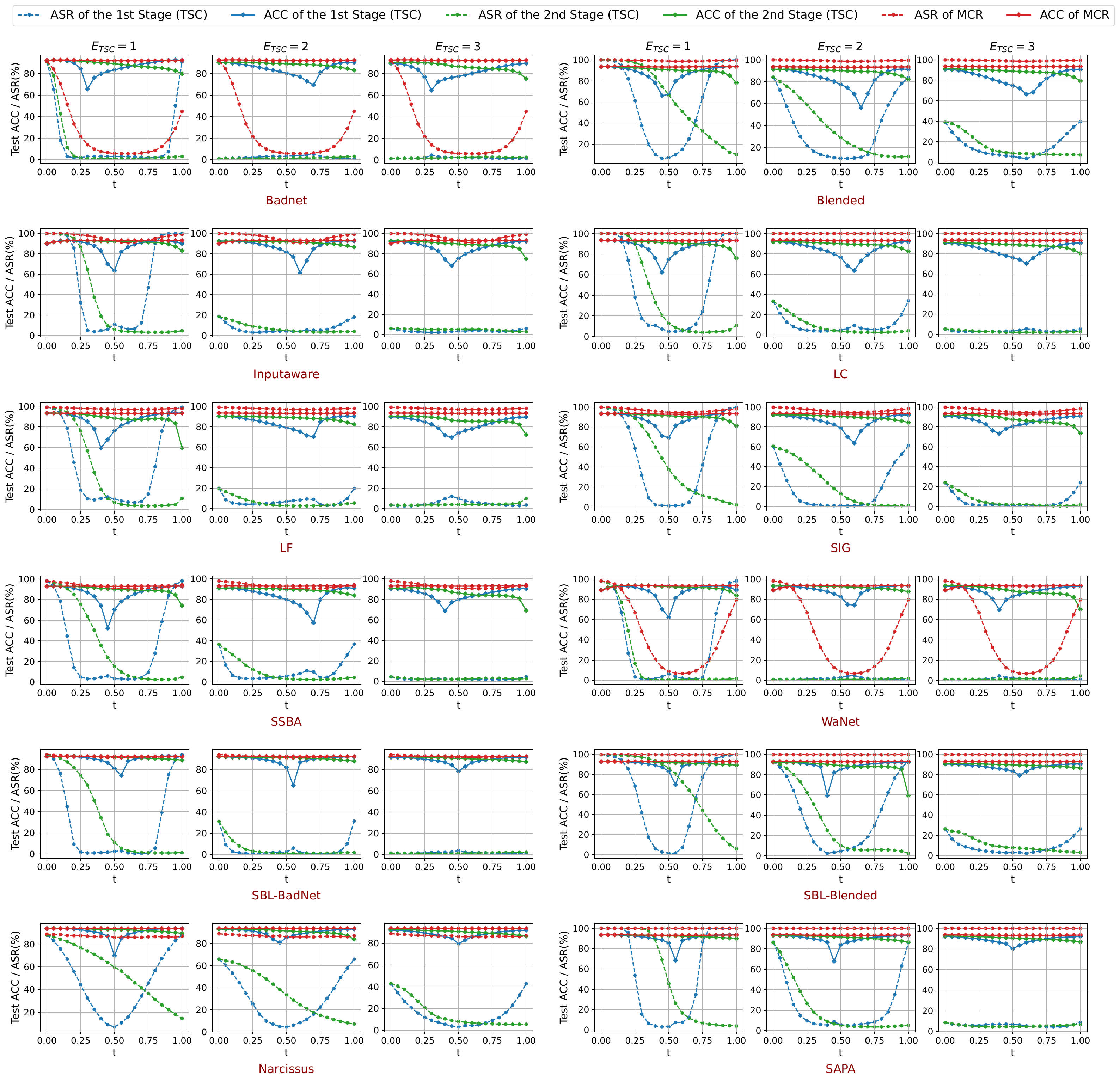}
    \vspace*{-1.5em}
    \caption{
        \textbf{(Supervised Learning)} Performance of \shortname\ and MCR on CIFAR-10 with 5\% poisoning rate using PreAct-ResNet18.
        We select model points along the curve at $t=0.4$ for each stage and round.
    }
       \label{fig:curve_acc_cifar10_0.05}
\end{figure*}

\begin{figure*}[t]
    \centering
    \includegraphics[width=1\linewidth]{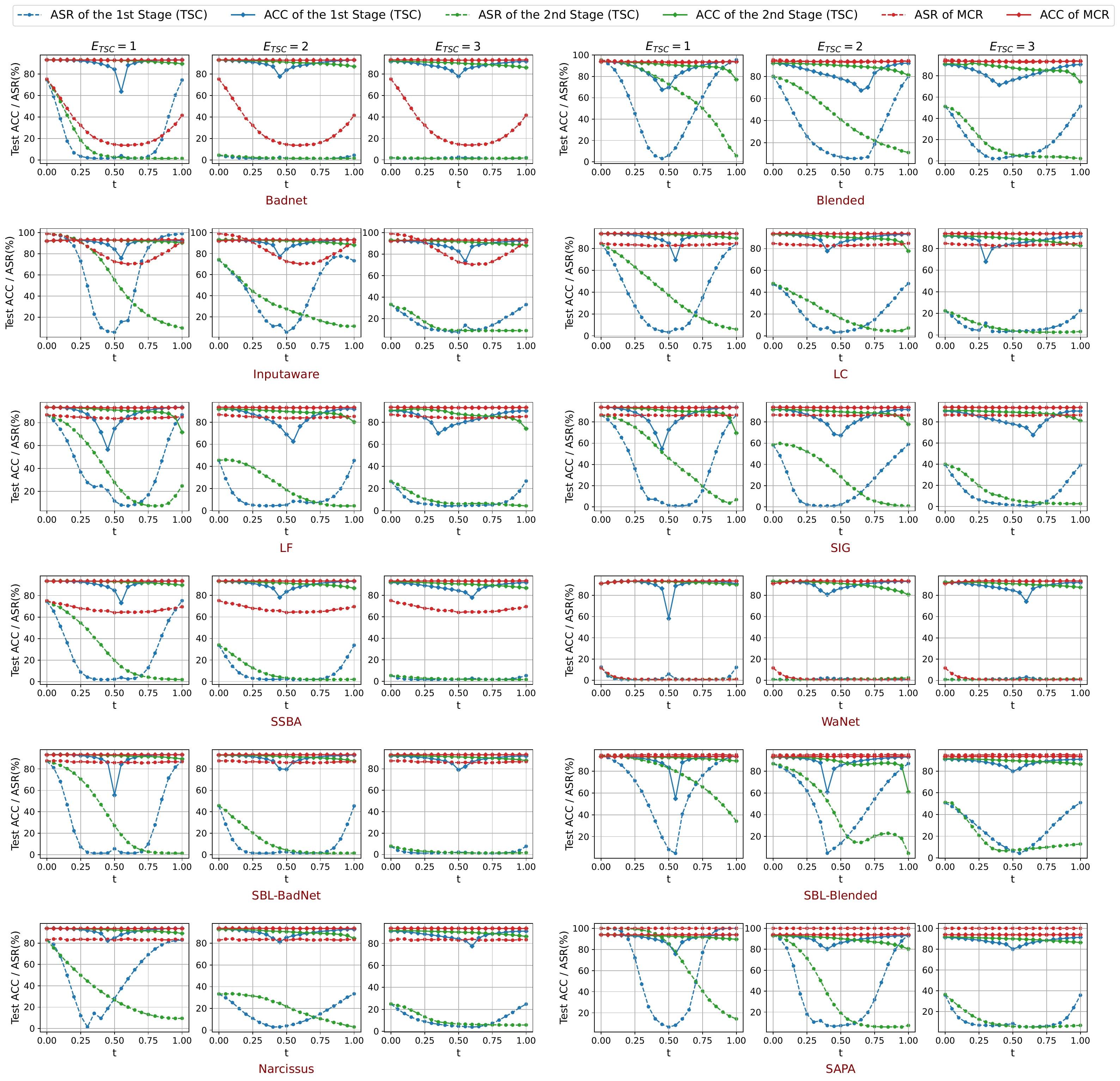}
    \vspace*{-2.5em}
    \caption{
        \textbf{(Supervised Learning)} Performance of \shortname\ and MCR on CIFAR-10 with 1\% poisoning rate using PreAct-ResNet18.
        We select model points along the curve at $t=0.4$ for each stage and round.
    }
       \label{fig:curve_acc_cifar10_0.01}
\end{figure*}

\begin{figure*}[b]
    \centering
    \includegraphics[width=1\linewidth]{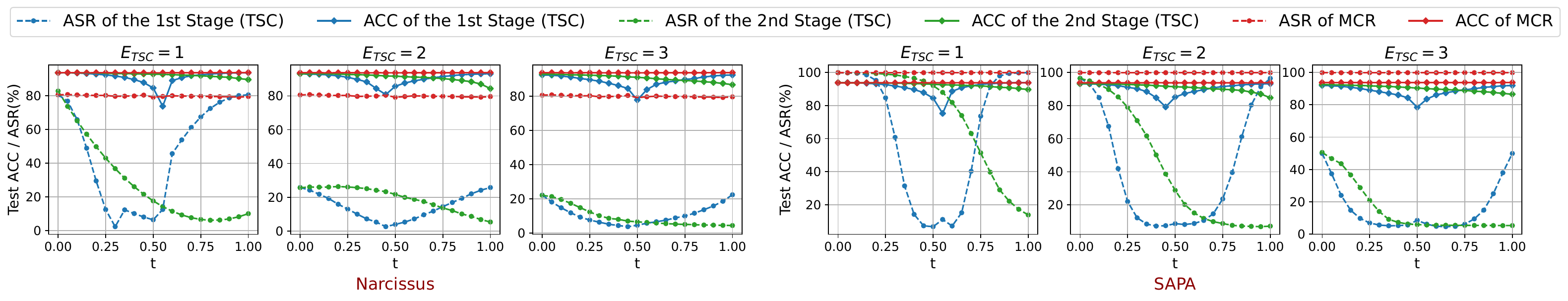}
    \vspace*{-2.5em}
    \caption{
        \textbf{(Supervised Learning)} Performance of \shortname\ and MCR on CIFAR-10 with 0.5\% poisoning rate using PreAct-ResNet18 against Narcissus and SAPA attacks.
        We select model points along the curve at $t=0.4$ for each stage and round.
    }
       \label{fig:curve_acc_cifar10_0.005}
\end{figure*}

\begin{figure*}[!h]
    \centering
    \includegraphics[width=1\linewidth]{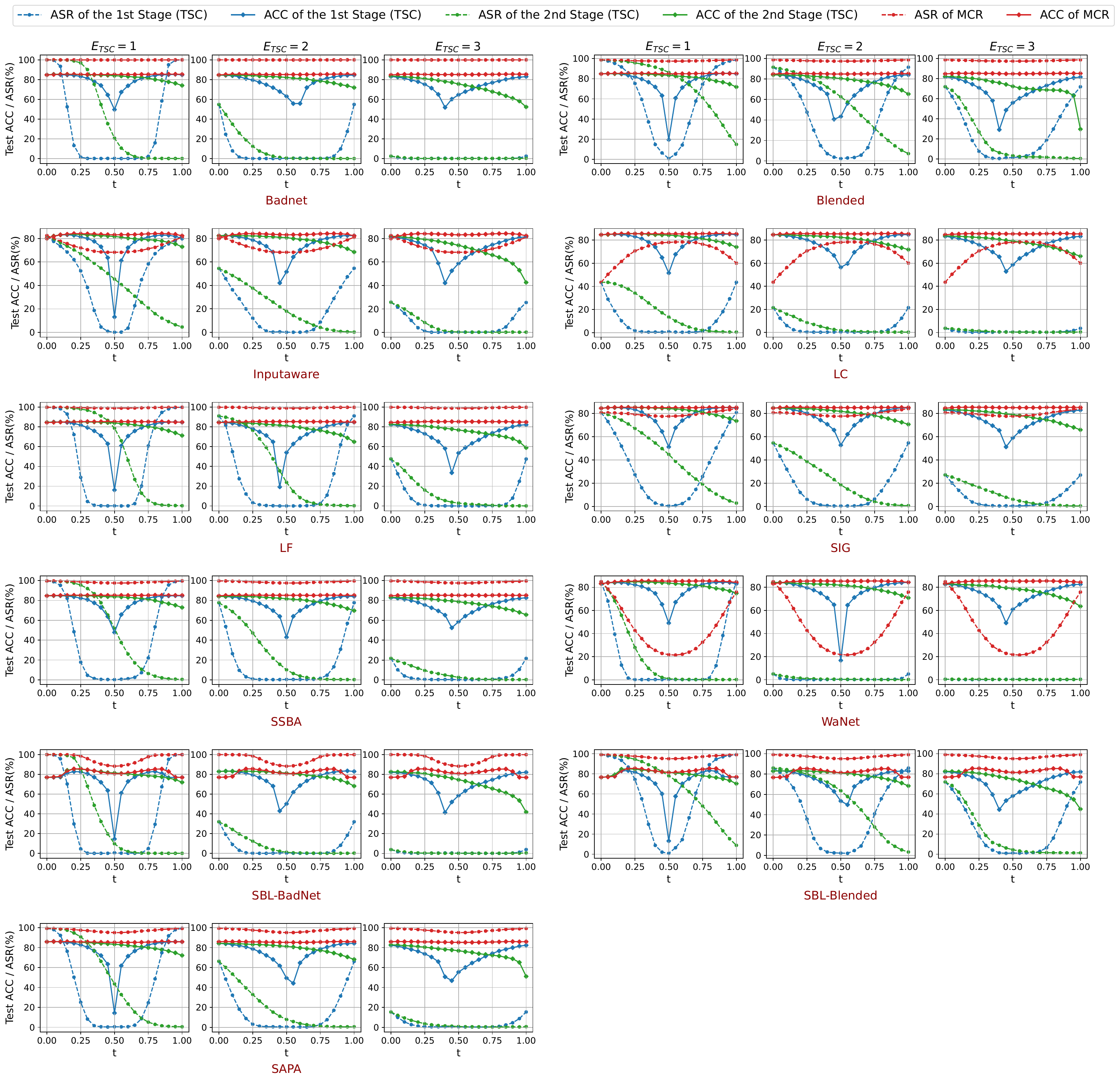}
    \vspace*{-1.5em}
    \caption{
        \textbf{(Supervised Learning)} Performance of \shortname\ and MCR on ImageNet100 with 1\% poisoning rate using ResNet50.
        We select model points along the curve at $t=0.4$ for each stage and round.
    }
       \label{fig:curve_acc_in100_0.01}
\end{figure*}

\begin{figure*}[!h]
    \centering
    \includegraphics[width=1\linewidth]{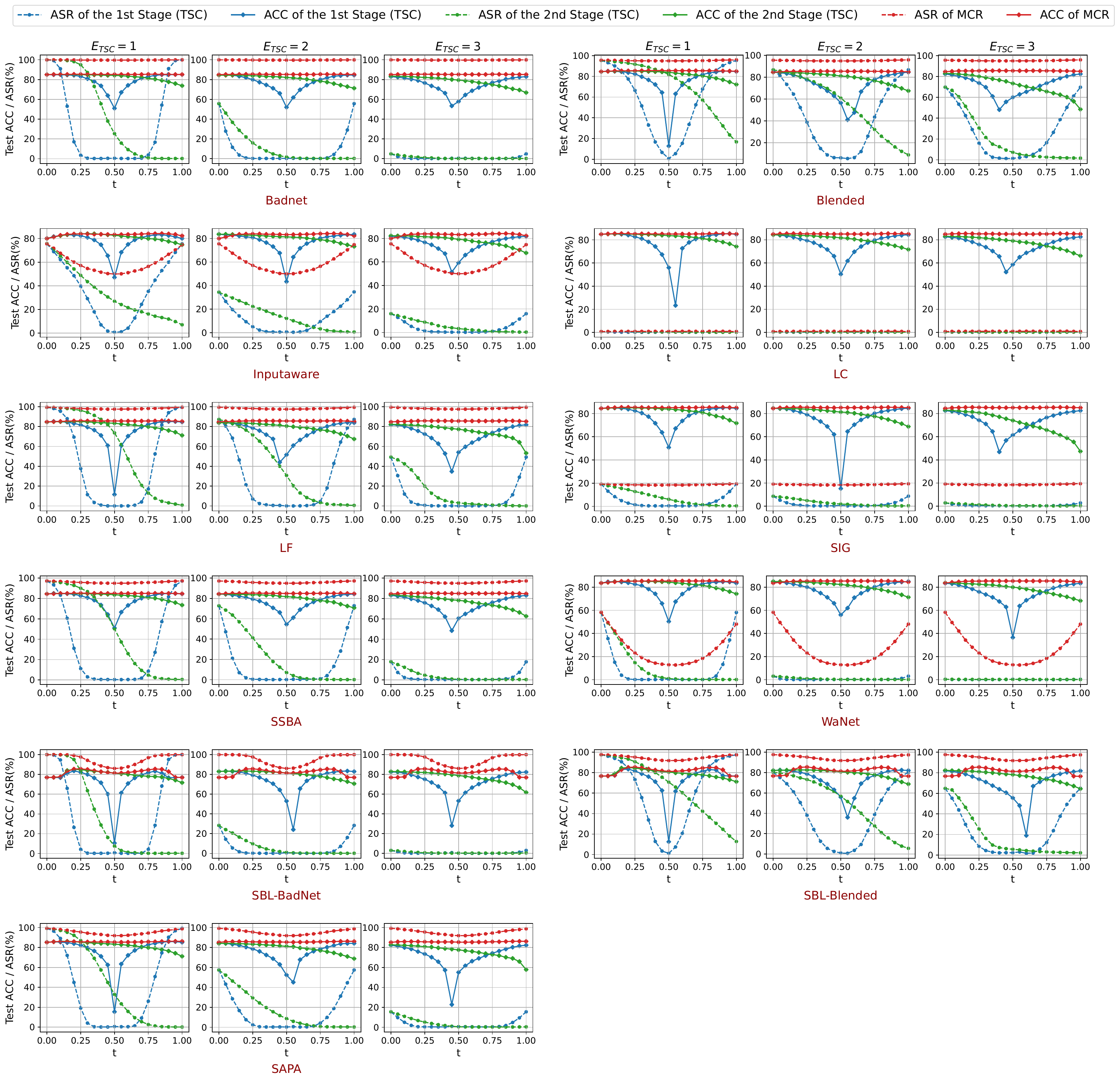}
    \vspace*{-1.5em}
    \caption{
        \textbf{(Supervised Learning)} Performance of \shortname\ and MCR on ImageNet100 with 0.5\% poisoning rate using ResNet50.
        We select model points along the curve at $t=0.4$ for each stage and round.
    }
       \label{fig:curve_acc_in100_0.005}
\end{figure*}

\cref{fig:curve_acc_cifar10_0.1,fig:curve_acc_cifar10_0.005,fig:curve_acc_cifar10_0.05,fig:curve_acc_cifar10_0.01}
illustrates the performance of \shortname\ and MCR on CIFAR10 with 10\%, 5\%, and 1\% poisoning rates, respectively.
\cref{fig:curve_acc_in100_0.01,fig:curve_acc_in100_0.005} present the results on ImageNet100 with 1\% and 0.5\% poisoning rates, respectively.
Taking CIFAR10 as an example, 
we observe that backdoors are more effectively eliminated by \shortname\ when the poisoning rate is relatively high. 
When the poisoning rate is 10\%, 
a small number of global epochs $E_{\shortname}$ is sufficient to remove backdoors implanted 
by attacks such as BadNet, InputAware, and LC. 
However, 
when the poisoning rate decreases to 1\%, $E_{\shortname}=3$ 
is required to defend against all attacks. 
\shortname\ demonstrates similar behavior on the ImageNet100 dataset.
These findings further indicate that conservative hyperparameter settings are reasonable 
in supervised learning scenarios.

\cref{tab:cifar10_defense,tab:imagnet_defense,tab:gtsrb_defense} 
summarize the defense results of \shortname\ and other baseline defenses 
on CIFAR10, ImageNet100, and GTSRB, respectively.
\cref{tab:cifar10_defense,tab:imagnet_defense,tab:gtsrb_defense} 
summarize the defense results of \shortname\ and 
other baseline defenses on CIFAR10, 
ImageNet100, and GTSRB, respectively. 

The experimental results reveal that 
while lower poisoning rates can increase the robustness of attack, 
excessively low poisoning rates sometimes 
result in diminished ASRs on the original model. 
For instance, with a 1\% poisoning rate, 
WaNet attack achieves only 12.63\% initial ASR on CIFAR10; 
with a 0.5\% poisoning rate, 
LC attack yields merely 0.61\% ASR on ImageNet100. 
Although these attacks are not considered successful backdoor attacks, 
we include them in our evaluation for completeness.

Notably, \shortname\ successfully reduces the attack success rate of all attacks to below 15\%. 
As mentioned in our main paper, 
ANP \cite{ANP}, FT-SAM \cite{FT_SAM}, and I-BAU \cite{I_BAU} 
perform well on smaller datasets like CIFAR10 and GTSRB, 
particularly with higher poisoning rates.
However, their effectiveness is limited on the ImageNet100 dataset. 
While SAU \cite{sau} demonstrates good defense capabilities, 
it sometimes reduces clean accuracy (ACC) to suboptimal levels. 
We attribute this phenomenon to the lack of theoretical convergence guarantees 
for the unlearning loss function employed by SAU.


\vspace*{1em}
\vspace*{\fill}

\clearpage

\vspace*{\fill}

\vspace*{\fill}

\clearpage

\clearpage
\vspace*{\fill}
\vspace*{\fill}
\clearpage

\vspace*{\fill}

\vspace*{\fill}
\clearpage

\vspace*{\fill}
\input{tab/cifar10.tex}
\vspace*{\fill}
\clearpage

\vspace*{\fill}
\input{tab/imagenet.tex}
\vspace{2em}
\input{tab/gtsrb.tex}
\vspace*{\fill}
\clearpage

\newpage
\section{Additional Results for Self-supervised Learning}
\label{additional_ssl_results}

\vspace*{-0.5em}
\subsection{Self-supervised Learning with SimCLR}
\begin{figure*}[h]
    \centering
    \includegraphics[width=0.95\linewidth]{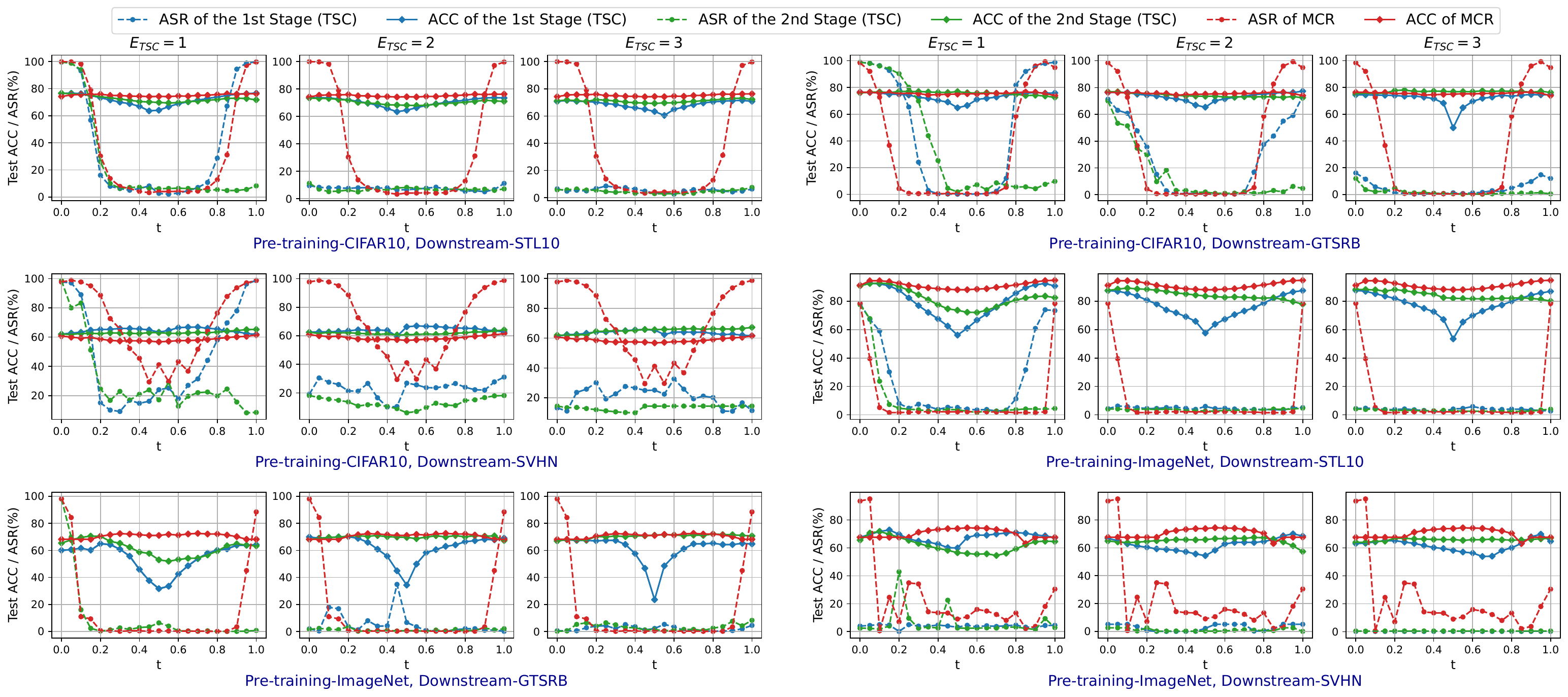}
    \vspace*{-0.5em}
    \caption{
        \textbf{(Self-supervised Learning, SimCLR)} Performance comparison of \shortname\ and MCR against BadEncoder \cite{BadEncoder} attacks using CIFAR10 and ImageNet as pre-training datasets. Model checkpoints are selected at $t=0.25$ for each stage and round.
    }
       \label{fig:curve_acc_simclr}
\end{figure*}

\vspace*{-0.5em}
\cref{fig:curve_acc_simclr} illustrates the performance of \shortname\ and MCR against BadEncoder \cite{BadEncoder} on CIFAR10 and ImageNet using SimCLR \cite{simCLR}. 
\cref{tab:simclr_defense_full} presents the corresponding ultimate defense results of \shortname\ and other baseline defenses. 
\cref{tab:simclr_defense_full} shows the defense results against CTRL \cite{embarrassing} on CIFAR10 and ImageNet100.

Notably, for the BadEncoder attack, 
we utilize publicly available backdoored model checkpoints \footnote{https://github.com/jinyuan-jia/BadEncoder}
as the original model, which was trained on ImageNet 
containing 1,000 classes with image dimensions of $224 \times 224$. 
As the original model for the CTRL attack is not publicly available, 
we follow the settings described in the original paper and train the corresponding encoder on ImageNet-100. 
This version of ImageNet-100 contains 100 classes. 
During training, we scale the image dimensions to $64 \times 64$.
Furthermore, we adopt the evaluation methodology from the BadEncoder paper, which employs linear probe evaluation on downstream tasks. 
For the CTRL attack, 
we also follow the evaluation methods from the original paper: in addition to using linear probe, 
we employ the K-Nearest Neighbor (KNN) method to evaluate ACC and ASR on the pre-training dataset.

In defending against BadEncoder attacks, 
we observe that MCR, SSL-Cleanse \cite{SSL_Cleanse}, 
and \shortname\ all successfully reduce the ASR values. 
However, MCR fails to counter backdoor attacks targeting SVHN downstream task. 
SSL-Cleanse and \shortname\ show similar defense effectiveness, 
successfully defending against BadEncoder attacks while maintaining high ACC values on downstream tasks. 
For CTRL attacks, MCR is only effective against STL10 dataset. 
In contrast, both SSL-Cleanse and \shortname\ 
consistently maintain strong defense performance.

\vspace*{-0.5em}
\subsection{Self-supervised Learning with CLIP}
\vspace*{-0.5em}

\cref{tab:clip_defense_full} presents the defense results of \shortname\ and MCR against BadEncoder under CLIP \cite{clip}.
Since the unlearning algorithm used in SSL-Cleanse defense is based on the simCLR training method, 
and the original paper of SSL-Cleanse did not conduct experiments in the CLIP learning scenario, 
we do not test the defense performance of SSL-Cleanse 
against BadEncoder in this work.

We test on downstream datasets (STL10, GTSRB, CIFAR10, Food101 \cite{food101}, and Pascal VOC 2007 \cite{pascal-voc-2007}) 
and report ACC and ASR using zero-shot a
nd linear probe methods. 
\shortname\ maintains strong defense performance against BadEncoder, 
though it shows lower ACC on GTSRB and CIFAR10.
In contrast, it outperforms the backdoored model on Pascal VOC 2007.
This may be due to the use of MS-COCO \cite{MSCOCO} for post-training. 
MS-COCO shares more similarity with Pascal VOC 2007, while its images differ from those in GTSRB and CIFAR10, leading the CLIP model to ``forget'' features for the latter datasets.

\begin{table}[!ht]
    \centering
    \begin{minipage}[c]{0.8\linewidth}
    \input{tab/simCLR_full.tex}

    \end{minipage}
\end{table}
\newpage

\begin{table}[!ht]
    \centering
    \begin{minipage}[c]{0.8\linewidth}
    \vspace*{0.5em}
    \input{tab/simCLR_ctrl.tex}

    \end{minipage}
\end{table}

\begin{table}[!ht]
    \centering
    \begin{minipage}[c]{0.7\linewidth}
    \vspace*{0.5em}
    \input{tab/CLIP_full.tex}

    \end{minipage}
\end{table}


\vspace{1em}
\subsection{Evaluation of I-BAU and SAU in Self-supervised Learning}
Previously, we have presented the defense results of \shortname\ against BadEncoder and CTRL attacks.
Our defense, as shown in \cref{fig:overview}, specifically targets self-supervised learning (SSL) scenarios by directly purifying the encoder.
The other SSL defenses we evaluated, such as MCR and SSL-Cleanse, also follow this workflow.
This design enables \shortname\ to be effectively applied to zero-shot scenarios, such as CLIP, where neither a linear classifier nor fine-tuning is required.

It might be noted that other defenses designed for supervised learning (SL) settings, such as I-BAU and SAU, could also be applied to the combined encoder and linear classifier after fine-tuning.
Initially, we excluded those defenses from SSL comparisons to maintain fairness and methodological consistency. 
To address this, we conducted additional experiments to evaluate the performance of I-BAU and SAU against BadEncoder with 5\% clean downstream training data.
The other defense settings are consistent with the SL scenarios involving models trained on CIFAR10 and ImageNet100.
\cref{tab:simclr_sau} presents the corresponding results.

It's clear that while I-BAU and SAU reduce the ASR, they significantly degrade benign accuracy (ACC). 
For instance, on CIFAR10-STL10 settings, the ACC dropped from 76.73\% to 30.13\% with I-BAU and further to 21.52\% with SAU. 
We suspect this decline occurs because I-BAU and SAU employ post-training methods analogous to adversarial training in SL, potentially harming the representation extraction capability of encoders trained via SSL methods like SimCLR.
As these specific findings extend beyond the primary scope of this paper, we reserve a more extensive exploration for future work.

\begin{table}[!ht]
    \centering
    \begin{minipage}[c]{0.6\linewidth}
    \vspace*{0.5em}
    \input{tab/simCLR_sau.tex}

    \end{minipage}
\end{table}

\section{Clean Accuracy Drops for Non-backdoored Models}
\label{appendix:clean_acc_drop}
To illustrate the impact of \shortname\ on non-backdoored models,
in this section, we provide additional results on the clean accuracy (ACC) drops for no poison scenarios when applying \shortname\ and other baseline defenses.
\cref{tab:sl_clean,tab:ssl_clean} present the clean ACC drops of non-backdoored models under supervised learning and self-supervised learning (SimCLR), respectively.

Except for settings specific to the attacks, the experimental configurations are consistent with those detailed in previous tables.
While the ACC drops for \shortname\ are not the lowest among all defenses, they are considered acceptable.
Furthermore, the reason why NC always achieves the same ACC as the original model is that NC would check if the model is backdoored before applying it's corresponding unlearning method.
The intermediate results of \cref{tab:sl_clean} show that NC successfully classifies the model as non-backdoored, thus it does not apply any unlearning method. 

\begin{table}[!ht]
    \centering
    \begin{minipage}[c]{0.72\linewidth}
    \input{tab/sl_clean.tex}
    \end{minipage}
\end{table}

\begin{table}[!ht]
    \centering
    \begin{minipage}[c]{0.65\linewidth}
    \vspace*{0.5em}
    \input{tab/ssl_clean.tex}
    \end{minipage}
\end{table}

\newpage

\section{Experimental Setup}
\label{appendix:Experimental_Detials}

Our deep learning training algorithm is implemented using PyTorch.
All experiments were run on one Ubuntu 18.04 server equipped with four NVIDIA RTX V100 GPUs.
Our implementation for supervised learning is mainly based on BackdoorBench \footnote{https://github.com/SCLBD/BackdoorBench/} \cite{backdoorbench}.
For self-supervised learning, we use the official implementation of BadEncoder \footnote{https://github.com/jinyuan-jia/BadEncoder} \cite{BadEncoder}
and CTRL \footnote{https://github.com/meet-cjli/CTRL} \cite{embarrassing} to conduct attacks and defenses experiments.

\subsection{Attack Settings}
\input{fig_tex/attack_lf.tex}

(1) \textbf{Supervised Learning.}
We employed the stochastic gradient descent (SGD) optimization method with a batch size of 256. 
We set the initial learning rate to 0.01 and decayed it using the cosine annealing strategy \cite{SGDR}. 
For the CIFAR-10 and GTSRB datasets, we trained for a total of 100 epochs. 
In the case of the ImageNet100 dataset, we trained for 200 epochs.

Generally, on the CIFAR10, ImageNet100 and GTSRB datasets, the target label for all backdoor attacks is class 0, corresponding to the specific class names ``airplane'', ``tench'' and ``Speed limit 20km/h''. 
The CIFAR10 and GTSRB images are resized to $32 \times 32$, while the ImageNet100 images are resized to $224 \times 224$.

The trigger patterns for BadNet \cite{badnets}, Blended \cite{TargetGlasses}, SSBA \cite{ISSBA}, LF \cite{lf}, WaNet \cite{wanet}, Inputaware \cite{Inputaware}, LC \cite{firstCleanLabel}, SIG \cite{SIG}, SBL \cite{sbl}, Narcissus \cite{narcissus} and SAPA \cite{sapa} are shown in \cref{attack_samples_imagenet}.
BadNet, Blended, SSBA, LF, WaNet, Inputaware, SBL and SAPA are label-flipping attacks, which turns the original label of the poisoned data to the target label.
LC, SIG and Narcissus are clean label attacks, which utilize the data beloning to the target class to generate poisoned samples.
The implementation details for each backdoor attack are as follows:
\begin{itemize}
    \item BadNets attack \cite{badnets} employs a white square placed at the bottom-right corner as the trigger pattern.
    \item Blended attack \cite{TargetGlasses} poisons the data by introducing a Hello Kitty image trigger. We implement the blended injection strategy, denoted as $\alpha t + (1 - \alpha)x$, to incorporate the trigger $t$ into the benign sample $x$ with a value of $\alpha=0.2$.
    \item SSBA \cite{ISSBA} utilizes the StegaStemp algorithm \cite{Stega} to generate specific triggers for poison samples across various classes.
    \item LF \cite{lf} employs frequency domain analysis and optimization algorithms to create poison samples.
    \item InputAware \cite{Inputaware} attack requires the attacker to control the entire training process. During training, the attacker not only trains the model but also trains a generator to produce unique triggers for different samples. The generator continuously optimizes the trigger design while minimizing its size, ensuring a high attack success rate.
    \item WaNet \cite{wanet} uses image embedding to generate invisible triggers for poisoned samples. To enhance attack robustness, WaNet introduces Gaussian noise to poisoned samples with a certain probability during training and restores the original labels of these poisoned samples.
    \item LC attack \cite{firstCleanLabel} utilizes a checkboard pattern positioned in the four corners as the trigger. 
    To establish a link between the trigger and the target label, LC attacks initially employ the Projection Gradient Descent (PGD) method to introduce adversarial perturbations to the images before incorporating the trigger.
    \item SIG attack \cite{SIG} utilizes a sinusoidal signal that is seamlessly integrated into the image as the trigger.
    \item SBL-BadNet and SBL-Blended attacks \cite{sbl} employ continual learning algorithm to fine-tune poisoned models to generate backdoors that are resilient against previous fine-tuning defenses. We used EWC \cite{ewc} as the continual learning algorithm, with BadNet and Blended as base attacks.
    \item Narcissus attack \cite{narcissus} optimizes a universal trigger pattern based a the surrogate (target) model. For fair comparison, we used the open-source trigger for CIFAR10 from \citet{narcissus}, with target label ``bird''.
    \item SAPA attack \cite{sapa} combines sharpness-aware minimization \cite{sharpnessaware} with the Sleeper-Agent \cite{sleepagent} backdoor attack to smooth the poison loss landscape.
    We utilized the colorful patch from \cite{sleepagent} as the trigger.
For CIFAR10 and GTSRB, we used an $8 \times 8$ pixel patch; for ImageNet100, a $16 \times 16$ pixel patch was employed.
Other recommended parameters for generating the poisoned samples were adopted from the SAPA and Sleeper-Agent. These included: sharpness sigma = 0.01, number of source samples = 1000, $R=250$ optimization steps, $T=4$ retraining periods, and an $L_\infty$-norm perturbation bound of $16/255$.

\end{itemize}

(2) \textbf{Self-Supervised Learning.}

\begin{figure}[h]
    \centering
    \begin{subfigure}[BadEncoder\label{fig:sample_badencoer_stl10}]{
        \includegraphics[width=0.42\linewidth]{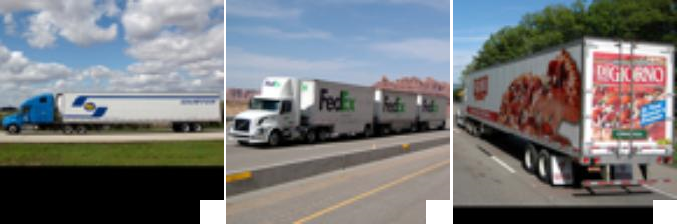}
    }    \end{subfigure}
    \begin{subfigure}[CTRL\label{fig:sample_ctrl_stl10}]{
        \includegraphics[width=0.42\linewidth]{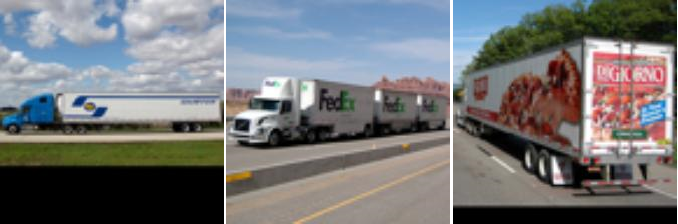}
    }\end{subfigure}
    \caption{Examples of 2 self-supervised learning backdoor trigger patterns on STL10.}
    \label{attack_samples_stl10_ssl}
\end{figure}
    
    We use publicly available backdoored ResNet18 and ResNet50 models as encoders with SimCLR in the BadEncoder attack \cite{BadEncoder}. 
    For CLIP, following the BadEncoder approach, 
    we fine-tune a pre-trained CLIP ResNet50 on ImageNet100 using SimCLR for 200 epochs to inject backdoors.
    For the CTRL \cite{embarrassing} attack with SimCLR, we apply the cosine annealing strategy for the learning rate. 
    For the CIFAR10 dataset under CTRL, 
    we set the initial learning rate to 0.06 and run for 800 epochs. 
    For the ImageNet100 dataset under CTRL, 
    the initial learning rate is set to 0.02 with 1000 training epochs.
    
    \cref{tab:badencoder_settings,tab:ctrl_settings} show the dataset and target class settings for BadEncoder and CTRL in different learning scenarios. 
    The post-training dataset refers to the fine-tuning dataset used by each defense method. 
    When the downstream dataset lacks the target class, 
    we add the corresponding class and include target class data from the pre-training dataset to evaluate ACC and ASR. 
    To align with the CTRL paper’s settings, 
    we use a $64 \times 64$ ImageNet100 dataset as the pre-training dataset.

\begin{table}[th]

    \caption{
        Dataset and target class settings for BadEncoder attack.
        If the ``Requiring extra data'' column is marked as `\Checkmark',
        it indicates that the downstream dataset lacks the target class and we manually add the target class and data to the dataset for evaluating ACC and ASR.
        }
        \centering
        \renewcommand{\arraystretch}{1}
    \resizebox{0.98\linewidth}{!}{
    \begin{tabular}{ccccccc}
        \toprule
    Method & Image Size & Pre-training & Downstream        & Post-training & Target class       & Requiring extra data \\
    \hline
    simCLR & 32$\times$32      & CIFAR10      & STL10             & CIFAR10       & truck              & \XSolidBrush                   \\
    simCLR & 32$\times$32      & CIFAR10      & GTSRB             & CIFAR10       & priority road sign & \XSolidBrush                   \\
    simCLR & 32$\times$32      & CIFAR10      & SVHN              & CIFAR10       & 1                  & \XSolidBrush                   \\
    simCLR & 224$\times$224    & ImageNet     & STL10             & ImageNet100   & truck              & \XSolidBrush                   \\
    simCLR & 224$\times$224    & ImageNet     & GTSRB             & ImageNet100   & priority road sign & \XSolidBrush                   \\
    simCLR & 224$\times$224    & ImageNet     & SVHN              & ImageNet100   & 1                  & \XSolidBrush                   \\
    CLIP   & 224$\times$224    & CLIP         & STL10             & MS-COCO   & truck              & \XSolidBrush                   \\
    CLIP   & 224$\times$224    & CLIP         & GTSRB             & MS-COCO   & stop sign          & \XSolidBrush                   \\
    CLIP   & 224$\times$224    & CLIP         & SVHN              & MS-COCO   & 0                  & \XSolidBrush                   \\
    CLIP   & 224$\times$224    & CLIP         & Food101, VOC 2007 & MS-COCO   & truck              & \Checkmark                 \\
    \bottomrule

    \end{tabular}
    }\label{tab:badencoder_settings}
    \end{table}

    \begin{table}[ht]
        \centering
            \caption{
                Dataset and target class settings for CTRL attack.
                }
                \centering
                \renewcommand{\arraystretch}{1}
            \resizebox{0.94\linewidth}{!}{
            \begin{tabular}{ccccccc}
                \toprule
            Method & Image Size & Pre-training & Downstream        & Post-training & Target class       & Requiring extra data \\
            \hline
            simCLR                   & 32$\times$32                       & CIFAR10                  & STL10                                 & CIFAR10                         & airplane                      & \XSolidBrush                      \\
            simCLR                   & 32$\times$32                       & CIFAR10                  & GTSRB                                 & CIFAR10                         & airplane                      & \Checkmark                     \\
            simCLR                   & 32$\times$32                       & CIFAR10                  & SVHN                                  & CIFAR10                         & airplane                      & \Checkmark                     \\
            simCLR                   & 64$\times$64                       & ImageNet100              & STL10                                 & ImageNet100                     & tench                         & \XSolidBrush                      \\
            simCLR                   & 64$\times$64                       & ImageNet100              & GTSRB                                 & ImageNet100                     & tench                         & \Checkmark                     \\
            simCLR                   & 64$\times$64                       & ImageNet100              & SVHN                                  & ImageNet100                     & tench                         & \Checkmark                     \\             
    \bottomrule
        
            \end{tabular}
            }    \label{tab:ctrl_settings}
            \end{table}
    
The implementation details for BadEncoder and CTRL attacks are as follows:
\begin{itemize}
    \item BadEncoder attack \cite{BadEncoder}
    requires the attacker to control the training process. During training, the attacker uses the SimCLR loss function to optimize the similarity between poisoned samples and target class images. Similar to BadNets, we use a white square in the bottom-right corner as the trigger. The loss function assigns equal weight to clean and poisoned samples, i.e., $\lambda_1 = \lambda_2 = 1$.

    \item CTRL attack \cite{embarrassing}
    uses a invisible frequency trigger to generate poisoned samples.
     Following the recommended settings, 
     we set the poisoning rate to 0.5\% and the trigger window size to 32. 
     For CIFAR10, the trigger magnitude is set to 100. 
     For ImageNet100, we use a magnitude of 100 for the first 700 epochs and a magnitude of 200 for the remaining 300 epochs.

\end{itemize}

\subsection{Defense Settings}
We allocate 5\% of the clean training samples from each dataset to the defender.
For all other settings not specified below,
we follow the default settings outlined in their publications or public implementation.

Since we use the original mathematical symbols from each publication,
please note that some symbols and terms may conflict with each other.
All symbols and terms below are specific to the parameters in their respective papers.

(1) \textbf{Supervised Learning.}
\begin{itemize}
    \item FP \cite{FinePruning}:
    In the experiment, we use SGD as the base fine-tuning method with a learning rate of 0.01. The pruning rate for all models is set to 1\%, and the number of fine-tuning epochs is 100.
    
    \item NC \cite{NeuralCleanse}: 
    We set the threshold of the Anomaly Index at 2 for all the datasets. 
    For models with an Anomaly Index higher than 2 (marked as attacked), 
    we conduct the unlearning procedure for 40 epochs, 
    utilizing 5\% of the training data and applying the reversed trigger to 20\% of these samples.

    \item ANP \cite{ANP}:
    We set the learning rate to 0.2 for optimizing the neuron mask with SGD. 
    After optimization, neurons with a mask value smaller than the threshold of 0.2 are pruned. 
    Additionally, we set the tradeoff coefficient $\alpha$ to 0.2 and the perturbation budget $\epsilon$ to 0.4 for a total of 2000 iterations.

    \item FT-SAM \cite{FT_SAM}:
    SGD is used as the base fine-tuning method with a learning rate of 0.01. The label smoothing rate is set to 0.1, and the SAM (Sharpness-aware Minimization) \cite{sharpnessaware} optimization method uses a Neighborhood size of $\rho = 0.2$. The number of fine-tuning epochs is 100.
    
    \item I-BAU \cite{I_BAU}:
    We follow the default settings provided in the public implementation of I-BAU. 
    We use Adam with a learning rate of $1e^{-3}$ to optimize the outer loop. 
    And, we set the maximum number of unlearning rounds to 5 and the maximum number of fixed-point iterations to 5.

    \item SAU \cite{sau}: 
    We iterate 5 times to compute the adversarial perturbation, using an $\ell_\infty$ norm with perturbation bound $0.2$ and adversarial learning rate $0.2$. The total epochs for CIFAR10 and GTSRB are 100, and for ImageNet100, 50 epochs. We set the unlearning weights as $\lambda_1 = \lambda_2 = \lambda_4 = 1$ and $\lambda_3 = 0.1$.
    
    \item TSC (ours):
    We set the global number of epochs $E_{\shortname} = 3$, curve index $t = 0.4$, and curve training epochs $e = 200$. The initial learning rate is set to $0.02$, with cosine annealing used for post-training.
    After every round of \shortname, we slightly fine-tune the model extracted from the second stage for $5$ epochs with a learning rate of $1e^{-4}$.

    \item MCR \cite{MCR}:
    For a fair comparison with \shortname, we set the curve index $t = 0.4$ and curve training epochs $e = 200$. The initial learning rate is $0.02$, and cosine annealing is used for post-training.

\end{itemize}

(2) \textbf{Self-Supervised Learning.}
\begin{itemize}        

    \item SSL-Cleanse \cite{SSL_Cleanse}:
    This method is specifically designed for defending against backdoors in self-supervised learning. 
    Similar to NC, 
    it first attempts to restore the trigger pattern and then 
    applies an unlearning method to remove the backdoor. 
    We set the initial value of $\lambda$ to 0.01 and 
    use a learning rate of 0.01.
    
    \item TSC (ours):
    We set the global number of epochs $E_{\shortname} = 2$ 
    and the curve index $t = 0.25$. 
    All learning rates use a cosine annealing strategy for iteration.
    (1) For SimCLR, we set the batch size to 256, 
    the curve training epochs $e = 200$, 
    and use Adam with an initial learning rate of $5e^{-3}$ 
    for CIFAR10 and $2e^{-3}$ for ImageNet100. 
    After every round of \shortname, we slightly fine-tune the model extracted from the second stage for $5$ epochs with a learning rate of $5e^{-5}$.
    (2) For CLIP, we follow the training method proposed by \citet{clip}, 
    setting the batch size to 32768 and using Adam with an initial learning rate of $1e^{-4}$ ($\beta_1 = 0.9$, $\beta_2 = 0.999$). 
    For such a large batch size, 
    we use gradient accumulation to update the model, 
    with each stage training for only 2 epochs.
    After every round of \shortname, we do not fine-tune the model extracted from the second stage.

    \item MCR \cite{MCR}:
    We set the curve index $t = 0.25$. All other parameters are the same as those used in TSC.

\end{itemize}

%% file: mathematics_proof/proof_1.tex
\begin{proof}
    For clarity, we first introduce some notation  about $\tilde{\bm{\theta}}_{A,B}(t)$.
    These notations can also be applied to $\tilde{\bm{\theta}}_{A,C}(t)$ 
    by substituting appropriate symbols. 

    We use $\bm{\theta}_{A,B}(t)$ to denote the point sampled from the curve $\bm{\gamma}_{\tilde{\bm{\theta}}_{A,B}}(t)$.
    Meanwhile, we consider a dataset containing a single sample  $\bm{x}_0$. 
    Then, $\forall l \in \{1, 2, \ldots, L\}$, the condition can be reformulated as:
    \begin{equation}
        \label{feature_condition}
        M_l(\bm{\theta}_A, \bm{\theta}_B; \bm{x}_0) \leq M_l(\bm{\theta}_A, \bm{\theta}_C; \bm{x}_0).
    \end{equation}
    Moreover, we use $\tilde{\bm{W}}^{A,B}_l$ to indicate the weight of the $l$\textsuperscript{th} layer of $\tilde{\bm{\theta}}_{A,B}$.  
    This allows us to construct a feedforward network respect to $\bm{\theta}_{A,B}(t)$:
    \begin{align}
         \bm{f}(\bm{x}_0,\bm{\theta}_{A,B}(t))\nonumber = & \left( (1-t)\bm{W}^{A}_{L}+t\bm{W}^{B}_{L}+2t(1-t)\tilde{\bm{W}}^{A,B}_L \right) \sigma \left( (1-t)\bm{W}^{A}_{L-1}+t\bm{W}^{B}_{L-1}+2t(1-t) \tilde{\bm{W}}^{A,B}_{L-1} \right) \\
         & \ldots \sigma \left( (1-t)\bm{W}^{A}_{1}+t\bm{W}^{B}_{1}+2t(1-t)\tilde{\bm{W}}^{A,B}_1 \right) \bm{x}_0.
    \end{align}
    Then, we define the pre-activation and post-activation for each layer as follows:
    \begin{align}
        \label{pre_activation}
        \bm{x}_1^{A,B}(t) = &\left( (1-t)\bm{W}^{A}_{1}+t\bm{W}^{B}_{1}+2t(1-t)\tilde{\bm{W}}^{A,B}_1 \right) \bm{x}_0, \\
        \bm{x}_l^{A,B}(t) = &\Bigr(\; (1-t)\bm{W}^{A}_{l-1}\; +\; t\bm{W}^{B}_{l-1} +\; 2t(1-t)\tilde{\bm{W}}^{A,B}_{l-1} \; \Bigl)\ \sigma \ \bm{x}_{l-1}^{A,B}(t).
    \end{align}

    Now, we consider the $L_2$ norm distance between $\bm{x}_1^{A,B}(t)$ and the endpoints
    defining the following distances:
    \begin{align}
        d^{A,B}_{1}(t,0)
        &= \| \bm{x}_1^{A,B}(t)-\bm{x}_1^{A,B}(0) \|  \nonumber \\[0.25em] 
        &= \Vert t(\bm{x}_1^{A,B}(1)-\bm{x}_1^{A,B}(0)) \ + 2t(1-t)\tilde{\bm{W}}^{A,B}_1\bm{x}_0 \Vert,
    \end{align}
    \begin{align}
        d^{A,B}_{1}(t,1)
        &= \| \bm{x}_1^{A,B}(t)-\bm{x}_1^{A,B}(1) \|  \nonumber \\[0.25em] 
        &= \Vert (1-t)(\bm{x}_1^{A,B}(1)-\bm{x}_1^{A,B}(0)) + 2t(1-t)\tilde{\bm{W}}^{A,B}_1\bm{x}_0 \Vert,
    \end{align}
    Then, we can use Triangle Inequality and Cauchy-Schwarz Inequality 
    to get an upper bound $U(d^{A,B}_{1}(t,0))$ for $d^{A,B}_{1}(t,0)$:
    \begin{align}
        d^{A,B}_{1}(t,0) \le t&\Vert \bm{x}_1^{A,B}(1)-\bm{x}_1^{A,B}(0) \Vert + 2t(1-t) \Vert \tilde{\bm{W}}^{A,B}_1\bm{x}_0 \Vert  \ &(\text{Triangle Inequality}) \nonumber\\[0.25em]
        \le t&\Vert \bm{x}_1^{A,B}(1)-\bm{x}_1^{A,B}(0) \Vert + 2t(1-t) \Vert \tilde{\bm{W}}^{A,B}_1\Vert \cdot \Vert\bm{x}_0 \Vert  \ &(\text{Cauchy-Schwarz Inequality}) \nonumber \\[0.25em]
        = t&M_1(\bm{\theta}_A, \bm{\theta}_B; \bm{x}_0) + 2t(1-t)\Vert \tilde{\bm{W}}^{A,B}_1\Vert \cdot \Vert\bm{x}_0\Vert\ \nonumber\\[0.25em]
        =\ \; &U(d^{A,B}_{1}(t,0))\ .
    \end{align}
    Similarly, we can get
    \begin{align}
        U(d^{A,B}_{1}(t,1))=&(1-t)M_1(\bm{\theta}_A, \bm{\theta}_B; \bm{x}_0) + 2t(1-t)\Vert \tilde{\bm{W}}^{A,B}_1\Vert \cdot \Vert\bm{x}_0\Vert,
    \end{align}
    \begin{align}
        U(d^{A,C}_{1}(t,0))=&tM_1(\bm{\theta}_A, \bm{\theta}_C; \bm{x}_0)+ 2t(1-t)\Vert \tilde{\bm{W}}^{A,C}_1\Vert \cdot \Vert\bm{x}_0\Vert,
    \end{align}
    \begin{align}
        U(d^{A,C}_{1}(t,1))=&(1-t)M_1(\bm{\theta}_A, \bm{\theta}_C; \bm{x}_0) + 2t(1-t)\Vert \tilde{\bm{W}}^{A,C}_1\Vert \cdot \Vert\bm{x}_0\Vert,
    \end{align} 
    Since \( \tilde{\bm{\theta}}_{A,B}\) and \(\tilde{\bm{\theta}}_{A,C}\) are \(L_2\)-norm-consistent, 
    and given that $\bm{\theta}_A$, $\bm{\theta}_B$ and $\bm{\theta}_C$ satisfy the condition (\ref{feature_condition}),
    we can conclude that:
    \begin{align}
        U(d^{A,B}_{1}(t,0))\le U(d^{A,C}_{1}(t,0)),
    \end{align}
    \begin{align}
        U(d^{A,B}_{1}(t,1))\le U(d^{A,C}_{1}(t,1)).
    \end{align}
    We also consider the $L_2$ norm distance $b^{A,B}_{1}(t,0)$ between the post-activation $\sigma \bm{x}_1^{A,B}(t)$ and the endpoints.
    Assuming that $\sigma$ is Lipschitz continuous,
    we have:
    \begin{align}
        b^{A,B}_{1}(t,0) &=\|\sigma \bm{x}_1^{A,B}(t)-\sigma\bm{x}_1^{A,B}(0) \| \nonumber \\[0.25em]
                        &\le L_\sigma \ U(d^{A,B}_{1}(t,0)),
    \end{align}
    where the $L_\sigma$ is the Lipschitz constant of $\sigma$.
    This property holds for other post-activations as well.
    Thus, both pre-activation distance $d^{A,B}_{1}(t,0)$ and post-activation distance $b^{A,B}_{1}(t,0)$ 
    for the first layer of $\bm{\theta}_{A,B}(t)$
    have tighter upper bounds than those for $\bm{\theta}_{A,C}(t)$.

    Let $\bm{z}_1^{A,B}(t) =  \sigma \bm{x}_l^{A,B}(t)$ be the post-activation for the $l$\textsuperscript{th} layer of $\bm{\theta}_{A,B}(t)$.
    We can also derive the the upper bound for the pre-activation distance $d^{A,B}_{l}(t,0)$ for deep layer:
    \begin{align}
        d^{A,B}_{l}(t,0)= \ & \| \bm{x}_l^{A,B}(t)-\bm{x}_l^{A,B}(0) \|  \nonumber \\[0.25em] 
        =\ &\Vert (1-t)\bm{W}^{A}_l (\bm{z}_{l-1}^{A,B}(t)-\bm{z}_{l-1}^{A,B}(0))\nonumber + t\bm{W}^{B}_l (\bm{z}_{l-1}^{A,B}(t)-\bm{z}_{l-1}^{A,B}(1)) \nonumber \\[0.25em]
        &\ + t(\bm{W}^{B}_l\bm{z}_{l-1}^{A,B}(1)-\bm{W}^{A}_l\bm{z}_{l-1}^{A,B}(0)) \nonumber + 2t(1-t)\tilde{\bm{W}}^{A,B}_l(\bm{z}_{l-1}^{A,B}(t)-\bm{z}_{l-1}^{A,B}(0)) \nonumber \\[0.25em]
        &\ + 2t(1-t)\tilde{\bm{W}}^{A,B}_l\bm{z}_{l-1}^{A,B}(0) \Vert  \\[0.5em]
        \le\ &(1-t)\Vert \bm{W}^{A}_l\Vert b^{A,B}_{l-1}(t,0) \nonumber + t\ \Vert\bm{W}^{B}_l\Vert b^{A,B}_{l-1}(t,1) \nonumber + t\ d^{A,B}_{l}(1,0) \nonumber\\[0.25em]
        \ & + 2t(1-t)\Vert \tilde{\bm{W}}^{A,B}_l \Vert b^{A,B}_{l-1}(t,0) \nonumber + 2t(1-t)\Vert \tilde{\bm{W}}^{A,B}_l \Vert \bm{z}_{l-1}^{A}  \\[0.5em]
        \le\ &(1-t)L_\sigma \Vert \bm{W}^{A}_l\Vert U(d^{A,B}_{l-1}(t,0)) \nonumber  + t L_\sigma \ \Vert\bm{W}^{B}_l\Vert U(d^{A,B}_{l-1}(t,1)) \nonumber + t M_l(\bm{\theta}_A, \bm{\theta}_B; \bm{x}_0) \nonumber \\[0.25em] 
        \ & + 2t(1-t) L_\sigma \Vert \tilde{\bm{W}}^{A,B}_l \Vert U(d^{A,B}_{l-1}(t,0)) \nonumber + 2t(1-t)\Vert \tilde{\bm{W}}^{A,B}_l \Vert \bm{z}_{l-1}^{A} \nonumber \\[0.25em]
        &\ = U(d^{A,B}_{l}(t,0))
    \end{align}
    To streamline the proof, we ignore the other upper bounds here.
    Similar to our demonstration for the first layer, we have:
    \begin{align}
        U(d^{A,B}_{l}(t,0))\le U(d^{A,C}_{l}(t,0)),\\
        U(d^{A,B}_{l}(t,1))\le U(d^{A,C}_{l}(t,1)).
    \end{align}
    Finally, since we assume that $\mathcal{L}$ is Lipschitz continuous,
    there exists a constant $L_\mathcal{L}$ such that:
    \begin{align}
        \mathcal{L}(\bm{f}(\bm{x}_0,\bm{\theta}_{A,B}(t))-\bm{y}) &\le L_\mathcal{L} \Vert \bm{f}(\bm{x}_0,\bm{\theta}_{A,B}(t))-\bm{y} \Vert \nonumber \\[0.25em]
        &= L_\mathcal{L} \Vert \bm{x}_L^{A,B}(t) -\bm{y} \Vert \nonumber \\[0.25em]
        &\le L_\mathcal{L} U(\Vert \bm{x}_L^{A,B}(t) -\bm{y} \Vert) \nonumber \\[0.25em]
        & = U_{A,B}(t),
    \end{align}
    where $\bm{y}$ is the ground truth label (or feature).
    As $\bm{\theta}_A$, $\bm{\theta}_B$, and $\bm{\theta}_C$ are three optimal networks,
    we can find a constant $\epsilon$ and specify them as $\epsilon$ optimal networs such that $\Vert \bm{f}(\bm{x}_0,\bm{\theta}_{\alpha}(t))-\bm{y} \Vert$,
    for $\alpha \in \{A,B,C\}$. Then, we have:
    \begin{align}
        \Vert \bm{x}_L^{A,B}(t) -\bm{y} \Vert =\ & \Vert (1-t)\bm{W}^{A}_L (\bm{z}_{L-1}^{A,B}(t)-\bm{z}_{L-1}^{A,B}(0)) \nonumber + t\bm{W}^{B}_L (\bm{z}_{L-1}^{A,B}(t)-\bm{z}_{L-1}^{A,B}(1)) \nonumber \\[0.25em]
        &\ + (1-t)\bm{W}^{A}_L (\bm{z}_{L-1}^{A,B}(0)-\bm{y}) \nonumber + t\bm{W}^{B}_L (\bm{z}_{L-1}^{A,B}(1)-\bm{y}) \nonumber \\[0.25em]
        &\ + 2t(1-t)\tilde{\bm{W}}^{A,B}_L\bm{z}_{L-1}^{A,B}(t) \Vert \\[0.5em]
        \le\ &(1-t)L_\sigma \Vert \bm{W}^{A}_L\Vert U(d^{A,B}_{L-1}(t,0)) \nonumber + t L_\sigma \ \Vert\bm{W}^{B}_L\Vert U(d^{A,B}_{L-1}(t,1))+\epsilon \nonumber \\[0.25em]
        &\ + 2t(1-t) L_\sigma \Vert \tilde{\bm{W}}^{A,B}_L \Vert U(d^{A,B}_{L-1}(t,0)) \nonumber + 2t(1-t)\Vert \tilde{\bm{W}}^{A,B}_L \Vert \bm{z}_{L-1}^{A} \nonumber \\[0.25em]
        &\ = U(\Vert \bm{x}_L^{A,B}(t) -\bm{y} \Vert).
    \end{align}
    Following the above scaling, we can also get $U_{A,C}(t)$
    and derive $U_{A,B}(t) \le U_{A,C}(t)$.
    Thus, for the upper bounds of loss value over the curves $\bm{\gamma}_{\tilde{\bm{\theta}}_{A,B}}(t)$ and $\bm{\gamma}_{\tilde{\bm{\theta}}_{A,C}}(t)$,
    we have $\ell(\tilde{\bm{\theta}}_{A,B}) \le U_{{A,B}}, \ \ell(\tilde{\bm{\theta}}_{A,C}) \le U_{{A,C}}$, where $U_{{A,B}} \le U_{{A,C}}$,
    finishing the proof.
\end{proof}

%% file: mathematics_proof/proof_2.tex
\begin{proof}
    Since $S(\bm{P}')$ and $S(\hat{\bm{P}})$ are the sets consisting of permutation matrices, 
    the Frobenius norms of the weights in each layer of $\bm{\theta}_{\hat{B}}$ and $\bm{\theta}_{B'}$ are identical to those of  $\bm{\theta}_{B}$.
    Thus, $\bm{\theta}_{\hat{B}}$, $\bm{\theta}_{B'}$, and $\bm{\theta}_{B}$ are \(L_2\)-norm-consistent with each other, according to \Cref{weight_consistency}.

    As $S(\bm{\hat{P}})$ and $S(\bm{P}')$ are solutions to the optimization problems  (\ref{l2_minimization}) and (\ref{l2_maximization}), respectively, 
    we derive the following relations for all $l \in \{1, 2, \ldots, L\}$:
    $$
        M_l(\bm{\theta}_A, \bm{\theta}_{\hat{B}}; D) \leq M_l(\bm{\theta}_A, \bm{\theta}_B; D) \leq M_l(\bm{\theta}_A, \bm{\theta}_{B'}; D).
    $$
    By \Cref{general_theorem}, 
    we conclude that $U_{{A,\hat{B}}} \le U_{{A,B}} \le U_{{A,B'}}$, completing the proof.

\end{proof}

%% file: algorithms/permutation_layer.tex
\begin{algorithm}[h]
\caption{\textsc{PermuteLayers} (Compute Permutation Matrices for Layer Alignment/Un-alignment)}
    \label{alg:permutation_layer}
\begin{algorithmic}
    \REQUIRE{model $\bm{\theta}_A$, \ model $\bm{\theta}_B$, \ dataset $D$, \ optimization \textsc{opt};}
    \STATE Set assignment problem $\bm{O}$ with minimization/maximization objective according to \textsc{opt};
    \FOR{each layer $l$ in $\{1, \ 2, \ \dots, L-1\}$}
        \FOR{each sample $\bm{x}_{i, \; 0}$ in dataset $D$}
            \STATE Compute $\bm{x}^A_{i, \; l} \gets \sigma \circ \bm{W}^A_{l} \circ ... \circ \sigma \circ \bm{W}^A_{1} \bm{x}_{i, \; 0}$, \quad $\bm{x}^B_{i, \; l} \gets \sigma \circ \bm{W}^B_{l} \circ ... \circ \sigma \circ \bm{W}^B_{1} \bm{x}_{i, \; 0}$;
        \ENDFOR
        \STATE Compute the correlation matrix $R_l$ using \cref{correlation_matrix};
        \STATE Compute $P_l$ by solving the $\bm{O}$ with $R_l$ using Hungarian algorithm;
        \STATE Update $\bm{W}^B_{l} \gets P_l \bm{W}^B_{l}$, \quad $\bm{W}^B_{l+1} \gets \bm{W}^B_{l+1} P_l^{\top}$;
    \ENDFOR
    \ENSURE{permuted model $\bm{\theta}_B$;}
\end{algorithmic}
\end{algorithm}

%% file: algorithms/curve_training.tex
\begin{algorithm}[h]
    \caption{\textsc{FitQuadCurve} (Train Quadratic Bézier Curve)}
    \label{alg:fit_curve}
 \begin{algorithmic}
    \REQUIRE{model $\bm{\theta}_A$, \ model $\bm{\theta}_B$, \ training method $\mathcal{F}$, \ dataset $D$, \ curve training epoch $e$;}
    \STATE Initialize \; $\bm{\theta}_{A,B} \gets \frac{1}{2}(\bm{\theta}_{A}+\bm{\theta}_{B})$;
    \STATE Initialize parametric curve \; $\bm{\gamma}_{\bm{\theta}_{A, B}}(t) \gets (1 -t)^2  \bm{\theta}_{A} + 2t(1-t) \bm{\theta}_{A, B} + t^2  \bm{\theta}_{B}$;
    \FOR{$i=1$ {\bfseries to} $e$}
    \STATE Sample $\hat t$ from the distribution $U(0,1)$;
    \STATE $\bm{\theta}_{\hat t} \gets \textsc{RetrievePoint}(\bm{\gamma}_{\bm{\theta}_{A, B}} \ ,\ \hat t)$;
    \STATE Update $\bm{\theta}_{A, B}$, the weights of the curve $\bm{\gamma}_{\bm{\theta}_{A, B}}(t)$, using the loss computed by $\mathcal{F}$ with respect to $\bm{\theta}_{\hat t}$ over the dataset $D$;
    \ENDFOR
    \ENSURE{quadratic Bézier curve $\bm{\gamma}_{\bm{\theta}_{A, B}}$;}
 \end{algorithmic}
\end{algorithm}

%% file: algorithms/adaptive_learning_subspace.tex
\begin{algorithm}[h]
    \caption{Adaptive Attack against \shortname}
    \label{alg:adaptive}
 \begin{algorithmic}
    \REQUIRE{backdoored model $\bm{\theta}_{adv}$,\ training method $\mathcal{F}$,\ poisoned dataset $D_{adv}$,\ curve training epoch $e$;}
   \STATE $\vartriangleright $ \; Projecting backdoored model to symmetric subspace
   \STATE $\bm{\theta}_{adv'} \gets \textsc{PermuteLayers}(\bm{\theta}_{adv}, \bm{\theta}_{adv}, D_{adv}, \textsc{max})$;
    \STATE Initialize \; $\bm{\theta}_{m-adv} \gets \frac{1}{2}(\bm{\theta}_{adv}+\bm{\theta}_{adv'})$;
    \STATE Initialize parametric curve \; $\bm{\gamma}_{\bm{\theta}_{adv}}(t) \gets (1 -t)^2  \bm{\theta}_{adv} + 2t(1-t) \bm{\theta}_{m-adv} + t^2  \bm{\theta}_{adv'}$;
   \STATE $\vartriangleright $ \; Learning Symmetric Backdoored Subspace
    \FOR{$i=1$ {\bfseries to} $e$}
    \STATE Sample $\hat t$ from the distribution $U(0,1)$;
    \STATE $\bm{\theta}_{\hat t,adv} \gets \textsc{RetrievePoint}(\bm{\gamma}_{\bm{\theta}_{adv}} \ ,\ \hat t)$;
    \STATE Update the weights of $\bm{\theta}_{adv}$, $\bm{\theta}_{m-adv}$ and $\bm{\theta}_{adv'}$ simultaneously, using the loss computed by $\mathcal{F}$ with respect to $\bm{\theta}_{\hat t,adv}$ over the dataset $D_{adv}$;
    \ENDFOR
    \ENSURE{backdoored model $\bm{\theta}_{adv}$;}
 \end{algorithmic}
\end{algorithm}

%% file: tab/adap_cifar10_sl.tex
    \caption{
        Performance of \shortname\ against adaptive attacks on \textbf{CIFAR-10} under \textbf{supervised learning}.
        } 
    \vspace*{0.1em}
    \centering 
    \renewcommand{\arraystretch}{1.25}
    \resizebox{1\linewidth}{!}{
        \begin{tabular}{ccccccc}
            \toprule
            CIFAR10& & \multicolumn{2}{c}{No Defense} & & \multicolumn{2}{c}{\shortname} \\
            \cline{3-4} \cline{6-7}
            (Poisoning rate-5\%) &
            &ACC($\uparrow$) & ASR($\downarrow$) & & ACC($\uparrow$) & ASR($\downarrow$) \\[2pt]
            \hline
            Adap-BadNet	&	&92.43	&\light 91.26	&	&91.47	&\grc 2.07	\\
            Adap-Blended	&	&93.12	&\light 99.23	&	&90.88	&\grc 7.26	\\
            Adap-Inputaware	&	&92.11	&\light 90.18	&	&90.21	&\grc 4.85	\\
            Adap-LC	&	&93.27	&\light 98.14	&	&89.87	&\grc 2.92	\\
            Adap-LF	&	&92.46	&\light 97.91	&	&88.66	&\grc 3.39	\\
            Adap-SIG	&	&92.33	&\light 94.99	&	&90.34	&\grc 1.82	\\
            Adap-SSBA	&	&92.39	&\light 94.87	&	&89.33	&\grc 2.68	\\
            Adap-WaNet	&	&92.61	&\light 87.29	&	&90.19	&\grc 1.59	\\            
            \bottomrule

        \end{tabular}
        }
    \label{tab:adap_cifar10_sl}
    \vspace*{-0.5em}

%% file: tab/adap_in100_sl.tex
    \caption{
        Performance of \shortname\ against adaptive attacks on \textbf{ImageNet100} under \textbf{supervised learning}. 
  } 
    \vspace*{0.1em}
    \centering 
    \renewcommand{\arraystretch}{1.25}
    \resizebox{1\linewidth}{!}{
        \begin{tabular}{ccccccc}
            \toprule
            ImageNet100& & \multicolumn{2}{c}{No Defense} & & \multicolumn{2}{c}{\shortname} \\
            \cline{3-4} \cline{6-7}
            (Poisoning rate-1\%) &
            &ACC($\uparrow$) & ASR($\downarrow$) & & ACC($\uparrow$) & ASR($\downarrow$) \\[2pt]
            \hline
            Adap-BadNet	&	&83.64	&\light 99.47	&	&78.22	&\grc 0.23	\\
            Adap-Blended	&	&84.42	&\light 98.15	&	&78.85	&\grc 9.61	\\
            Adap-Inputaware	&	&79.42	&\light 70.44	&	&77.04	&\grc 0.82	\\
            Adap-LC	&	&84.22	&\light 42.67	&	&80.44	&\grc 0.36	\\
            Adap-LF	&	&83.61	&\light 99.66	&	&78.18	&\grc 3.26	\\
            Adap-SIG	&	&83.39	&\light 69.86	&	&79.55	&\grc 5.61	\\
            Adap-SSBA	&	&83.13	&\light 99.32	&	&80.19	&\grc 7.67	\\
            Adap-WaNet	&	&81.75	&\light 88.04	&	&80.31	&\grc 0.26	\\      
            \bottomrule
        \end{tabular}
        }
    \label{tab:adap_in100_sl}
    \vspace*{-0.5em}

%% file: tab/adap_simclr_ssl.tex
\caption{
        Performance of \shortname\ against adaptive attacks using \textbf{SimCLR} with CIFAR-10 and ImageNet100 pretraining (\textbf{self-supervised learning}).
        } 
    \centering 
    \renewcommand{\arraystretch}{1.25}
    \resizebox{1\linewidth}{!}{
        \begin{tabular}{cccccccc}
            \toprule
            \multirow{2}*{\shortstack{Pre-training\\Dataset}} &
            \multirow{2}*{\shortstack{Downstream\\Dataset}} & &
            \multicolumn{2}{c}{No Defense} & &
            \multicolumn{2}{c}{TSC (ours)} \\[2pt]
            \cline{4-5} \cline{7-8}
            & & &  ACC($\uparrow$) & ASR($\downarrow$) & & ACC($\uparrow$) & ASR($\downarrow$) \\[2pt]
            \hline
            \multirow{3}*{\shortstack{CIFAR10}}	&STL10	&	&76.51	&\light 99.80	&	&72.62	&\grc 4.83	\\
            &GTSRB	&	&82.47	&\light 98.94	&	&77.77	&\grc 1.90	\\
            &SVHN	&	&64.90	&\light 98.74	&	&64.41	&\grc 5.13	\\ \hline
        \multirow{3}*{\shortstack{ImageNet}}	&STL10	&	&94.94	&\light 98.87	&	&88.24	&\grc 3.42	\\
            &GTSRB	&	&75.81	&\light 99.90	&	&70.94	&\grc 6.13	\\
            &SVHN	&	&73.62	&\light 99.32	&	&68.18	&\grc 3.42	\\
             
            \bottomrule
        \end{tabular}
        }
    \label{tab:adap_simclr_ssl}
    \vspace*{-0.5em}

%% file: tab/ablation_model_cifar10_vgg.tex
    \caption{
        (\textbf{Supervised Learning}) Performance of MCR and \shortname\ on CIFAR-10 with 5\% poisoning rate using \textbf{VGG19-BN}.
        } 
    \vspace*{0.1em}
    \centering 
    \renewcommand{\arraystretch}{1.25}
    \resizebox{1\linewidth}{!}{
        \begin{tabular}{cccccccccc}
            \toprule
            CIFAR10& & \multicolumn{2}{c}{No Defense} & & \multicolumn{2}{c}{MCR} & & \multicolumn{2}{c}{\shortname} \\
            \cline{3-4} \cline{6-7} \cline{9-10}
            (Poisoning rate-5\%) &
            &ACC($\uparrow$) & ASR($\downarrow$) & & ACC($\uparrow$) & ASR($\downarrow$) & & ACC($\uparrow$) & ASR($\downarrow$) \\[2pt]
            \hline
            BadNet	&	&91.19	&\light 93.92	&	&90.44	&\light 55.21	&	&90.69	&\grc 1.76	\\
            Blended	&	&92.24	&\light 99.43	&	&91.29	&\light 96.56	&	&90.46	&\grc 9.09	\\
            Inputaware	&	&89.22	&\light 93.34	&	&90.96	&\grc 5.44	&	&90.58	&\grc 6.32	\\
            LC	&	&91.78	&\light 99.21	&	&91.05	&\light 100.00	&	&89.82	&\grc 3.23	\\
            LF	&	&89.27	&\light 96.29	&	&90.22	&\grc 1.02	&	&88.97	&\grc 1.38	\\
            SIG	&	&91.91	&\light 97.23	&	&91.09	&\light 99.87	&	&89.25	&\grc 8.22	\\
            SSBA	&	&91.53	&\light 90.39	&	&91.05	&\light 81.84	&	&91.03	&\grc 7.92	\\
            WaNet	&	&87.42	&\light 94.32	&	&92.04	&\grc 2.45	&	&90.36	&\grc 1.91	\\    
            \bottomrule

        \end{tabular}
        }
    \label{tab:ablation_model_cifar10_vgg}
    \vspace*{-0.5em}

%% file: tab/ablation_model_cifar10_iv3.tex
    \caption{
        (\textbf{Supervised Learning}) Performance of MCR and \shortname\ on CIFAR-10 with 5\% poisoning rate using \textbf{Inception-v3}.
        } 
    \vspace*{0.1em}
    \centering 
    \renewcommand{\arraystretch}{1.25}
    \resizebox{1\linewidth}{!}{
        \begin{tabular}{cccccccccc}
            \toprule
            CIFAR10& & \multicolumn{2}{c}{No Defense} & & \multicolumn{2}{c}{MCR} & & \multicolumn{2}{c}{\shortname} \\
            \cline{3-4} \cline{6-7} \cline{9-10}
            (Poisoning rate-5\%) &
            &ACC($\uparrow$) & ASR($\downarrow$) & & ACC($\uparrow$) & ASR($\downarrow$) & & ACC($\uparrow$) & ASR($\downarrow$) \\[2pt]
            \hline
            BadNet	&	&90.23	&\light 95.82	&	&89.74	&\light 70.16	&	&90.78	&\grc 1.88	\\
            Blended	&	&90.16	&\light 99.40	&	&90.69	&\light 91.57	&	&90.36	&\grc 6.93	\\
            Inputaware	&	&89.01	&\light 93.50	&	&90.94	&\grc 1.42	&	&90.71	&\grc 2.27	\\
            LC	&	&92.19	&\light 99.31	&	&90.06	&\light 96.57	&	&90.76	&\grc 3.09	\\
            LF	&	&91.43	&\light 93.12	&	&90.28	&\light 22.22	&	&91.00	&\grc 1.31	\\
            SIG	&	&90.89	&\light 91.29	&	&91.49	&\light 92.04	&	&89.44	&\grc 2.36	\\
            SSBA	&	&90.53	&\light 89.24	&	&90.86	&\light 62.65	&	&90.81	&\grc 3.85	\\
            WaNet	&	&89.34	&\light 91.26	&	&90.13	&\grc 5.44	&	&90.51	&\grc 1.87	\\              
            \bottomrule

        \end{tabular}
        }
    \label{tab:ablation_model_cifar10_inceptionv3}
    \vspace*{-0.5em}

%% file: tab/cifar10.tex
\begin{table*}[!th]
    \caption{
        Results on CIFAR10 under \textbf{supervised learning} scenarios. 
        Attack Success Rates (ASRs) below 15\% are highlighted in \grc{blue} to indicate a successful defense, 
        while ASRs above 15\% are denoted in \light{red} as failed defenses. 
    } 
    \vspace*{0.5em}
    \centering 
    \renewcommand{\arraystretch}{1.4}
    \resizebox{0.98\linewidth}{!}{
        \setlength\arrayrulewidth{1pt}
        \begin{tabular}{c|c|c *{9}{cc}}
        \hline
        \multirow{2}*{} &
        \multirow{2}*{Attacks} & \multirow{2}*{\shortstack{Poison\\Rate}} &
        \multicolumn{2}{c}{No Defense} & 
        \multicolumn{2}{c}{FP} &
        \multicolumn{2}{c}{NC} &
        \multicolumn{2}{c}{MCR} &
        \multicolumn{2}{c}{ANP} &
        \multicolumn{2}{c}{FT-SAM} &
        \multicolumn{2}{c}{I-BAU} &
        \multicolumn{2}{c}{SAU} &
        \multicolumn{2}{c}{TSC (ours)} \\[2pt]
        \cline{4-21}
        & & & ACC($\uparrow$) & ASR($\downarrow$) & ACC($\uparrow$) & ASR($\downarrow$) & 
        ACC($\uparrow$) & ASR($\downarrow$) & ACC($\uparrow$) & ASR($\downarrow$) & 
        ACC($\uparrow$) & ASR($\downarrow$) & ACC($\uparrow$) & ASR($\downarrow$) & 
        ACC($\uparrow$) & ASR($\downarrow$) & ACC($\uparrow$) & ASR($\downarrow$) & ACC($\uparrow$) & ASR($\downarrow$) \\[2pt]
        \hline
        \multirow{36}*{\rotatebox{90}{CIFAR10}}
        &\multirow{3}*{\shortstack{BadNet}}	&10\%	&91.63	&\light 93.88	&91.88	&\grc 0.82	&90.47	&\grc 1.08	&90.98	&\grc 2.01	&84.03	&\grc 0.00	&91.84	&\grc 1.63	&88.45	&\grc 2.40	&90.74	&\grc 1.08	&90.09	&\grc 1.16	\\
        &&5\%	&92.64	&\light 88.74	&92.26	&\grc 1.17	&90.53	&\grc 1.01	&92.17	&\grc 7.62	&86.45	&\grc 0.02	&92.19	&\grc 3.50	&88.66	&\grc 0.92	&89.32	&\grc 1.74	&89.19	&\grc 1.90	\\
        &&1\%	&93.14	&\light 74.73	&92.59	&\grc 2.29	&92.07	&\grc 0.77	&92.90	&\light 18.06	&85.82	&\grc 0.04	&92.39	&\grc 1.57	&87.80	&\grc 2.29	&65.38	&\grc 2.06	&90.71	&\grc 1.26	\\ \cline{2-21}
    &\multirow{3}*{\shortstack{Blended}}	&10\%	&93.46	&\light 99.78	&91.97	&\light 18.42	&90.75	&\grc 2.44	&92.93	&\light 97.87	&84.91	&\grc 6.14	&92.48	&\grc 11.38	&88.86	&\grc 9.52	&90.39	&\grc 9.57	&90.33	&\grc 8.79	\\
        &&5\%	&93.66	&\light 99.61	&92.70	&\light 49.47	&93.67	&\light 99.61	&93.23	&\light 99.01	&88.95	&\light 18.76	&93.00	&\light 29.59	&88.07	&\light 34.86	&90.69	&\grc 7.74	&90.14	&\grc 10.53	\\
        &&1\%	&93.76	&\light 94.88	&92.92	&\light 69.74	&93.76	&\light 94.88	&93.62	&\light 93.10	&89.69	&\light 60.52	&93.00	&\light 49.36	&89.62	&\light 25.74	&90.02	&\light 36.16	&91.12	&\grc 12.46	\\ \cline{2-21}
    &\multirow{3}*{\shortstack{Inputaware}}	&10\%	&91.54	&\light 88.34	&93.29	&\light 15.36	&92.77	&\grc 5.57	&93.25	&\light 60.84	&87.30	&\grc 0.20	&93.32	&\grc 2.66	&90.62	&\grc 0.78	&91.94	&\grc 1.40	&92.05	&\grc 3.11	\\
        &&5\%	&91.51	&\light 90.20	&93.25	&\light 35.21	&91.52	&\light 90.20	&92.94	&\light 95.49	&88.75	&\grc 0.22	&93.32	&\grc 2.88	&91.31	&\grc 8.43	&91.62	&\grc 1.67	&90.40	&\grc 5.07	\\
        &&1\%	&91.74	&\light 79.18	&93.16	&\grc 8.58	&91.74	&\light 79.19	&93.09	&\light 79.62	&83.95	&\grc 1.32	&93.83	&\grc 10.42	&90.98	&\grc 6.36	&91.60	&\grc 2.53	&92.04	&\grc 9.52	\\ \cline{2-21}
    &\multirow{3}*{\shortstack{LF}}	&10\%	&93.19	&\light 99.28	&92.37	&\light 42.14	&91.43	&\grc 2.50	&92.73	&\light 94.66	&87.60	&\grc 0.74	&92.68	&\grc 7.10	&86.45	&\light 28.03	&85.40	&\grc 2.01	&90.66	&\grc 3.90	\\
        &&5\%	&93.36	&\light 98.03	&92.84	&\light 59.12	&90.98	&\grc 2.43	&93.07	&\light 97.32	&84.20	&\grc 2.46	&92.89	&\grc 7.44	&88.64	&\light 45.66	&90.60	&\grc 1.71	&88.50	&\grc 3.78	\\
        &&1\%	&93.56	&\light 86.44	&92.45	&\light 65.80	&93.56	&\light 86.46	&93.09	&\light 84.11	&86.27	&\grc 11.28	&93.47	&\grc 11.71	&90.53	&\light 69.28	&91.58	&\light 18.12	&90.68	&\grc 11.67	\\ \cline{2-21}
    &\multirow{3}*{\shortstack{SSBA}}	&10\%	&92.88	&\light 97.07	&92.15	&\light 19.20	&92.88	&\light 97.07	&92.49	&\light 88.49	&84.86	&\grc 0.03	&92.18	&\grc 4.07	&88.61	&\grc 3.84	&88.75	&\grc 1.80	&90.54	&\grc 3.67	\\
        &&5\%	&93.27	&\light 94.91	&92.55	&\light 16.27	&93.27	&\light 94.91	&92.94	&\light 92.06	&88.72	&\grc 0.13	&92.71	&\grc 2.87	&89.65	&\grc 1.54	&91.30	&\grc 2.06	&89.43	&\grc 2.18	\\
        &&1\%	&93.43	&\light 73.44	&93.01	&\grc 7.68	&91.60	&\grc 0.46	&93.33	&\light 65.88	&85.33	&\grc 0.31	&93.02	&\grc 1.49	&89.56	&\grc 4.87	&91.38	&\grc 0.99	&91.18	&\grc 2.18	\\ \cline{2-21}
    &\multirow{3}*{\shortstack{WaNet}}	&10\%	&90.56	&\light 96.92	&93.18	&\grc 0.81	&90.56	&\light 96.92	&93.10	&\grc 0.71	&89.11	&\grc 0.42	&93.73	&\grc 0.74	&91.94	&\grc 13.44	&91.73	&\grc 0.80	&92.00	&\grc 0.98	\\
        &&5\%	&91.76	&\light 85.50	&93.66	&\grc 7.51	&91.76	&\light 85.50	&93.25	&\light 20.83	&87.64	&\grc 0.72	&93.85	&\grc 1.00	&90.66	&\grc 4.43	&91.70	&\grc 1.98	&90.46	&\grc 1.34	\\
        &&1\%	&90.65	&\grc 12.63	&93.47	&\grc 0.51	&92.55	&\grc 0.64	&93.48	&\grc 0.77	&83.25	&\grc 0.12	&93.96	&\grc 0.72	&91.87	&\grc 1.32	&91.90	&\grc 1.20	&91.40	&\grc 0.87	\\ \cline{2-21}
    &\multirow{3}*{\shortstack{SBL-BadNet}}	&10\%	&91.30	&\light 95.11	&92.00	&\light 91.07	&91.63	&\grc 0.34	&91.37	&\light 70.99	&90.72	&\grc 0.00	&91.41	&\light 89.13	&88.99	&\light 17.09	&90.63	&\grc 1.52	&90.36	&\grc 0.21	\\
        &&5\%	&90.79	&\light 93.48	&92.59	&\grc 1.13	&92.22	&\grc 0.59	&92.26	&\light 91.82	&82.82	&\light 51.63	&92.16	&\light 60.03	&90.67	&\light 27.06	&91.31	&\grc 0.60	&91.02	&\grc 1.12	\\
        &&1\%	&91.71	&\light 88.64	&93.10	&\light 31.77	&91.82	&\grc 0.72	&93.23	&\light 86.11	&82.71	&\light 81.48	&92.77	&\light 59.58	&90.63	&\grc 2.00	&92.32	&\grc 1.01	&91.54	&\grc 1.93	\\ \cline{2-21}
    &\multirow{3}*{\shortstack{SBL-Blended}}	&10\%	&90.46	&\light 94.12	&92.49	&\light 29.61	&90.46	&\light 88.12	&92.51	&\light 99.91	&86.32	&\light 52.96	&92.40	&\light 74.02	&91.44	&\light 22.79	&88.11	&\grc 9.09	&90.98	&\grc 8.27	\\
        &&5\%	&91.70	&\light 97.67	&92.97	&\light 79.74	&91.70	&\light 97.67	&92.75	&\light 99.61	&85.13	&\light 20.48	&92.50	&\light 77.90	&89.65	&\light 57.41	&91.43	&\grc 11.53	&90.47	&\grc 8.94	\\
        &&1\%	&92.07	&\light 91.84	&93.43	&\light 83.80	&92.07	&\light 91.84	&93.37	&\light 95.02	&85.30	&\light 58.19	&93.31	&\light 82.64	&90.67	&\light 64.08	&92.34	&\light 16.31	&90.11	&\grc 6.70	\\ \cline{2-21}
    &\multirow{3}*{\shortstack{SAPA}}	&5\%	&93.57	&\light 100.00	&92.56	&\light 41.88	&92.76	&\grc 2.51	&93.25	&\light 100.00	&84.83	&\grc 1.14	&92.80	&\grc 8.40	&88.51	&\grc 1.44	&91.39	&\grc 3.30	&91.13	&\grc 4.51	\\
        &&1\%	&94.01	&\light 99.97	&92.34	&\light 92.22	&92.80	&\grc 2.14	&93.83	&\light 100.00	&86.06	&\light 92.68	&93.06	&\light 79.80	&86.69	&\light 15.17	&91.83	&\grc 1.96	&90.37	&\grc 7.41	\\
        &&0.5\%	&93.77	&\light 84.80	&88.82	&\light 82.76	&92.74	&\grc 1.52	&93.78	&\light 80.83	&87.99	&\light 81.52	&93.23	&\light 82.02	&90.16	&\light 26.48	&91.75	&\grc 0.68	&90.98	&\grc 7.32	\\ \cline{2-21}
    &\multirow{3}*{\shortstack{LC}}	&10\%	&84.49	&\light 99.68	&89.90	&\grc 2.82	&90.17	&\grc 2.02	&89.95	&\grc 12.82	&84.08	&\light 90.01	&91.32	&\grc 3.09	&86.28	&\grc 3.80	&88.78	&\grc 0.36	&89.95	&\grc 2.74	\\
        &&5\%	&93.31	&\light 98.33	&92.19	&\light 72.99	&92.32	&\grc 0.64	&92.94	&\light 99.94	&88.15	&\grc 13.83	&92.59	&\light 57.18	&90.15	&\grc 1.99	&91.53	&\grc 1.50	&90.04	&\grc 2.38	\\
        &&1\%	&93.79	&\light 75.93	&92.86	&\light 29.86	&92.31	&\grc 0.68	&93.67	&\light 82.54	&86.58	&\light 31.46	&92.83	&\light 39.40	&89.78	&\grc 0.71	&92.16	&\grc 3.77	&90.08	&\grc 5.78	\\ \cline{2-21}
    &\multirow{3}*{\shortstack{SIG}}	&10\%	&84.48	&\light 97.43	&89.95	&\grc 0.69	&84.48	&\light 97.43	&90.06	&\grc 1.33	&80.94	&\grc 0.01	&91.61	&\grc 0.24	&86.39	&\grc 3.61	&89.40	&\grc 2.87	&89.67	&\grc 0.64	\\
        &&5\%	&93.29	&\light 95.06	&92.81	&\light 43.02	&93.28	&\light 95.06	&92.96	&\light 95.82	&86.51	&\grc 4.76	&92.62	&\grc 0.94	&87.98	&\grc 7.54	&90.63	&\grc 0.40	&90.39	&\grc 2.19	\\
        &&1\%	&93.82	&\light 83.40	&93.00	&\light 70.51	&93.82	&\light 83.40	&93.56	&\light 86.31	&91.49	&\light 52.69	&93.20	&\light 37.74	&88.18	&\light 38.01	&91.07	&\light 33.71	&91.67	&\grc 10.97	\\ \cline{2-21}
    &\multirow{3}*{\shortstack{Narcissus}}	&5\%	&93.72	&\light 90.91	&91.93	&\light 68.61	&93.72	&\light 80.91	&93.63	&\light 86.64	&87.87	&\light 49.27	&93.19	&\light 27.92	&87.82	&\light 73.79	&90.72	&\grc 1.57	&91.35	&\grc 14.48	\\
	&&1\%	&93.68	&\light 82.87	&92.29	&\light 44.88	&93.68	&\light 47.87	&93.61	&\light 49.79	&92.01	&\light 27.01	&93.05	&\light 26.80	&90.21	&\light 18.67	&91.36	&\grc 3.24	&90.65	&\grc 7.88	\\
	&&0.5\%	&93.68	&\light 80.58	&92.94	&\light 29.59	&93.67	&\light 32.57	&93.69	&\light 32.96	&89.35	&\light 16.78	&93.06	&\grc 14.08	&89.16	&\light 21.09	&91.74	&\grc 5.81	&91.71	&\grc 8.02	\\ 

    \hline
        \end{tabular}
        }
    \label{tab:cifar10_defense}
    \vspace*{-0.5em}
\end{table*}

%% file: tab/imagenet.tex
\begin{table*}[th]
    \caption{
        Results on ImageNet100 under \textbf{supervised learning} scenarios. 
    } 
    \vspace*{0.5em}
    \centering 
    \renewcommand{\arraystretch}{1.4}
    \resizebox{0.98\linewidth}{!}{
        \setlength\arrayrulewidth{1pt}
        \begin{tabular}{c|c|c *{9}{cc}}
        \hline
        \multirow{2}*{} &
        \multirow{2}*{Attacks} & \multirow{2}*{\shortstack{Poison\\Rate}} &
        \multicolumn{2}{c}{No Defense} & 
        \multicolumn{2}{c}{FP} &
        \multicolumn{2}{c}{NC} &
        \multicolumn{2}{c}{MCR} &
        \multicolumn{2}{c}{ANP} &
        \multicolumn{2}{c}{FT-SAM} &
        \multicolumn{2}{c}{I-BAU} &
        \multicolumn{2}{c}{SAU} &
        \multicolumn{2}{c}{TSC (ours)} \\[2pt]
        \cline{4-21}
        & & & ACC($\uparrow$) & ASR($\downarrow$) & ACC($\uparrow$) & ASR($\downarrow$) & 
        ACC($\uparrow$) & ASR($\downarrow$) & ACC($\uparrow$) & ASR($\downarrow$) & 
        ACC($\uparrow$) & ASR($\downarrow$) & ACC($\uparrow$) & ASR($\downarrow$) & 
        ACC($\uparrow$) & ASR($\downarrow$) & ACC($\uparrow$) & ASR($\downarrow$) & ACC($\uparrow$) & ASR($\downarrow$) \\[2pt]
        \hline
    \multirow{22}*{\rotatebox{90}{ImageNet100}}
    &\multirow{2}*{\shortstack{BadNet}}	&0.5\%	&84.30	&\light 99.78	&83.36	&\grc 9.80	&81.92	&\grc 0.52	&85.24	&\light 99.66	&78.44	&\light 94.18	&83.70	&\grc 9.45	&73.70	&\grc 8.34	&73.86	&\grc 0.28	&80.20	&\grc 0.22	\\
	&&1\%	&84.56	&\light 99.86	&83.10	&\grc 9.58	&81.92	&\grc 0.49	&85.08	&\light 99.86	&79.48	&\light 93.64	&83.88	&\light 24.14	&71.46	&\light 43.66	&72.84	&\grc 0.26	&78.06	&\grc 0.14	\\ \cline{2-21}
&\multirow{2}*{\shortstack{Blended}}	&0.5\%	&84.44	&\light 94.32	&82.80	&\light 63.25	&84.44	&\light 94.32	&85.58	&\light 94.97	&84.56	&\light 93.27	&83.40	&\light 75.43	&74.22	&\light 62.34	&73.84	&\grc 3.72	&76.58	&\grc 12.63	\\
	&&1\%	&84.90	&\light 98.04	&83.36	&\light 69.21	&80.21	&\light 70.21	&85.04	&\light 97.58	&84.54	&\light 97.70	&83.86	&\light 82.00	&73.10	&\light 61.25	&69.24	&\grc 0.53	&75.88	&\grc 6.35	\\ \cline{2-21}
&\multirow{2}*{\shortstack{Inputaware}}	&0.5\%	&76.62	&\light 69.94	&83.56	&\light 16.57	&76.62	&\light 69.94	&83.56	&\light 51.52	&73.82	&\light 42.20	&82.96	&\light 39.23	&71.80	&\light 43.11	&76.18	&\grc 1.27	&80.10	&\grc 4.63	\\
	&&1\%	&77.66	&\light 65.13	&83.54	&\light 25.23	&72.66	&\light 43.62	&83.54	&\light 68.55	&70.56	&\light 56.24	&82.60	&\light 54.46	&73.64	&\light 56.53	&70.86	&\light 24.63	&76.46	&\grc 1.09	\\ \cline{2-21}
&\multirow{2}*{\shortstack{LF}}	&0.5\%	&84.24	&\light 98.87	&83.10	&\light 50.26	&84.24	&\light 98.87	&85.70	&\light 97.70	&81.32	&\light 86.20	&83.80	&\light 70.48	&74.36	&\light 74.97	&75.22	&\grc 0.18	&78.78	&\grc 5.39	\\
	&&1\%	&83.92	&\light 99.56	&83.00	&\light 35.82	&76.76	&\light 49.87	&85.30	&\light 99.03	&81.10	&\light 88.53	&83.40	&\light 70.69	&71.06	&\light 22.32	&67.38	&\grc 2.93	&78.58	&\grc 5.41	\\ \cline{2-21}
&\multirow{2}*{\shortstack{SSBA}}	&0.5\%	&84.30	&\light 95.31	&83.34	&\light 46.75	&84.30	&\light 95.31	&85.04	&\light 95.13	&76.96	&\grc 6.18	&83.16	&\light 15.70	&71.52	&\grc 1.19	&76.12	&\grc 0.89	&79.56	&\grc 1.45	\\
	&&1\%	&84.02	&\light 99.43	&83.34	&\light 59.68	&78.47	&\light 70.78	&85.14	&\light 97.72	&80.22	&\light 22.77	&83.50	&\light 20.30	&72.58	&\grc 7.45	&73.94	&\grc 0.36	&79.88	&\grc 4.91	\\ \cline{2-21}
&\multirow{2}*{\shortstack{WaNet}}	&0.5\%	&83.34	&\light 53.92	&84.04	&\grc 0.95	&77.34	&\light 22.66	&85.30	&\grc 14.34	&79.66	&\grc 0.00	&83.90	&\grc 0.12	&72.46	&\grc 0.00	&76.18	&\grc 0.18	&80.70	&\grc 0.28	\\
	&&1\%	&79.62	&\light 90.69	&84.90	&\grc 4.87	&78.34	&\grc 14.12	&85.44	&\light 25.52	&78.56	&\grc 0.04	&83.28	&\grc 0.26	&71.24	&\grc 1.56	&76.42	&\grc 1.23	&80.20	&\grc 0.32	\\ \cline{2-21}
&\multirow{2}*{\shortstack{SBL-BadNet}}	&0.5\%	&72.50	&\light 100.00	&85.10	&\light 67.07	&70.15	&\grc 3.62	&82.76	&\light 88.61	&65.40	&\light 89.49	&83.98	&\light 40.71	&72.34	&\grc 3.52	&78.54	&\grc 2.57	&80.32	&\grc 0.34	\\
	&&1\%	&72.64	&\light 100.00	&85.36	&\light 79.82	&69.97	&\grc 1.95	&82.24	&\light 90.44	&67.36	&\light 89.49	&83.68	&\light 64.97	&74.02	&\light 19.33	&78.14	&\grc 0.36	&77.48	&\grc 0.22	\\ \cline{2-21}
&\multirow{2}*{\shortstack{SBL-Blended}}	&0.5\%	&72.52	&\light 97.56	&85.10	&\light 89.29	&72.52	&\light 97.56	&82.54	&\light 92.63	&68.28	&\light 97.05	&83.84	&\light 89.07	&70.78	&\light 37.35	&73.66	&\grc 14.87	&79.42	&\grc 7.18	\\
	&&1\%	&72.68	&\light 99.17	&83.41	&\light 68.24	&71.42	&\light 70.14	&82.82	&\light 95.78	&72.72	&\light 99.15	&83.92	&\light 92.69	&73.52	&\light 20.30	&76.78	&\light 39.29	&77.16	&\grc 8.87	\\ \cline{2-21}
&\multirow{2}*{\shortstack{SAPA}}	&0.5\%	&85.04	&\light 98.83	&83.44	&\light 20.53	&78.57	&\grc 9.20	&85.60	&\light 93.05	&80.42	&\light 96.59	&83.82	&\light 30.04	&69.32	&\light 18.34	&73.34	&\grc 1.07	&79.00	&\grc 1.74	\\
	&&1\%	&85.50	&\light 98.79	&83.34	&\light 27.86	&77.42	&\grc 3.52	&85.42	&\light 95.80	&83.12	&\light 93.76	&83.96	&\light 45.88	&69.12	&\light 41.64	&75.42	&\grc 1.23	&78.14	&\grc 1.41	\\ \cline{2-21}
&\multirow{2}*{\shortstack{LC}}	&0.5\%	&84.22	&\grc 0.61	&83.34	&\grc 0.20	&84.22	&\grc 0.61	&85.04	&\grc 0.93	&84.48	&\grc 0.57	&83.86	&\grc 0.24	&69.72	&\grc 0.36	&76.20	&\grc 0.16	&80.36	&\grc 0.22	\\
	&&1\%	&84.10	&\light 32.97	&83.48	&\grc 4.28	&81.13	&\grc 0.42	&85.48	&\light 76.75	&84.22	&\light 32.42	&83.58	&\grc 8.06	&73.52	&\grc 1.76	&70.50	&\grc 0.79	&80.18	&\grc 0.57	\\ \cline{2-21}
&\multirow{2}*{\shortstack{SIG}}	&0.5\%	&84.20	&\light 16.22	&83.26	&\grc 2.22	&78.48	&\grc 0.43	&85.18	&\light 18.34	&84.00	&\light 15.86	&83.88	&\grc 4.51	&70.76	&\grc 3.43	&75.22	&\grc 0.12	&77.20	&\grc 0.55	\\
	&&1\%	&84.16	&\light 70.08	&83.40	&\light 20.48	&79.58	&\grc 0.89	&85.02	&\light 77.84	&80.40	&\light 65.68	&83.36	&\light 44.81	&70.98	&\light 19.98	&73.76	&\grc 0.30	&80.22	&\grc 9.98	\\
    \hline
        \end{tabular}
        }
    \label{tab:imagnet_defense}
\end{table*}

%% file: tab/gtsrb.tex
\begin{table*}[!h]
    \caption{
        Results on GTSRB under \textbf{supervised learning} scenarios. 
  } 
    \vspace*{0.5em}
    \centering 
    \renewcommand{\arraystretch}{1.4}
    \resizebox{0.98\linewidth}{!}{
        \setlength\arrayrulewidth{1pt}
        \begin{tabular}{c|c|c *{9}{cc}}
        \hline
        \multirow{2}*{} &
        \multirow{2}*{Attacks} & \multirow{2}*{\shortstack{Poison\\Rate}} &
        \multicolumn{2}{c}{No Defense} & 
        \multicolumn{2}{c}{FP} &
        \multicolumn{2}{c}{NC} &
        \multicolumn{2}{c}{MCR} &
        \multicolumn{2}{c}{ANP} &
        \multicolumn{2}{c}{FT-SAM} &
        \multicolumn{2}{c}{I-BAU} &
        \multicolumn{2}{c}{SAU} &
        \multicolumn{2}{c}{TSC (ours)} \\[2pt]
        \cline{4-21}
        & & & ACC($\uparrow$) & ASR($\downarrow$) & ACC($\uparrow$) & ASR($\downarrow$) & 
        ACC($\uparrow$) & ASR($\downarrow$) & ACC($\uparrow$) & ASR($\downarrow$) & 
        ACC($\uparrow$) & ASR($\downarrow$) & ACC($\uparrow$) & ASR($\downarrow$) & 
        ACC($\uparrow$) & ASR($\downarrow$) & ACC($\uparrow$) & ASR($\downarrow$) & ACC($\uparrow$) & ASR($\downarrow$) \\[2pt]
        \hline
    \multirow{26}*{\rotatebox{90}{GTSRB}}
    &\multirow{3}*{\shortstack{BadNet}}	&10\%	&97.62	&\light 95.48	&98.21	&\grc 0.09	&97.48	&\grc 0.01	&98.68	&\grc 4.18	&95.86	&\grc 0.00	&98.82	&\grc 0.31	&96.47	&\grc 0.02	&97.75	&\grc 0.02	&98.05	&\grc 0.01	\\
	&&5\%	&97.89	&\light 93.00	&97.86	&\grc 2.11	&97.09	&\grc 0.00	&97.99	&\grc 4.02	&92.72	&\grc 0.00	&98.96	&\grc 0.09	&14.43	&\grc 0.00	&94.65	&\grc 0.00	&98.15	&\grc 0.00	\\
	&&1\%	&98.39	&\light 79.23	&98.48	&\grc 0.01	&97.20	&\grc 0.00	&98.32	&\grc 3.33	&93.99	&\grc 0.00	&98.73	&\grc 0.00	&93.80	&\grc 0.02	&10.30	&\grc 0.00	&98.20	&\grc 0.01	\\ \cline{2-21}
&\multirow{3}*{\shortstack{Blended}}	&10\%	&98.62	&\light 100.00	&98.38	&\light 100.00	&97.76	&\grc 8.03	&98.62	&\light 100.00	&95.85	&\light 42.80	&98.38	&\light 49.82	&92.35	&\light 86.35	&96.15	&\grc 6.51	&97.67	&\grc 11.95	\\
	&&5\%	&98.86	&\light 99.96	&98.75	&\light 99.93	&96.64	&\grc 3.14	&98.79	&\light 99.87	&93.92	&\light 68.09	&98.48	&\light 25.18	&84.63	&\grc 0.00	&92.80	&\grc 4.53	&97.18	&\grc 12.59	\\
	&&1\%	&98.80	&\light 96.95	&98.71	&\light 95.87	&97.00	&\grc 10.88	&98.75	&\light 96.00	&92.78	&\light 79.88	&98.57	&\light 38.01	&93.98	&\light 27.85	&95.32	&\grc 2.55	&97.11	&\grc 12.98	\\ \cline{2-21}
&\multirow{3}*{\shortstack{Inputaware}}	&10\%	&98.76	&\light 95.93	&98.91	&\grc 4.46	&98.76	&\light 95.92	&99.00	&\grc 13.88	&98.19	&\grc 0.00	&99.45	&\grc 0.07	&97.09	&\grc 0.52	&98.37	&\grc 0.08	&99.11	&\grc 0.02	\\
	&&5\%	&98.26	&\light 92.84	&98.93	&\grc 4.07	&98.26	&\light 92.84	&99.21	&\light 17.58	&98.82	&\grc 0.00	&99.53	&\grc 0.01	&98.61	&\grc 0.07	&98.64	&\grc 0.00	&99.37	&\grc 0.01	\\
	&&1\%	&98.75	&\grc 7.05	&99.45	&\grc 0.03	&99.12	&\grc 0.61	&99.15	&\grc 11.13	&98.44	&\grc 0.03	&99.62	&\grc 0.14	&97.87	&\grc 0.00	&98.43	&\grc 0.37	&99.40	&\grc 0.02	\\ \cline{2-21}
&\multirow{3}*{\shortstack{LF}}	&10\%	&97.89	&\light 99.36	&98.28	&\light 82.28	&97.23	&\grc 0.27	&97.93	&\light 98.81	&96.07	&\grc 0.00	&98.03	&\grc 3.93	&95.78	&\light 16.15	&96.45	&\grc 0.76	&98.33	&\grc 3.91	\\
	&&5\%	&98.16	&\light 98.81	&97.89	&\light 98.17	&97.74	&\grc 0.02	&97.95	&\light 98.90	&89.10	&\grc 2.16	&98.47	&\grc 0.99	&87.54	&\grc 1.40	&95.88	&\grc 0.01	&98.31	&\grc 2.05	\\
	&&1\%	&98.17	&\light 96.11	&97.63	&\light 71.56	&97.55	&\grc 0.49	&98.11	&\light 95.86	&88.60	&\light 15.49	&98.34	&\grc 1.28	&96.70	&\light 68.50	&94.74	&\grc 0.01	&97.52	&\grc 2.62	\\ \cline{2-21}
&\multirow{3}*{\shortstack{SSBA}}	&10\%	&97.90	&\light 99.47	&97.75	&\light 99.46	&97.72	&\grc 0.29	&97.95	&\light 99.33	&88.47	&\grc 0.00	&98.32	&\light 34.71	&96.14	&\grc 1.88	&96.55	&\grc 0.32	&97.63	&\grc 6.23	\\
	&&5\%	&98.01	&\light 99.22	&98.12	&\light 99.12	&97.03	&\grc 0.33	&97.97	&\light 98.98	&90.29	&\grc 6.09	&97.66	&\grc 4.93	&6.12	&\grc 0.00	&6.32	&\grc 0.00	&98.03	&\grc 3.12	\\
	&&1\%	&98.84	&\light 94.51	&98.78	&\light 91.38	&97.51	&\grc 0.12	&98.78	&\light 91.65	&89.87	&\grc 0.02	&98.40	&\grc 6.55	&86.37	&\grc 0.96	&93.94	&\grc 1.96	&97.48	&\grc 0.64	\\ \cline{2-21}
&\multirow{3}*{\shortstack{WaNet}}	&10\%	&97.74	&\light 94.25	&97.62	&\light 88.07	&98.25	&\grc 0.00	&98.52	&\grc 1.94	&97.08	&\grc 0.00	&0.00	&\grc 0.00	&96.72	&\grc 0.00	&98.91	&\grc 0.04	&98.71	&\grc 0.00	\\
	&&5\%	&97.42	&\light 92.85	&98.00	&\grc 12.24	&97.42	&\light 92.85	&98.67	&\grc 5.80	&97.09	&\grc 0.00	&98.97	&\grc 0.00	&97.93	&\grc 0.25	&98.19	&\grc 0.05	&98.86	&\grc 0.00	\\
	&&1\%	&97.08	&\light 62.24	&98.08	&\grc 1.14	&98.88	&\grc 9.67	&98.51	&\light 53.94	&97.66	&\grc 0.00	&99.04	&\grc 0.35	&96.27	&\grc 0.00	&97.73	&\grc 0.06	&98.50	&\grc 0.00	\\ \cline{2-21}
&\multirow{3}*{\shortstack{SBL-BadNet}}	&10\%	&88.84	&\light 95.47	&97.50	&\grc 0.13	&96.44	&\grc 0.04	&96.95	&\grc 0.00	&98.52	&\grc 0.00	&92.06	&\grc 0.00	&97.13	&\grc 0.00	&97.13	&\grc 0.00	&97.56	&\grc 0.03 \\
	&&5\%	&89.21	&\light 95.10	&97.78	&\grc 0.18	&97.27	&\grc 0.00	&97.67	&\grc 1.36	&98.42	&\light 51.52	&92.84	&\grc 0.00	&93.73	&\grc 0.00	&93.73	&\grc 0.00	&98.09	&\grc 0.00 \\
	&&1\%	&88.93	&\light 89.72	&97.76	&\grc 10.56	&97.03	&\grc 0.00	&97.47	&\light 21.06	&98.43	&\grc 8.64	&92.29	&\grc 0.02	&96.86	&\grc 0.03	&96.86	&\grc 0.03	&98.05	&\grc 0.00 \\ \cline{2-21}
&\multirow{3}*{\shortstack{SAPA}}	&5\%	&98.31	&\light 100.00	&98.00	&\light 97.31	&96.18	&\grc 0.83	&98.24	&\light 100.00	&98.03	&\grc 0.02	&96.12	&\grc 0.00	&96.42	&\grc 0.06	&96.42	&\grc 0.06	&98.18	&\grc 0.10 \\
	&&1\%	&98.28	&\light 100.00	&98.16	&\light 89.42	&98.00	&\grc 0.25	&98.27	&\light 99.96	&98.06	&\grc 0.02	&90.67	&\grc 0.00	&96.22	&\grc 0.00	&96.22	&\grc 0.00	&97.56	&\grc 0.64 \\
	&&0.5\%	&98.44	&\light 100.00	&98.43	&\light 55.10	&97.95	&\grc 0.09	&98.46	&\grc 0.00	&98.63	&\grc 0.00	&93.15	&\grc 0.00	&68.65	&\grc 0.27	&68.65	&\grc 0.27	&97.64	&\grc 0.02 \\ \cline{2-21}
    &\multirow{2}*{\shortstack{LC}}	&0.5\%	&98.02	&\grc 0.00	&97.76	&\grc 0.01	&98.02	&\grc 0.00	&98.03	&\grc 0.01	&98.02	&\grc 0.00	&97.93	&\grc 0.00	&94.66	&\grc 0.00	&68.80	&\grc 0.43	&98.01	&\grc 0.00	\\
	&&0.1\%	&98.46	&\grc 0.01	&98.37	&\grc 0.00	&98.46	&\grc 0.01	&98.53	&\grc 0.01	&98.46	&\grc 0.01	&98.27	&\grc 0.00	&93.29	&\grc 0.03	&95.34	&\grc 0.02	&98.10	&\grc 0.00	\\ \cline{2-21}
&\multirow{2}*{\shortstack{SIG}}	&0.5\%	&98.52	&\light 71.33	&98.38	&\light 71.26	&98.52	&\light 71.30	&98.57	&\light 68.54	&94.24	&\light 58.47	&97.98	&\grc 0.63	&68.45	&\grc 0.00	&94.39	&\grc 0.79	&98.03	&\grc 3.33	\\
	&&0.1\%	&98.69	&\light 58.09	&98.84	&\light 56.11	&98.69	&\light 58.09	&98.70	&\light 57.91	&91.21	&\light 33.44	&98.38	&\grc 0.91	&94.07	&\grc 0.57	&89.16	&\grc 0.36	&98.22	&\grc 0.08	\\

    \hline

        \end{tabular}
        }
    \label{tab:gtsrb_defense}
\end{table*}

%% file: tab/simCLR_full.tex
    \caption{
        Defense results under \textbf{self-supervised learning (SimCLR)} settings. 
        We evaluate MCR \cite{MCR}, SSL-Cleanse \cite{SSL_Cleanse}, 
        and TSC against the \textbf{BadEncoder} attack \cite{BadEncoder}. 
    } 
    \vspace*{0.4em}
    \centering 
    \renewcommand{\arraystretch}{1.2}
    \resizebox{1\linewidth}{!}{
        \setlength\arrayrulewidth{1.1pt}
        \begin{tabular}{cc *{4}{cc}}
        \hline
        \multirow{2}*{\shortstack{Pre-training\\Dataset}} &
        \multirow{2}*{\shortstack{Downstream\\Dataset}} &
        \multicolumn{2}{c}{No Defense} & 
        \multicolumn{2}{c}{MCR} &
        \multicolumn{2}{c}{SSL-Cleanse} &
        \multicolumn{2}{c}{TSC (ours)} \\[2pt]
        \cmidrule(r){3-4} \cmidrule(lr){5-6} \cmidrule(l){7-8} \cmidrule(l){9-10}
        & & ACC($\uparrow$) & ASR($\downarrow$) & ACC($\uparrow$) & ASR($\downarrow$) & ACC($\uparrow$) & ASR($\downarrow$) &
        ACC($\uparrow$) & ASR($\downarrow$) \\[2pt]
        \hline
        \multirow{3}*{\shortstack{CIFAR10}}	&STL10	&76.74	&\light 99.65	&74.93	&\grc 7.92	&70.51	&\grc 2.44	&71.11	&\grc 4.44	\\
	&GTSRB	&81.12	&\light 98.79	&75.51	&\grc 0.54	&75.50	&\grc 1.23	&77.57	&\grc 1.68	\\
	&SVHN	&63.12	&\light 98.71	&57.35	&\light 65.58	&61.01	&\grc 7.95	&64.13	&\grc 10.26	\\ \hline
\multirow{3}*{\shortstack{ImageNet}}	&STL10	&94.93	&\light 98.99	&90.20	&\grc 2.08	&88.72	&\grc 1.69	&86.99	&\grc 3.11	\\
	&GTSRB	&75.94	&\light 99.76	&72.38	&\grc 0.13	&67.55	&\grc 1.81	&69.47	&\grc 6.47	\\
	&SVHN	&72.64	&\light 99.21	&71.27	&\light 34.15	&67.96	&\grc 8.00	&66.44	&\grc 3.64	\\ \hline
  
        \end{tabular}
        }
    \label{tab:simclr_defense_full}
    \vspace*{-0.5em}

%% file: tab/simCLR_ctrl.tex
    \caption{
        Defense results under \textbf{self-supervised learning (SimCLR)} settings. 
        We evaluate MCR \cite{MCR}, SSL-Cleanse \cite{SSL_Cleanse}, 
        and TSC against the \textbf{CTRL} attack \cite{embarrassing}. 
        Following CTRL's evaluation methodology, 
        we also assess ACC and ASR on the pre-training dataset using K-Nearest Neighbor (KNN) classification.
        } 
    \vspace*{0.4em}
    \centering 
    \renewcommand{\arraystretch}{1.2}
    \resizebox{1\linewidth}{!}{
        \setlength\arrayrulewidth{1.1pt}
        \begin{tabular}{cc *{4}{cc}}
        \hline
        \multirow{2}*{\shortstack{Pre-training\\Dataset}} &
        \multirow{2}*{\shortstack{Evaluation\\Dataset}} &
        \multicolumn{2}{c}{No Defense} & 
        \multicolumn{2}{c}{MCR} &
        \multicolumn{2}{c}{SSL-Cleanse} &
        \multicolumn{2}{c}{TSC (ours)} \\[2pt]
        \cmidrule(r){3-4} \cmidrule(lr){5-6} \cmidrule(l){7-8} \cmidrule(l){9-10}
        & & ACC($\uparrow$) & ASR($\downarrow$) & ACC($\uparrow$) & ASR($\downarrow$) & ACC($\uparrow$) & ASR($\downarrow$) &
        ACC($\uparrow$) & ASR($\downarrow$) \\[2pt]
        \hline
        \multirow{4}*{\shortstack{CIFAR10 \\ (Poisoning rate-5\%)}}	&CIFAR10 (KNN)	&84.66	&\light 93.84	&81.45	&\light 44.64	&80.05	&\grc 2.95	&82.25	&\grc 3.95	\\
	&STL10 (linear probe)	&74.69	&\light 25.60	&70.48	&\grc 13.72	&70.55	&\grc 1.28	&71.29	&\grc 2.45	\\
	&GTSRB (linear probe)	&72.65	&\light 41.80	&71.05	&\light 24.54	&69.55	&\grc 4.16	&70.95	&\grc 4.25	\\
	&SVHN (linear probe)	&60.42	&\light 62.64	&57.35	&\light 32.58	&59.22	&\grc 0.24	&58.91	&\grc 7.42	\\ \hline
\multirow{4}*{\shortstack{ImageNet100\\ (Poisoning rate-5\%)}}	&ImageNet100 (KNN)	&43.66	&\light 42.53	&42.39	&\light 34.24	&41.20	&\grc 1.86	&41.80	&\grc 0.24	\\
	&STL10 (linear probe)	&74.96	&\light 17.27	&70.24	&\grc 12.33	&69.47	&\grc 1.43	&72.05	&\grc 3.37	\\
	&GTSRB (linear probe)	&63.58	&\light 68.20	&60.85	&\light 50.41	&61.85	&\grc 5.41	&60.73	&\grc 1.62	\\
	&SVHN (linear probe)	&56.73	&\light 90.77	&56.27	&\light 52.77	&53.70	&\grc 10.04	&54.39	&\grc 1.21	\\ \hline

        \end{tabular}
        }
    \label{tab:simclr_ctrl_defense}
    \vspace*{-0.5em}

%% file: tab/CLIP_full.tex
    \caption{
        Defense results under \textbf{self-supervised learning (CLIP)} settings. 
        We evaluate MCR \cite{MCR} and  
        and \shortname\ against the \textbf{BadEncoder} attack \cite{BadEncoder}.
        } 
    \vspace*{0.4em}
    \centering 
    \renewcommand{\arraystretch}{1.05}
    \resizebox{1\linewidth}{!}{
        \setlength\arrayrulewidth{1.2pt}
        \begin{tabular}{cc *{3}{cc}}
        \hline
        \multirow{2}*{\shortstack{Pre-training\\Dataset}} &
        \multirow{2}*{\shortstack{Downstream\\Dataset}} &
        \multicolumn{2}{c}{No Defense} & 
        \multicolumn{2}{c}{MCR} &
        \multicolumn{2}{c}{TSC (ours)} \\[2pt]
        \cmidrule(r){3-4} \cmidrule(lr){5-6} \cmidrule(l){7-8}
        & & ACC($\uparrow$) & ASR($\downarrow$) & ACC($\uparrow$) & ASR($\downarrow$) & 
        ACC($\uparrow$) & ASR($\downarrow$) \\[2pt]
        \hline
        \multirow{5}*{\shortstack{CLIP\\(linear probe)}}	&STL10	&97.07	&\light 99.33	&96.43	&\light 99.86	&94.15	&\grc 0.67	\\
        &GTSRB	&82.36	&\light 99.40	&78.22	&\light 99.21	&72.42	&\grc 5.32	\\
        &CIFAR10	&86.36	&\light 99.98	&84.36	&\light 99.45	&79.31	&\grc 1.32	\\
        &Food101	&72.58	&\light 97.91	&72.36	&\light 96.62	&69.33	&\grc 1.04	\\
        &Pascal VOC 2007	&76.07	&\light 99.83	&75.47	&\light 99.92	&78.42	&\grc 0.34	\\ \hline
    \multirow{5}*{\shortstack{CLIP\\(zero-shot)}}	&STL10	&94.06	&\light 99.86	&91.51	&\light 99.85	&90.25	&\grc 0.88	\\
        &GTSRB	&29.94	&\light 99.44	&24.66	&\light 99.34	&17.24	&\grc 2.01	\\
        &CIFAR10	&69.95	&\light 99.39	&62.51	&\light 99.10	&41.59	&\grc 1.28	\\
        &Food101	&67.72	&\light 99.96	&66.51	&\light 99.56	&61.69	&\grc 0.28	\\
        &Pascal VOC 2007	&71.22	&\light 99.92	&70.09	&\light 99.12	&75.08	&\grc 1.45	\\ \hline
    
        \end{tabular}
        }
    \label{tab:clip_defense_full}
    \vspace*{-0.5em}

%% file: tab/simCLR_sau.tex
    \caption{
        Defense results of I-BAU \cite{I_BAU} and SAU \cite{sau} under SimCLR training scenario, 
        where linear probing is used to evaluate the downstream tasks.
    } 
    \vspace*{0.4em}
    \centering 
    \renewcommand{\arraystretch}{1.05}
    \resizebox{1\linewidth}{!}{
        \setlength\arrayrulewidth{1.2pt}
        \begin{tabular}{cc *{3}{cc}}
        \hline
        \multirow{2}*{\shortstack{Pre-training\\Dataset}} &
        \multirow{2}*{\shortstack{Downstream\\Dataset}} &
        \multicolumn{2}{c}{No Defense} & 
        \multicolumn{2}{c}{I-BAU} &
        \multicolumn{2}{c}{SAU} \\[2pt]
        \cmidrule(r){3-4} \cmidrule(lr){5-6} \cmidrule(l){7-8}
        & & ACC($\uparrow$) & ASR($\downarrow$) & ACC($\uparrow$) & ASR($\downarrow$) & 
        ACC($\uparrow$) & ASR($\downarrow$) \\[2pt]
        \hline
        \multirow{3}*{\shortstack{CIFAR10}}	&STL10	&76.74	&\light 99.65	&30.13	&\grc 12.42	&21.52	&\grc 7.22		\\
            &GTSRB	&81.12	&\light 98.79	&22.36	&\light 15.10	&47.45	&\grc 5.52		\\
            &SVHN	&63.12	&\light 98.71	&43.44	&\light 17.11	&24.07	&\grc 7.25		\\ \hline
        \multirow{3}*{\shortstack{ImageNet100}}	&STL10	&94.93	&\light 98.99	&74.62	&\grc 10.32	&81.24	&\grc 11.83		\\
            &GTSRB	&75.94	&\light 99.76	&39.85	&\grc 7.34	&10.75	&\grc 0.80		\\
            &SVHN	&72.64	&\light 99.21	&25.27	&\grc 13.70	&24.37	&\grc 4.10		\\ \hline
        \end{tabular}
        }
    \label{tab:simclr_sau}

%% file: tab/sl_clean.tex
\caption{
        (\textbf{Supervised Learning}) Clean Accuracy (ACC) drop results of various defenses on CIFAR10, GTSRB and ImageNet100 with poisoning rate of 0 (no poison).
        } 
    \vspace*{0.1em}
    \centering 
    \renewcommand{\arraystretch}{1.4}
    \resizebox{1\linewidth}{!}{
    \begin{tabular}{c|ccccccccc}
    \toprule
    Dataset     & No Defense & FP    & NC             & MCR   & ANP   & FT-SAM & I-BAU & SAU   & TSC (ours)   \\
    \hline
    CIFAR10     & 93.12      & 92.10 & \textbf{93.12} & 90.98 & 83.09 & 91.32  & 87.47 & 89.97 & 91.12 \\
    GTSRB       & 99.20      & 99.11 & \textbf{99.20} & 98.14 & 95.43 & 98.04  & 95.40 & 96.84 & 98.31 \\
    ImageNet100 & 84.32      & 82.22 & \textbf{84.32} & 81.46 & 77.78 & 82.90  & 77.10 & 76.93 & 81.44\\
    \bottomrule    
\end{tabular}
        }
    \label{tab:sl_clean}
    \vspace*{-0.5em}


%% file: tab/ssl_clean.tex
    \caption{
        (\textbf{Self-supervised Learning}) Clean Accuracy (ACC) drop results for SimCLR of various defenses on CIFAR10 and ImageNet100 with poisoning rate of 0 (no poison).
        } 
    \vspace*{0.1em}
    \centering 
    \renewcommand{\arraystretch}{1.25}
    \resizebox{1\linewidth}{!}{
    \begin{tabular}{cc|cccc}
            \toprule
    Pre-training & Downstream & No Defense & MCR            & SSL-Cleanse    & TSC (ours)   \\
    \hline
    & STL10      & 79.50      & \textbf{77.63} & 73.01          & 74.60 \\
    CIFAR10      & GTSRB      & 83.68      & 78.02          & \textbf{78.30} & 80.39 \\
                & SVHN       & 66.57      & 60.19          & \textbf{63.55} & 63.41 \\
    \hline
                & STL10      & 95.62      & \textbf{90.92} & 89.27          & 89.81 \\
    ImageNet100  & GTSRB      & 77.58      & \textbf{74.41} & 69.90          & 71.74 \\
                & SVHN       & 74.98      & \textbf{72.98} & 70.14          & 68.10 \\
    \bottomrule
            \end{tabular}
        }
    \label{tab:ssl_clean}
    \vspace*{-0.5em}

%% file: fig_tex/attack_lf.tex
\begin{figure*}[th]
    \centering
    \begin{subfigure}[BadNets (SBL-BadNets)\label{fig:sample_badnets_imagenet}]{
        \includegraphics[width=0.23\linewidth]{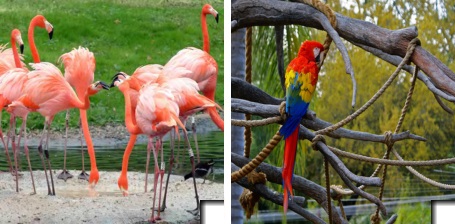}
    }    \end{subfigure}
    \hspace*{-0.5em}
    \begin{subfigure}[Blend (SBL-Blended)\label{fig:sample_blend_imagenet}]{
        \includegraphics[width=0.23\linewidth]{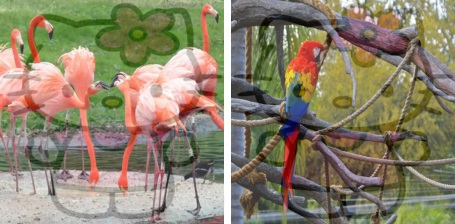}
    }\end{subfigure}
    \begin{subfigure}[SSBA\label{fig:sample_issba_imagenet}]{
        \includegraphics[width=0.23\linewidth]{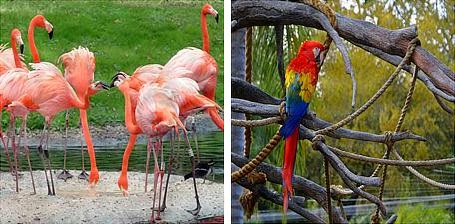}
    }\end{subfigure}
    \begin{subfigure}[LF\label{fig:sample_lf_imagenet}]{
        \includegraphics[width=0.23\linewidth]{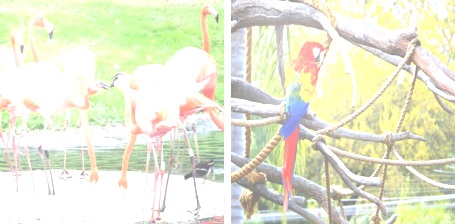}
    }\end{subfigure}\\
    \begin{subfigure}[InputAware\label{fig:sample_tsc_InputAware_imagenet}]{
        \includegraphics[width=0.23\linewidth]{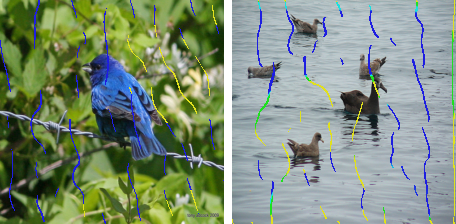}
    }    \end{subfigure}
    \hspace*{-0.5em}
    \begin{subfigure}[WaNet\label{fig:sample_tsc_wanet_imagenet}]{
        \includegraphics[width=0.23\linewidth]{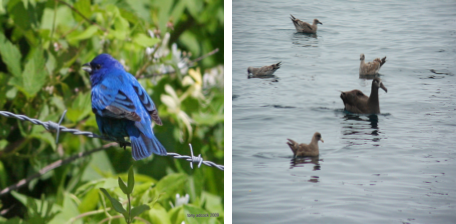}
    }\end{subfigure}
    \begin{subfigure}[LC\label{fig:sample_lc_imagenet}]{
        \includegraphics[width=0.23\linewidth]{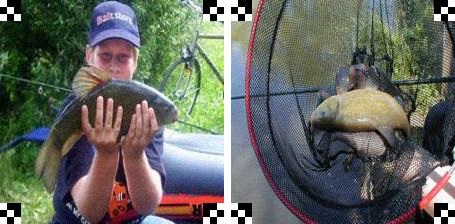}
    }
    \end{subfigure}
    \hspace*{-0.5em}
    \begin{subfigure}[SIG\label{fig:sample_sig_imagenet}]{
        \includegraphics[width=0.23\linewidth]{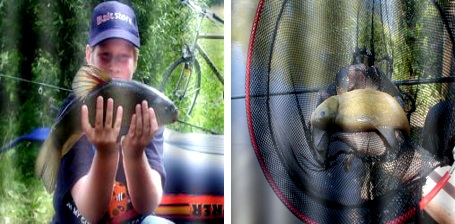}
    }\end{subfigure}
    \\
    \begin{subfigure}[SAPA\label{fig:sample_sapa_imagenet}]{
        \includegraphics[width=0.23\linewidth]{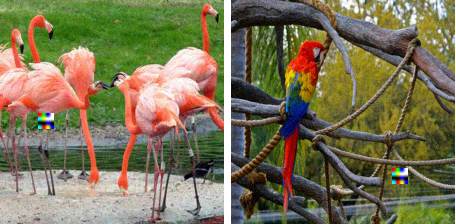}
    }
    \end{subfigure}
    \hspace*{-0.5em}
    \begin{subfigure}[Narcissus\label{fig:sample_narcissus_imagenet}]{
        \includegraphics[width=0.23\linewidth]{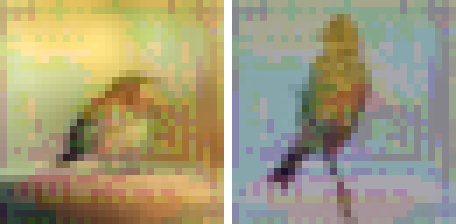}
    }\end{subfigure}
    \\
    \caption{Examples of 12 supervised learning backdoor trigger patterns on ImageNet100.
    The triggers of SBL-BadNets and SBL-Blended are as same as those of BadNets and Blend, respectively.
    }
    \label{attack_samples_imagenet}
\end{figure*}

%% file: manuscript_icml.bbl
\begin{thebibliography}{82}
\providecommand{\natexlab}[1]{#1}
\providecommand{\url}[1]{\texttt{#1}}
\expandafter\ifx\csname urlstyle\endcsname\relax
  \providecommand{\doi}[1]{doi: #1}\else
  \providecommand{\doi}{doi: \begingroup \urlstyle{rm}\Url}\fi

\bibitem[Ainsworth et~al.(2023)Ainsworth, Hayase, and Srinivasa]{git_rebasin}
Ainsworth, S., Hayase, J., and Srinivasa, S.
\newblock Git re-basin: Merging models modulo permutation symmetries.
\newblock In \emph{The Eleventh International Conference on Learning Representations}, 2023.
\newblock URL \url{https://openreview.net/forum?id=CQsmMYmlP5T}.

\bibitem[Alain \& Bengio(2017)Alain and Bengio]{linear_probe}
Alain, G. and Bengio, Y.
\newblock Understanding intermediate layers using linear classifier probes, 2017.
\newblock URL \url{https://openreview.net/forum?id=ryF7rTqgl}.

\bibitem[Barni et~al.(2019)Barni, Kallas, and Tondi]{SIG}
Barni, M., Kallas, K., and Tondi, B.
\newblock A new backdoor attack in cnns by training set corruption without label poisoning.
\newblock In \emph{2019 IEEE International Conference on Image Processing}, pp.\  101--105, 2019.

\bibitem[Bober-Irizar et~al.(2023)Bober-Irizar, Shumailov, Zhao, Mullins, and Papernot]{Architectural}
Bober-Irizar, M., Shumailov, I., Zhao, Y., Mullins, R., and Papernot, N.
\newblock Architectural backdoors in neural networks.
\newblock In \emph{Proceedings of the IEEE/CVF Conference on Computer Vision and Pattern Recognition (CVPR)}, pp.\  24595--24604, 2023.

\bibitem[Bossard et~al.(2014)Bossard, Guillaumin, and Van~Gool]{food101}
Bossard, L., Guillaumin, M., and Van~Gool, L.
\newblock Food-101 -- mining discriminative components with random forests.
\newblock In \emph{European Conference on Computer Vision}, 2014.

\bibitem[Carlini \& Terzis(2022)Carlini and Terzis]{poisoningCLIP}
Carlini, N. and Terzis, A.
\newblock Poisoning and backdooring contrastive learning.
\newblock In \emph{International Conference on Learning Representations}, 2022.
\newblock URL \url{https://openreview.net/forum?id=iC4UHbQ01Mp}.

\bibitem[Carlini et~al.(2023)Carlini, Jagielski, Choquette-Choo, Paleka, Pearce, Anderson, Terzis, Thomas, and Tram{\`e}r]{web_scale_poisoning}
Carlini, N., Jagielski, M., Choquette-Choo, C.~A., Paleka, D., Pearce, W., Anderson, H., Terzis, A., Thomas, K., and Tram{\`e}r, F.
\newblock Poisoning web-scale training datasets is practical.
\newblock \emph{arXiv preprint arXiv:2302.10149}, 2023.

\bibitem[Chai \& Chen(2022)Chai and Chen]{AWM}
Chai, S. and Chen, J.
\newblock One-shot neural backdoor erasing via adversarial weight masking.
\newblock In \emph{Thirty-Sixth Conference on Neural Information Processing Systems (NeurIPS 2022)}, 2022.

\bibitem[Chen et~al.(2018)Chen, Carvalho, Baracaldo, Ludwig, Edwards, Lee, Molloy, and Srivastava]{Clustering}
Chen, B., Carvalho, W., Baracaldo, N., Ludwig, H., Edwards, B., Lee, T., Molloy, I., and Srivastava, B.
\newblock Detecting backdoor attacks on deep neural networks by activation clustering.
\newblock \emph{ArXiv}, abs/1811.03728, 2018.

\bibitem[Chen et~al.(2020)Chen, Kornblith, Norouzi, and Hinton]{simCLR}
Chen, T., Kornblith, S., Norouzi, M., and Hinton, G.
\newblock A simple framework for contrastive learning of visual representations.
\newblock In \emph{Proceedings of the 37th International Conference on Machine Learning}, volume 119 of \emph{Proceedings of Machine Learning Research}, pp.\  1597--1607. PMLR, 2020.
\newblock URL \url{https://proceedings.mlr.press/v119/chen20j.html}.

\bibitem[Chen et~al.(2017)Chen, Liu, Li, Lu, and Song]{TargetGlasses}
Chen, X., Liu, C., Li, B., Lu, K., and Song, D.
\newblock Targeted backdoor attacks on deep learning systems using data poisoning.
\newblock \emph{arXiv preprint arXiv:1712.05526}, 2017.

\bibitem[Coates et~al.(2011)Coates, Ng, and Lee]{STL10}
Coates, A., Ng, A., and Lee, H.
\newblock An analysis of single-layer networks in unsupervised feature learning.
\newblock In \emph{Proceedings of the Fourteenth International Conference on Artificial Intelligence and Statistics}, volume~15 of \emph{Proceedings of Machine Learning Research}, pp.\  215--223, Fort Lauderdale, FL, USA, 11--13 Apr 2011. PMLR.
\newblock URL \url{https://proceedings.mlr.press/v15/coates11a.html}.

\bibitem[Deng et~al.(2009)Deng, Dong, Socher, Li, Li, and Fei-Fei]{ImageNet}
Deng, J., Dong, W., Socher, R., Li, L.-J., Li, K., and Fei-Fei, L.
\newblock Imagenet: A large-scale hierarchical image database.
\newblock In \emph{2009 IEEE Conference on Computer Vision and Pattern Recognition}, pp.\  248--255, 2009.
\newblock \doi{10.1109/CVPR.2009.5206848}.

\bibitem[Draxler et~al.(2018)Draxler, Veschgini, Salmhofer, and Hamprecht]{lossLandscape}
Draxler, F., Veschgini, K., Salmhofer, M., and Hamprecht, F.
\newblock Essentially no barriers in neural network energy landscape.
\newblock In \emph{Proceedings of the 35th International Conference on Machine Learning}, volume~80, pp.\  1309--1318. PMLR, 10--15 Jul 2018.

\bibitem[Entezari et~al.(2022)Entezari, Sedghi, Saukh, and Neyshabur]{permutation}
Entezari, R., Sedghi, H., Saukh, O., and Neyshabur, B.
\newblock The role of permutation invariance in linear mode connectivity of neural networks.
\newblock In \emph{International Conference on Learning Representations}, 2022.
\newblock URL \url{https://openreview.net/forum?id=dNigytemkL}.

\bibitem[Everingham et~al.()Everingham, Van~Gool, Williams, Winn, and Zisserman]{pascal-voc-2007}
Everingham, M., Van~Gool, L., Williams, C. K.~I., Winn, J., and Zisserman, A.
\newblock The {PASCAL} {V}isual {O}bject {C}lasses {C}hallenge 2007 {(VOC2007)} {R}esults.
\newblock http://www.pascal-network.org/challenges/VOC/voc2007/workshop/index.html.

\bibitem[Feng et~al.(2023)Feng, Tao, Cheng, Shen, Xu, Liu, Zhang, Ma, and Zhang]{detect_encoder}
Feng, S., Tao, G., Cheng, S., Shen, G., Xu, X., Liu, Y., Zhang, K., Ma, S., and Zhang, X.
\newblock Detecting backdoors in pre-trained encoders.
\newblock In \emph{2023 IEEE/CVF Conference on Computer Vision and Pattern Recognition (CVPR)}, pp.\  16352--16362, 2023.
\newblock URL \url{https://doi.ieeecomputersociety.org/10.1109/CVPR52729.2023.01569}.

\bibitem[Foret et~al.(2021)Foret, Kleiner, Mobahi, and Neyshabur]{sharpnessaware}
Foret, P., Kleiner, A., Mobahi, H., and Neyshabur, B.
\newblock Sharpness-aware minimization for efficiently improving generalization.
\newblock In \emph{International Conference on Learning Representations}, 2021.

\bibitem[Frankle et~al.(2020)Frankle, Dziugaite, Roy, and Carbin]{LMC}
Frankle, J., Dziugaite, G.~K., Roy, D., and Carbin, M.
\newblock Linear mode connectivity and the lottery ticket hypothesis.
\newblock In \emph{Proceedings of the 37th International Conference on Machine Learning}, volume 119, pp.\  3259--3269. PMLR, 2020.
\newblock URL \url{https://proceedings.mlr.press/v119/frankle20a.html}.

\bibitem[Gao et~al.(2019)Gao, Xu, Wang, Chen, Ranasinghe, and Nepal]{STRIP}
Gao, Y., Xu, C., Wang, D., Chen, S., Ranasinghe, D.~C., and Nepal, S.
\newblock Strip: A defence against trojan attacks on deep neural networks.
\newblock In \emph{Proceedings of the 35th Annual Computer Security Applications Conference}, pp.\  113–125, 2019.
\newblock ISBN 9781450376280.

\bibitem[Garipov et~al.(2018)Garipov, Izmailov, Podoprikhin, Vetrov, and Wilson]{ModeConnectivity}
Garipov, T., Izmailov, P., Podoprikhin, D., Vetrov, D.~P., and Wilson, A.~G.
\newblock Loss surfaces, mode connectivity, and fast ensembling of dnns.
\newblock In \emph{Advances in Neural Information Processing Systems (NeurIPS 2018)}, volume~31. Curran Associates, Inc., 2018.

\bibitem[Gotmare et~al.(2018)Gotmare, Keskar, Xiong, and Socher]{SGDR_mc}
Gotmare, A.~D., Keskar, N.~S., Xiong, C., and Socher, R.
\newblock Using mode connectivity for loss landscape analysis.
\newblock \emph{ArXiv}, abs/1806.06977, 2018.

\bibitem[Gu et~al.(2017)Gu, Dolan-Gavitt, and Garg]{badnets}
Gu, T., Dolan-Gavitt, B., and Garg, S.
\newblock Badnets: Identifying vulnerabilities in the machine learning model supply chain.
\newblock \emph{arXiv preprint arXiv:1708.06733}, 2017.

\bibitem[Guo et~al.(2023)Guo, Li, Chen, Guo, Sun, and Liu]{scaleup}
Guo, J., Li, Y., Chen, X., Guo, H., Sun, L., and Liu, C.
\newblock {SCALE}-{UP}: An efficient black-box input-level backdoor detection via analyzing scaled prediction consistency.
\newblock In \emph{The Eleventh International Conference on Learning Representations}, 2023.
\newblock URL \url{https://openreview.net/forum?id=o0LFPcoFKnr}.

\bibitem[Hayase et~al.(2021)Hayase, Kong, Somani, and Oh]{SPECTRE}
Hayase, J., Kong, W., Somani, R., and Oh, S.
\newblock Spectre: defending against backdoor attacks using robust statistics.
\newblock In \emph{Proceedings of the 38th International Conference on Machine Learning}, volume 139 of \emph{Proceedings of Machine Learning Research}, pp.\  4129--4139. PMLR, 18--24 Jul 2021.

\bibitem[He et~al.(2016{\natexlab{a}})He, Zhang, Ren, and Sun]{Pre-Act-ResNet}
He, K., Zhang, X., Ren, S., and Sun, J.
\newblock Identity mappings in deep residual networks.
\newblock \emph{ArXiv}, abs/1603.05027, 2016{\natexlab{a}}.

\bibitem[He et~al.(2016{\natexlab{b}})He, Zhang, Ren, and Sun]{ResNet}
He, K., Zhang, X., Ren, S., and Sun, J.
\newblock Deep residual learning for image recognition.
\newblock In \emph{2016 IEEE Conference on Computer Vision and Pattern Recognition (CVPR)}, pp.\  770--778, jun 2016{\natexlab{b}}.

\bibitem[He et~al.(2022)He, Chen, Xie, Li, Doll\'ar, and Girshick]{MAE}
He, K., Chen, X., Xie, S., Li, Y., Doll\'ar, P., and Girshick, R.
\newblock Masked autoencoders are scalable vision learners.
\newblock In \emph{Proceedings of the IEEE/CVF Conference on Computer Vision and Pattern Recognition (CVPR)}, pp.\  16000--16009, June 2022.

\bibitem[He et~al.(2024)He, Xu, Ren, Cui, Zeng, Liu, Aggarwal, and Tang]{sapa}
He, P., Xu, H., Ren, J., Cui, Y., Zeng, S., Liu, H., Aggarwal, C., and Tang, J.
\newblock Sharpness-aware data poisoning attack.
\newblock In \emph{International Conference on Learning Representations}, 2024.

\bibitem[Hou et~al.(2024)Hou, Feng, Hua, Luo, Zhang, and Li]{IBD}
Hou, L., Feng, R., Hua, Z., Luo, W., Zhang, L.~Y., and Li, Y.
\newblock {IBD}-{PSC}: Input-level backdoor detection via parameter-oriented scaling consistency.
\newblock In \emph{Proceedings of the 41st International Conference on Machine Learning}, pp.\  18992--19022. PMLR, 2024.
\newblock URL \url{https://proceedings.mlr.press/v235/hou24a.html}.

\bibitem[Houben et~al.(2013)Houben, Stallkamp, Salmen, Schlipsing, and Igel]{gtsrb}
Houben, S., Stallkamp, J., Salmen, J., Schlipsing, M., and Igel, C.
\newblock Detection of traffic signs in real-world images: The {G}erman {T}raffic {S}ign {D}etection {B}enchmark.
\newblock In \emph{International Joint Conference on Neural Networks}, number 1288, 2013.

\bibitem[Jia et~al.(2022)Jia, Liu, and Gong]{BadEncoder}
Jia, J., Liu, Y., and Gong, N.~Z.
\newblock {BadEncoder}: Backdoor attacks to pre-trained encoders in self-supervised learning.
\newblock In \emph{IEEE Symposium on Security and Privacy}, 2022.

\bibitem[Jordan et~al.(2023)Jordan, Sedghi, Saukh, Entezari, and Neyshabur]{repair}
Jordan, K., Sedghi, H., Saukh, O., Entezari, R., and Neyshabur, B.
\newblock {REPAIR}: {RE}normalizing permuted activations for interpolation repair.
\newblock In \emph{The Eleventh International Conference on Learning Representations}, 2023.
\newblock URL \url{https://openreview.net/forum?id=gU5sJ6ZggcX}.

\bibitem[Khaddaj et~al.(2023)Khaddaj, Leclerc, Makelov, Georgiev, Salman, Ilyas, and Madry]{Rethinking_BD}
Khaddaj, A., Leclerc, G., Makelov, A., Georgiev, K., Salman, H., Ilyas, A., and Madry, A.
\newblock Rethinking backdoor attacks.
\newblock In \emph{Proceedings of the 40th International Conference on Machine Learning}, volume 202 of \emph{Proceedings of Machine Learning Research}, pp.\  16216--16236. PMLR, 23--29 Jul 2023.
\newblock URL \url{https://proceedings.mlr.press/v202/khaddaj23a.html}.

\bibitem[Kirkpatrick et~al.(2017)Kirkpatrick, Pascanu, Rabinowitz, Veness, Desjardins, Rusu, Milan, Quan, Ramalho, Grabska-Barwinska, Hassabis, Clopath, Kumaran, and Hadsell]{ewc}
Kirkpatrick, J., Pascanu, R., Rabinowitz, N., Veness, J., Desjardins, G., Rusu, A.~A., Milan, K., Quan, J., Ramalho, T., Grabska-Barwinska, A., Hassabis, D., Clopath, C., Kumaran, D., and Hadsell, R.
\newblock Overcoming catastrophic forgetting in neural networks.
\newblock \emph{Proceedings of the National Academy of Sciences}, 114\penalty0 (13):\penalty0 3521--3526, 2017.
\newblock \doi{10.1073/pnas.1611835114}.
\newblock URL \url{https://www.pnas.org/doi/abs/10.1073/pnas.1611835114}.

\bibitem[Krizhevsky(2009)]{CIFAR10}
Krizhevsky, A.
\newblock Learning multiple layers of features from tiny images.
\newblock 2009.

\bibitem[Kuhn(1955)]{hungarian}
Kuhn, H.~W.
\newblock The hungarian method for the assignment problem.
\newblock \emph{Naval research logistics quarterly}, 2\penalty0 (1-2):\penalty0 83--97, 1955.

\bibitem[Li et~al.(2023)Li, Pang, Xi, Du, Ji, Yao, and Wang]{embarrassing}
Li, C., Pang, R., Xi, Z., Du, T., Ji, S., Yao, Y., and Wang, T.
\newblock An embarrassingly simple backdoor attack on self-supervised learning.
\newblock In \emph{Proceedings of the IEEE/CVF International Conference on Computer Vision (ICCV)}, pp.\  4367--4378, October 2023.

\bibitem[Li et~al.(2021{\natexlab{a}})Li, Xue, Zhao, Zhu, and Zhang]{BackdoorStega}
Li, S., Xue, M., Zhao, B. Z.~H., Zhu, H., and Zhang, X.
\newblock Invisible backdoor attacks on deep neural networks via steganography and regularization.
\newblock \emph{IEEE Transactions on Dependable and Secure Computing}, 18:\penalty0 2088--2105, 2021{\natexlab{a}}.

\bibitem[Li et~al.(2015)Li, Yosinski, Clune, Lipson, and Hopcroft]{convergent_learning}
Li, Y., Yosinski, J., Clune, J., Lipson, H., and Hopcroft, J.
\newblock Convergent learning: Do different neural networks learn the same representations?
\newblock In \emph{Proceedings of the 1st International Workshop on Feature Extraction: Modern Questions and Challenges at NIPS 2015}, volume~44 of \emph{Proceedings of Machine Learning Research}, pp.\  196--212, Montreal, Canada, 11 Dec 2015. PMLR.
\newblock URL \url{https://proceedings.mlr.press/v44/li15convergent.html}.

\bibitem[Li et~al.(2021{\natexlab{b}})Li, Koren, Lyu, Lyu, Li, and Ma]{NAD}
Li, Y., Koren, N., Lyu, L., Lyu, X., Li, B., and Ma, X.
\newblock Neural attention distillation: Erasing backdoor triggers from deep neural networks.
\newblock \emph{ArXiv}, abs/2101.05930, 2021{\natexlab{b}}.

\bibitem[Li et~al.(2021{\natexlab{c}})Li, Li, Wu, Li, He, and Lyu]{ISSBA}
Li, Y., Li, Y., Wu, B., Li, L., He, R., and Lyu, S.
\newblock Invisible backdoor attack with sample-specific triggers.
\newblock In \emph{IEEE International Conference on Computer Vision (ICCV)}, 2021{\natexlab{c}}.

\bibitem[Li et~al.(2021{\natexlab{d}})Li, Lyu, Koren, Lyu, Li, and Ma]{ABL}
Li, Y., Lyu, X., Koren, N., Lyu, L., Li, B., and Ma, X.
\newblock Anti-backdoor learning: Training clean models on poisoned data.
\newblock \emph{Advances in Neural Information Processing Systems (NeurIPS 2021)}, 34:\penalty0 14900--14912, 2021{\natexlab{d}}.

\bibitem[Lin et~al.(2014)Lin, Maire, Belongie, Hays, Perona, Ramanan, Doll{\'a}r, and Zitnick]{MSCOCO}
Lin, T.-Y., Maire, M., Belongie, S., Hays, J., Perona, P., Ramanan, D., Doll{\'a}r, P., and Zitnick, C.~L.
\newblock Microsoft coco: Common objects in context.
\newblock In \emph{Computer Vision--ECCV 2014: 13th European Conference, Zurich, Switzerland, September 6-12, 2014, Proceedings, Part V 13}, pp.\  740--755. Springer, 2014.

\bibitem[Liu et~al.(2018)Liu, Dolan-Gavitt, and Garg]{FinePruning}
Liu, K., Dolan-Gavitt, B., and Garg, S.
\newblock Fine-pruning: Defending against backdooring attacks on deep neural networks.
\newblock In \emph{RAID}, 2018.

\bibitem[Loshchilov \& Hutter(2016)Loshchilov and Hutter]{SGDR}
Loshchilov, I. and Hutter, F.
\newblock Sgdr: Stochastic gradient descent with warm restarts.
\newblock \emph{arXiv preprint arXiv:1608.03983}, 2016.

\bibitem[Min et~al.(2023)Min, Qin, Shen, and Cheng]{towardsStable}
Min, R., Qin, Z., Shen, L., and Cheng, M.
\newblock Towards stable backdoor purification through feature shift tuning.
\newblock In \emph{Thirty-seventh Conference on Neural Information Processing Systems}, 2023.

\bibitem[Netzer et~al.(2011)Netzer, Wang, Coates, Bissacco, Wu, and Ng]{SVHN}
Netzer, Y., Wang, T., Coates, A., Bissacco, A., Wu, B., and Ng, A.~Y.
\newblock Reading digits in natural images with unsupervised feature learning.
\newblock In \emph{NIPS Workshop on Deep Learning and Unsupervised Feature Learning 2011}, 2011.
\newblock URL \url{http://ufldl.stanford.edu/housenumbers/nips2011_housenumbers.pdf}.

\bibitem[Nguyen \& Tran(2020)Nguyen and Tran]{Inputaware}
Nguyen, T.~A. and Tran, A.
\newblock Input-aware dynamic backdoor attack.
\newblock In \emph{Advances in Neural Information Processing Systems (NeurIPS 2020)}, volume~33, pp.\  3454--3464. Curran Associates, Inc., 2020.

\bibitem[Nguyen \& Tran(2021)Nguyen and Tran]{wanet}
Nguyen, T.~A. and Tran, A.~T.
\newblock Wanet - imperceptible warping-based backdoor attack.
\newblock In \emph{International Conference on Learning Representations}, 2021.
\newblock URL \url{https://openreview.net/forum?id=eEn8KTtJOx}.

\bibitem[Pan et~al.(2023)Pan, Zeng, Lyu, Lin, and Jia]{asset}
Pan, M., Zeng, Y., Lyu, L., Lin, X., and Jia, R.
\newblock {ASSET}: Robust backdoor data detection across a multiplicity of deep learning paradigms.
\newblock In \emph{32nd USENIX Security Symposium (USENIX Security 23)}, pp.\  2725--2742, Anaheim, CA, 2023. USENIX Association.
\newblock ISBN 978-1-939133-37-3.
\newblock URL \url{https://www.usenix.org/conference/usenixsecurity23/presentation/pan}.

\bibitem[Pang et~al.(2019)Pang, Shen, Zhang, Ji, Vorobeychik, Luo, Liu, and Wang]{Tale}
Pang, R., Shen, H., Zhang, X., Ji, S., Vorobeychik, Y., Luo, X., Liu, A.~X., and Wang, T.
\newblock A tale of evil twins: Adversarial inputs versus poisoned models.
\newblock \emph{Proceedings of the 2020 ACM SIGSAC Conference on Computer and Communications Security}, 2019.

\bibitem[Pham et~al.(2024)Pham, Ta, Tran, and Doan]{sbl}
Pham, H., Ta, T.-A., Tran, A., and Doan, K.~D.
\newblock Flatness-aware sequential learning generates resilient backdoors.
\newblock In \emph{Computer Vision -- ECCV 2024}, pp.\  89--107, Cham, 2024. Springer Nature Switzerland.

\bibitem[Qi et~al.(2023)Qi, Xie, Li, Mahloujifar, and Mittal]{adap_Backdoor}
Qi, X., Xie, T., Li, Y., Mahloujifar, S., and Mittal, P.
\newblock Revisiting the assumption of latent separability for backdoor defenses.
\newblock In \emph{The Eleventh International Conference on Learning Representations}, 2023.

\bibitem[Radford et~al.(2021)Radford, Kim, Hallacy, Ramesh, Goh, Agarwal, Sastry, Askell, Mishkin, Clark, Krueger, and Sutskever]{clip}
Radford, A., Kim, J.~W., Hallacy, C., Ramesh, A., Goh, G., Agarwal, S., Sastry, G., Askell, A., Mishkin, P., Clark, J., Krueger, G., and Sutskever, I.
\newblock Learning transferable visual models from natural language supervision.
\newblock In \emph{Proceedings of the 38th International Conference on Machine Learning}, volume 139 of \emph{Proceedings of Machine Learning Research}, pp.\  8748--8763. PMLR, 2021.
\newblock URL \url{https://proceedings.mlr.press/v139/radford21a.html}.

\bibitem[Saha et~al.(2020)Saha, Subramanya, and Pirsiavash]{HiddenTrigger}
Saha, A., Subramanya, A., and Pirsiavash, H.
\newblock Hidden trigger backdoor attacks.
\newblock In \emph{Proceedings of the AAAI conference on artificial intelligence}, volume~34, pp.\  11957--11965, 2020.

\bibitem[Saha et~al.(2022)Saha, Tejankar, Koohpayegani, and Pirsiavash]{backdoor_ssl_cvpr}
Saha, A., Tejankar, A., Koohpayegani, S.~A., and Pirsiavash, H.
\newblock Backdoor attacks on self-supervised learning.
\newblock In \emph{Proceedings of the IEEE/CVF Conference on Computer Vision and Pattern Recognition}, pp.\  13337--13346, 2022.

\bibitem[Shafahi et~al.(2018)Shafahi, Huang, Najibi, Suciu, Studer, Dumitras, and Goldstein]{PoisonFrog}
Shafahi, A., Huang, W.~R., Najibi, M., Suciu, O., Studer, C., Dumitras, T., and Goldstein, T.
\newblock Poison frogs! targeted clean-label poisoning attacks on neural networks.
\newblock In \emph{Advances in Neural Information Processing Systems (NeurIPS 2018)}, volume~31. Curran Associates, Inc., 2018.

\bibitem[Simonyan \& Zisserman(2014)Simonyan and Zisserman]{vgg19}
Simonyan, K. and Zisserman, A.
\newblock Very deep convolutional networks for large-scale image recognition.
\newblock \emph{CoRR}, abs/1409.1556, 2014.

\bibitem[Simsek et~al.(2021)Simsek, Ged, Jacot, Spadaro, Hongler, Gerstner, and Brea]{geometry_lossLandscape}
Simsek, B., Ged, F., Jacot, A., Spadaro, F., Hongler, C., Gerstner, W., and Brea, J.
\newblock Geometry of the loss landscape in overparameterized neural networks: Symmetries and invariances.
\newblock In \emph{Proceedings of the 38th International Conference on Machine Learning}, volume 139 of \emph{Proceedings of Machine Learning Research}, pp.\  9722--9732. PMLR, 18--24 Jul 2021.
\newblock URL \url{https://proceedings.mlr.press/v139/simsek21a.html}.

\bibitem[Singh \& Jaggi(2020)Singh and Jaggi]{model_fusion}
Singh, S.~P. and Jaggi, M.
\newblock Model fusion via optimal transport.
\newblock In Larochelle, H., Ranzato, M., Hadsell, R., Balcan, M., and Lin, H. (eds.), \emph{Advances in Neural Information Processing Systems}, volume~33, pp.\  22045--22055. Curran Associates, Inc., 2020.

\bibitem[Souri et~al.(2022)Souri, Fowl, Chellappa, Goldblum, and Goldstein]{sleepagent}
Souri, H., Fowl, L., Chellappa, R., Goldblum, M., and Goldstein, T.
\newblock Sleeper agent: Scalable hidden trigger backdoors for neural networks trained from scratch.
\newblock In \emph{Advances in Neural Information Processing Systems (NeurIPS 2022)}, volume~35, pp.\  19165--19178, 2022.

\bibitem[Szegedy et~al.(2016)Szegedy, Vanhoucke, Ioffe, Shlens, and Wojna]{inceptionv3}
Szegedy, C., Vanhoucke, V., Ioffe, S., Shlens, J., and Wojna, Z.
\newblock Rethinking the inception architecture for computer vision.
\newblock In \emph{Proceedings of the IEEE conference on computer vision and pattern recognition}, pp.\  2818--2826, 2016.

\bibitem[Tancik et~al.(2020)Tancik, Mildenhall, and Ng]{Stega}
Tancik, M., Mildenhall, B., and Ng, R.
\newblock Stegastamp: Invisible hyperlinks in physical photographs.
\newblock \emph{2020 IEEE/CVF Conference on Computer Vision and Pattern Recognition (CVPR)}, pp.\  2114--2123, 2020.

\bibitem[Tatro et~al.(2020)Tatro, Chen, Das, Melnyk, Sattigeri, and Lai]{optMC_na}
Tatro, N., Chen, P.-Y., Das, P., Melnyk, I., Sattigeri, P., and Lai, R.
\newblock Optimizing mode connectivity via neuron alignment.
\newblock In \emph{Advances in Neural Information Processing Systems}, volume~33, pp.\  15300--15311. Curran Associates, Inc., 2020.

\bibitem[Tran et~al.(2018)Tran, Li, and Madry]{PCA}
Tran, B., Li, J., and Madry, A.
\newblock Spectral signatures in backdoor attacks.
\newblock In \emph{Neural Information Processing Systems (NeurIPS 2018)}, 2018.

\bibitem[Turner et~al.(2019)Turner, Tsipras, and Madry]{firstCleanLabel}
Turner, A., Tsipras, D., and Madry, A.
\newblock Label-consistent backdoor attacks.
\newblock \emph{ArXiv}, abs/1912.02771, 2019.

\bibitem[Villani et~al.(2009)]{optimal_transport}
Villani, C. et~al.
\newblock \emph{Optimal transport: old and new}, volume 338.
\newblock Springer, 2009.

\bibitem[Wang et~al.(2019)Wang, Yao, Shan, Li, Viswanath, Zheng, and Zhao]{NeuralCleanse}
Wang, B., Yao, Y., Shan, S., Li, H., Viswanath, B., Zheng, H., and Zhao, B.~Y.
\newblock Neural cleanse: Identifying and mitigating backdoor attacks in neural networks.
\newblock \emph{2019 IEEE Symposium on Security and Privacy (SP)}, pp.\  707--723, 2019.

\bibitem[Wei et~al.(2023)Wei, Zhang, Zha, and Wu]{sau}
Wei, S., Zhang, M., Zha, H., and Wu, B.
\newblock Shared adversarial unlearning: Backdoor mitigation by unlearning shared adversarial examples.
\newblock In \emph{Thirty-seventh Conference on Neural Information Processing Systems}, 2023.
\newblock URL \url{https://openreview.net/forum?id=zqOcW3R9rd}.

\bibitem[Wenger et~al.(2022)Wenger, Bhattacharjee, Bhagoji, Passananti, Andere, Zheng, and Zhao]{FindingNaturally}
Wenger, E., Bhattacharjee, R., Bhagoji, A.~N., Passananti, J., Andere, E., Zheng, H., and Zhao, B.
\newblock Finding naturally occurring physical backdoors in image datasets.
\newblock In \emph{Advances in Neural Information Processing Systems}, volume~35, pp.\  22103--22116, 2022.

\bibitem[Wortsman et~al.(2021)Wortsman, Horton, Guestrin, Farhadi, and Rastegari]{learning_subspace}
Wortsman, M., Horton, M.~C., Guestrin, C., Farhadi, A., and Rastegari, M.
\newblock Learning neural network subspaces.
\newblock In \emph{Proceedings of the 38th International Conference on Machine Learning}, volume 139 of \emph{Proceedings of Machine Learning Research}, pp.\  11217--11227. PMLR, 18--24 Jul 2021.
\newblock URL \url{https://proceedings.mlr.press/v139/wortsman21a.html}.

\bibitem[Wu et~al.(2022)Wu, Chen, Zhang, Zhu, Wei, Yuan, and Shen]{backdoorbench}
Wu, B., Chen, H., Zhang, M., Zhu, Z., Wei, S., Yuan, D., and Shen, C.
\newblock Backdoorbench: A comprehensive benchmark of backdoor learning.
\newblock In \emph{Thirty-sixth Conference on Neural Information Processing Systems Datasets and Benchmarks Track}, 2022.

\bibitem[Wu \& Wang(2021)Wu and Wang]{ANP}
Wu, D. and Wang, Y.
\newblock Adversarial neuron pruning purifies backdoored deep models.
\newblock In \emph{Advances in Neural Information Processing Systems (NeurIPS 2021)}, volume~34, pp.\  16913--16925. Curran Associates, Inc., 2021.

\bibitem[Yao et~al.(2019)Yao, Li, Zheng, and Zhao]{Latent}
Yao, Y., Li, H., Zheng, H., and Zhao, B.~Y.
\newblock Latent backdoor attacks on deep neural networks.
\newblock In \emph{Proceedings of the ACM SIGSAC Conference on Computer and Communications Security}, CCS '19, pp.\  2041–2055, 2019.
\newblock URL \url{https://doi.org/10.1145/3319535.3354209}.

\bibitem[Zeng et~al.(2021)Zeng, Park, Mao, and Jia]{lf}
Zeng, Y., Park, W., Mao, Z., and Jia, R.
\newblock Rethinking the backdoor attacks’ triggers: A frequency perspective.
\newblock In \emph{2021 IEEE/CVF International Conference on Computer Vision (ICCV)}, pp.\  16453--16461, Los Alamitos, CA, USA, 2021. IEEE Computer Society.

\bibitem[Zeng et~al.(2022)Zeng, Chen, Park, Mao, Jin, and Jia]{I_BAU}
Zeng, Y., Chen, S., Park, W., Mao, Z., Jin, M., and Jia, R.
\newblock Adversarial unlearning of backdoors via implicit hypergradient.
\newblock In \emph{International Conference on Learning Representations}, 2022.

\bibitem[Zeng et~al.(2023)Zeng, Pan, Just, Lyu, Qiu, and Jia]{narcissus}
Zeng, Y., Pan, M., Just, H.~A., Lyu, L., Qiu, M., and Jia, R.
\newblock Narcissus: A practical clean-label backdoor attack with limited information.
\newblock In \emph{Proceedings of the 2023 ACM SIGSAC Conference on Computer and Communications Security}, pp.\  771–785, 2023.
\newblock ISBN 9798400700507.
\newblock \doi{10.1145/3576915.3616617}.

\bibitem[Zhao et~al.(2020)Zhao, Chen, Das, Ramamurthy, and Lin]{MCR}
Zhao, P., Chen, P.-Y., Das, P., Ramamurthy, K.~N., and Lin, X.
\newblock Bridging mode connectivity in loss landscapes and adversarial robustness.
\newblock In \emph{International Conference on Learning Representations}, 2020.

\bibitem[Zheng et~al.(2024)Zheng, Xue, Wang, Chen, Lou, Jiang, and Wang]{SSL_Cleanse}
Zheng, M., Xue, J., Wang, Z., Chen, X., Lou, Q., Jiang, L., and Wang, X.
\newblock Ssl-cleanse: Trojan detection and mitigation in self-supervised learning.
\newblock In \emph{Computer Vision -- ECCV 2024}, 2024.

\bibitem[Zhu et~al.(2019)Zhu, Huang, Li, Taylor, Studer, and Goldstein]{Transferable_clean-label}
Zhu, C., Huang, W.~R., Li, H., Taylor, G., Studer, C., and Goldstein, T.
\newblock Transferable clean-label poisoning attacks on deep neural nets.
\newblock In \emph{Proceedings of the 36th International Conference on Machine Learning}, volume~97, pp.\  7614--7623. PMLR, 09--15 Jun 2019.

\bibitem[Zhu et~al.(2023)Zhu, Wei, Shen, Fan, and Wu]{FT_SAM}
Zhu, M., Wei, S., Shen, L., Fan, Y., and Wu, B.
\newblock Enhancing fine-tuning based backdoor defense with sharpness-aware minimization.
\newblock In \emph{Proceedings of the IEEE/CVF International Conference on Computer Vision (ICCV)}, pp.\  4466--4477, October 2023.

\end{thebibliography}
